%% file: region.tex
\documentclass[11pt]{article}

\usepackage{amssymb,amsfonts,amsmath,amsthm,amscd,dsfont,mathrsfs}
\usepackage{graphicx,float,psfrag,epsfig}
\usepackage{wrapfig}
\usepackage{hyperref}
\usepackage{algorithm,algorithmic}

\usepackage{color}

\DeclareMathAlphabet{\mathpzc}{OT1}{pzc}{m}{it}

\newtheorem{propo}{Proposition}[]

\newtheorem{definition}[propo]{Definition}

\newtheorem{thm}[propo]{Theorem}

\newtheorem{remark}[propo]{Remark}

\def\de{{$(\varepsilon,\delta)$}}
\def\cR{{\cal R}}
\def\cX{{\cal X}}
\def\tX{{\tilde X}}

\def\reals{{\mathbb R}}
\def\prob{{\mathbb P}}

\def\E{\mathbb E}

\def\ind{{\mathbb I}}

\def\id{{\mathds I}}

\begin{document}

\title{PacGAN: The power of two samples in generative adversarial networks}

\author{
Zinan Lin$^\dagger$,  Ashish Khetan$^\ddagger$, Giulia Fanti$^\dagger$, Sewoong Oh$^\ddagger$\thanks{Author emails are {\text{zinanl@andrew.cmu.edu}, \text{khetan2@illinois.edu}, \text{gfanti@andrew.cmu.edu}, and \text{swoh@illinois.edu}}. 
This work used the Extreme Science and Engineering Discovery Environment (XSEDE), which is supported by National Science Foundation grant number OCI-1053575.  Specifically, it used the Bridges system, which is supported by NSF award number ACI-1445606, at the Pittsburgh Supercomputing Center (PSC).}\\
$^\dagger$Carnegie Mellon University, $^\ddagger$University of Illinois at Urbana-Champaign\\
}

\date{}

\maketitle

\begin{abstract}

Generative adversarial networks (GANs) are  innovative techniques for learning generative models of complex data distributions from samples. 
Despite remarkable recent improvements in generating  realistic images, one of their major shortcomings is the fact that in practice, they tend to produce samples with little diversity, even when trained on diverse datasets.
This phenomenon, known as mode collapse, has been the main focus 
of several recent advances in GANs. 
Yet there is little understanding of why mode collapse happens and 
why recently proposed approaches are able to  mitigate mode collapse. 
We propose a principled approach to handling mode collapse, 
which we call {\em packing}. 
The main idea is to modify the discriminator to make decisions based on multiple samples from the same class, 
either real or artificially generated.  
We borrow analysis tools from binary hypothesis testing---in particular the seminal result of Blackwell \cite{Bla53}---to prove a fundamental connection between packing and mode collapse. 
We show that packing  naturally penalizes generators with mode collapse, 
thereby favoring generator distributions with less mode collapse during the training process. 
Numerical experiments on benchmark datasets suggests that packing provides significant improvements in 
practice as well. 

\end{abstract}

%
%
\input{introduction.tex}

\input{section2-gf.tex}

\input{experiment.tex}

\section{Theoretical analyses of PacGAN}
\label{sec:theory}

In this section, we propose a formal and natural mathematical definition of mode collapse,  
which abstracts away domain-specific details (e.g.~images vs. time series). 
For a target distribution $P$ and a generator distribution $Q$,
this definition describes mode collapse through a two-dimensional representation of the pair $(P,Q)$ as a \emph{region}, which is motivated by the 
ROC (Receiver Operating Characteristic) curve representation of 
a binary hypothesis testing or a binary classification.

Mode collapse is a phenomenon commonly reported in the GAN literature \cite{Goo16,reed2016generative,TGB17,MT17,AZ17},
which can refer to two distinct concepts: 
$(i)$ the generative model loses some modes that are present in the samples of the target distribution. For example, despite being trained on a dataset of animal pictures that includes lizards, the model never generates images of lizards.
$(ii)$ Two distant points in the code vector $Z$ are mapped to the same or similar points in the sample space $X$.
For instance, two distant latent vectors $z_1$ and $z_2$ map to the same picture of a lizard \cite{Goo16}. 
Although these phenomena are different, and 
either one can occur without the other, 
they are generally not explicitly distinguished in the literature, 
and it has been suggested that the latter may cause the former \cite{Goo16}.  
In this paper, 
we focus on the former notion, as it does not depend on 
how the generator maps a code vector $Z$ to the sample $X$, 
and only focuses on the quality of the samples generated.  
In other words, we assume here that 
two generative models with the same marginal distribution over the generated samples should not be treated 
differently based on how random code vectors are mapped to the data sample space. 
The second notion of mode collapse would  differentiate two such architectures, and is beyond the scope of this work. 
The proposed region representation relies purely on the properties of the generated samples, 
and not on the generator's mapping between the latent and sample spaces.

We analyze how the proposed idea of packing changes the training of the generator. 
We view the discriminator's role as providing a surrogate for a desired loss to be minimized---surrogate in the sense that the actual desired losses, such as Jensen-Shannon divergence or total variation distances, 
cannot be computed exactly and need to be estimated. 
Consider the standard GAN discriminator with a cross-entropy loss: 
\begin{eqnarray}
	\label{eq:crossentropy}
	\min_G \;\;  \underbrace{\max_D \;\; \E_{X\sim P} [\log(D(X))] + \E_{G(Z)\sim Q}[ \log(1-D(G(Z)))]}_{\simeq \;\; d_{\rm KL}\big(P\| \frac{P+Q}{2}\big)+d_{\rm KL}\big(Q\| \frac{P+Q}{2}\big) + \log (1/4) } \;, 
\end{eqnarray} 
where the maximization is over the family of discriminators (or the discriminator weights, if the family is a neural network of a fixed architecture), 
the minimization is over the family of generators, and 
$X$ is drawn from the distribution $P$ of the real data,  
$Z$ is drawn from the distribution of the code vector, typically a low-dimensional Gaussian,
and we denote the resulting generator distribution as $G(Z)\sim Q$. 
The role of the discriminator under this GAN scenario is to provide the 
generator with an approximation (or a surrogate) of a loss, 
which in the case of cross entropy loss turns out to be the 
Jensen-Shannon divergence (up to a scaling and shift by a constant), defined as 
$d_{\rm JS}(P,Q) \triangleq (1/2)\,d_{\rm KL}(P\| {(P+Q)}/{2})+(1/2)\,d_{\rm KL}(Q\| {(P+Q)}/{2})$, where 
$d_{\rm KL}(\cdot)$ is the Kullback-Leibler divergence. 
This follows from the fact that, if we search for the maximizing discriminator over the space of all functions, 
 the maximizer turns out to be $D(X) = P(X)/(P(X)+Q(X))$ 
\cite{GPM14}.
In practice, we search over some parametric family of discriminators, 
and we can only compute sample average of the losses. 
This provides an approximation of the Jensen-Shannon divergence between $P$ and $Q$. 
The outer minimization over the generator tries to generate samples such that they are close to the real data in this (approximate) Jensen-Shannon divergence, 
which is one measure of how close the true distribution $P$ and the generator distribution $Q$ are.

In this section, we show a fundamental connection between  the principle of packing and mode collapse in GAN. 
We provide a complete understanding of how packing changes the  loss as seen by the generator, 
by focusing on (as we did to derive the Jensen-Shnnon divergence above) $(a)$ the optimal discriminator over a family of all measurable functions;  
$(b)$ the population expectation; and 
$(c)$  the $0$-$1$ loss function of the form:  
\begin{eqnarray*}
 	\max_D & & \E_{X\sim P} [\ind(D(X))]+ \E_{G(Z)\sim Q}[1-\ind(D(G(Z)))]\\
	\text{subject to} 	&& D(X) \in\{0,1\} \;.  
\end{eqnarray*}
The first assumption allows us to bypass the specific architecture of the discriminator used, 
which is common when analyzing neural network based discriminators (e.g.~\cite{BJPD17,BPD18}). 
The second assumption can be potentially relaxed and  the standard finite sample analysis can be 
applied to provide bounds similar to those in our main results in Theorems \ref{thm:main1}, \ref{thm:main2}, and \ref{thm:main3}. 
The last assumption gives a loss of the total variation distance 
$d_{\rm TV}(P,Q)\triangleq \sup_{S\subseteq \mathcal X} \{P(S)-Q(S)\}$  over the domain $\cX$. 
This follows from the fact that (e.g.~\cite{Goo16}), 
\begin{eqnarray*}
	\sup_D \big\{ \E_{X\sim P}[\ind(D(X))]+ \E_{G(Z)\sim Q}[1-\ind(D(G(Z)))]\big\} &=& \sup_S \big\{ P(S)+1-Q(S)\big\} \\ 
	&=&  1+d_{\rm TV}(P,Q) \;.
\end{eqnarray*}
This discriminator provides (an approximation of) the total variation distance, and 
 the generator  tries to minimize the total variation distance 
$d_{\rm TV}(P,Q)$. 
The reason we make this assumption is primarily for clarity and analytical tractability: total variation distance highlights the effect of packing in a way that is cleaner and easier to understand than if we were to analyze Jensen-Shannon divergence. 
We discuss this point in more detail in Section \ref{sec:productregion}.
In sum,  
these three assumptions allow us to focus purely on the impact of packing on the mode collapse of resulting discriminator.  

We want to understand how this 0-1 loss, as provided by such a discriminator, changes with the {\em degree of packing} $m$. 
As packed discriminators see $m$ packed samples, each drawn i.i.d. from one joint class 
(i.e.~either real or generated), 
we can consider these packed samples as a single sample that is drawn from the product distribution: $P^m$ for real and $Q^m$ for generated. 
The resulting loss provided by the packed discriminator is therefore  $d_{\rm TV}(P^m,Q^m)$. 

We first provide a formal mathematical definition of mode collapse in Section \ref{sec:definition}, 
which leads to a two-dimensional representation of any pair of distributions $(P,Q)$ as a {\em mode-collapse region}. 
This region representation provides not only  
 conceptual clarity regarding mode collapse, 
but also proof techniques that are essential to proving our main results on 
the fundamental connections between the strength of mode collapse in a pair $(P,Q)$ and 
the loss $d_{\rm TV}(P^m,Q^m)$ seen by a 
packed discriminator (Section \ref{sec:productregion}). 
The proofs of these results are provided in Section \ref{sec:proof}. 
In Section \ref{sec:region}, 
we show that the proposed mode collapse region is equivalent to
the ROC curve for binary hypothesis testing. 
This allows us to use powerful  
mathematical techniques from binary hypothesis testing 
including the data processing inequality and the reverse data processing inequalities. 


\subsection{Mathematical definition of mode collapse as a two-dimensional region} 
\label{sec:definition}
Although no formal and agreed-upon definition of mode collapse exists in the GAN literature, 
mode collapse is declared for a multimodal target distribution $P$
if the generator $Q$ assigns a significantly smaller probability density 
in the regions surrounding a particular subset of modes. 
One major challenge in addressing such a mode collapse is that 
it involves the geometry of $P$:
there is no standard partitioning of the domain respecting the modular topology of $P$, 
and even heuristic partitions are typically computationally intractable in high dimensions.
Hence, we drop this geometric constraint, and 
introduce a purely analytical definition. 
\begin{definition}
	\label{def:modecollapse}
	A target distribution $P$ and a generator $Q$ exhibit \de-{\em mode collapse} for some $0\leq \varepsilon < \delta\leq 1$ if there exists a set $S\subseteq \cX$ such that $P(S)\geq \delta$ and $Q(S) \leq \varepsilon$.
\end{definition} 
This definition provides a formal measure of mode collapse for a target $P$ and a generator $Q$; 
intuitively, larger $\delta$ and smaller $\varepsilon$ indicate more severe mode collapse.  
That is, if a large portion of the target $P(S)\geq \delta$ in some set $S$ in the domain $\cX$ is missing in the generator $Q(S)\leq \varepsilon$, then we declare $(\varepsilon,\delta)$-mode collapse. 

\begin{figure} [ht]
	\begin{center} 
	\includegraphics[width=.26\textwidth]{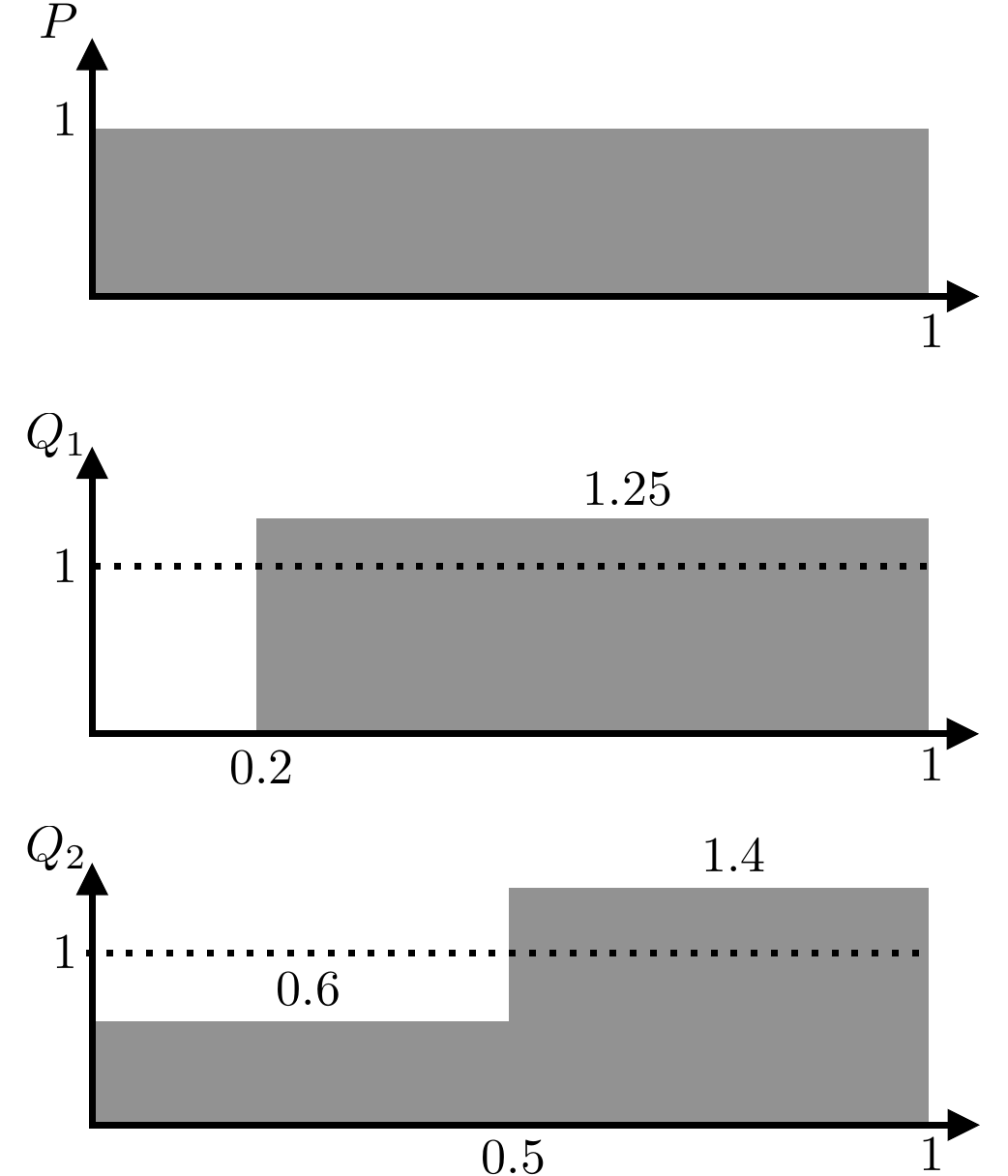}
	\includegraphics[width=.36\textwidth]{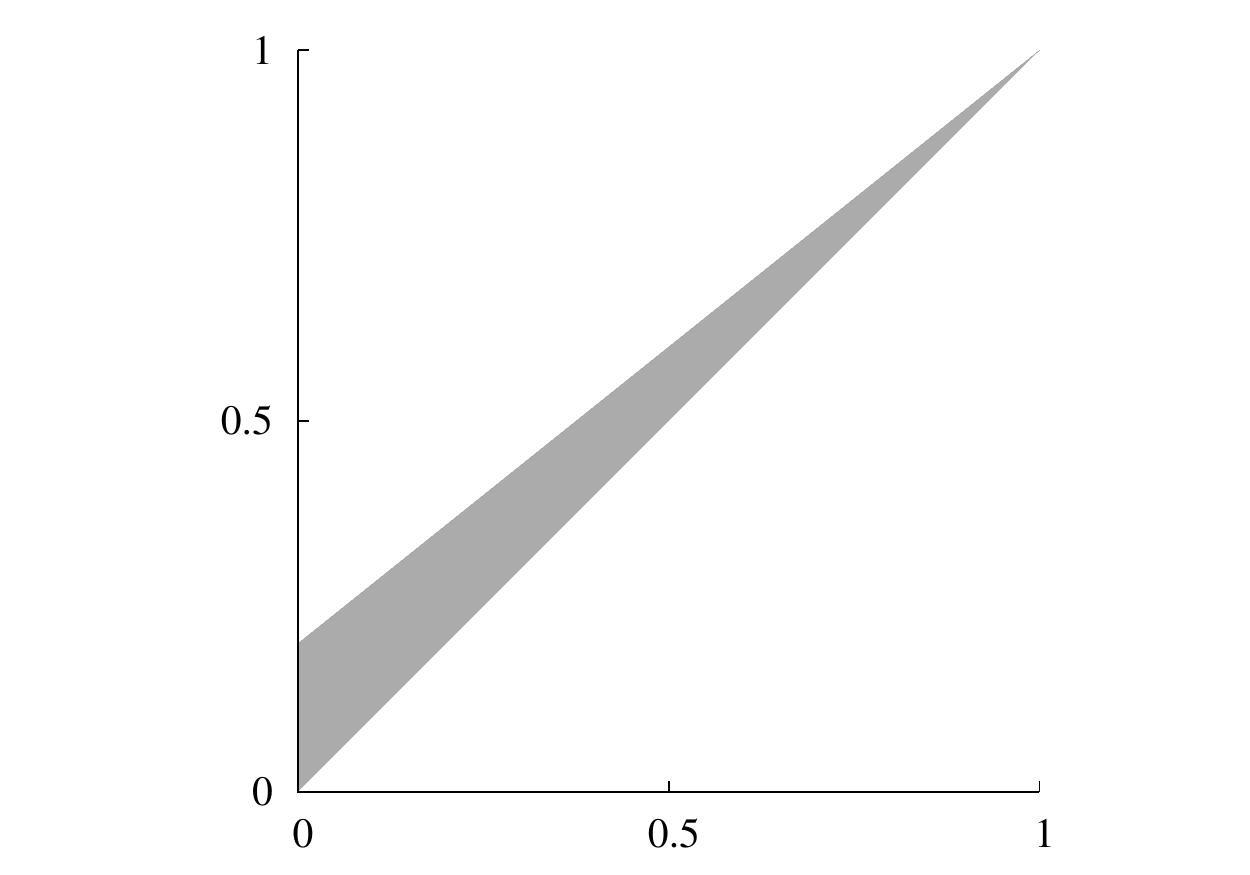}
	\put(-70,60){$\cR(P,Q_1)$}
	\put(-80,-4){$\varepsilon$}
	\put(-148,60){$\delta$}
	\includegraphics[width=.36\textwidth]{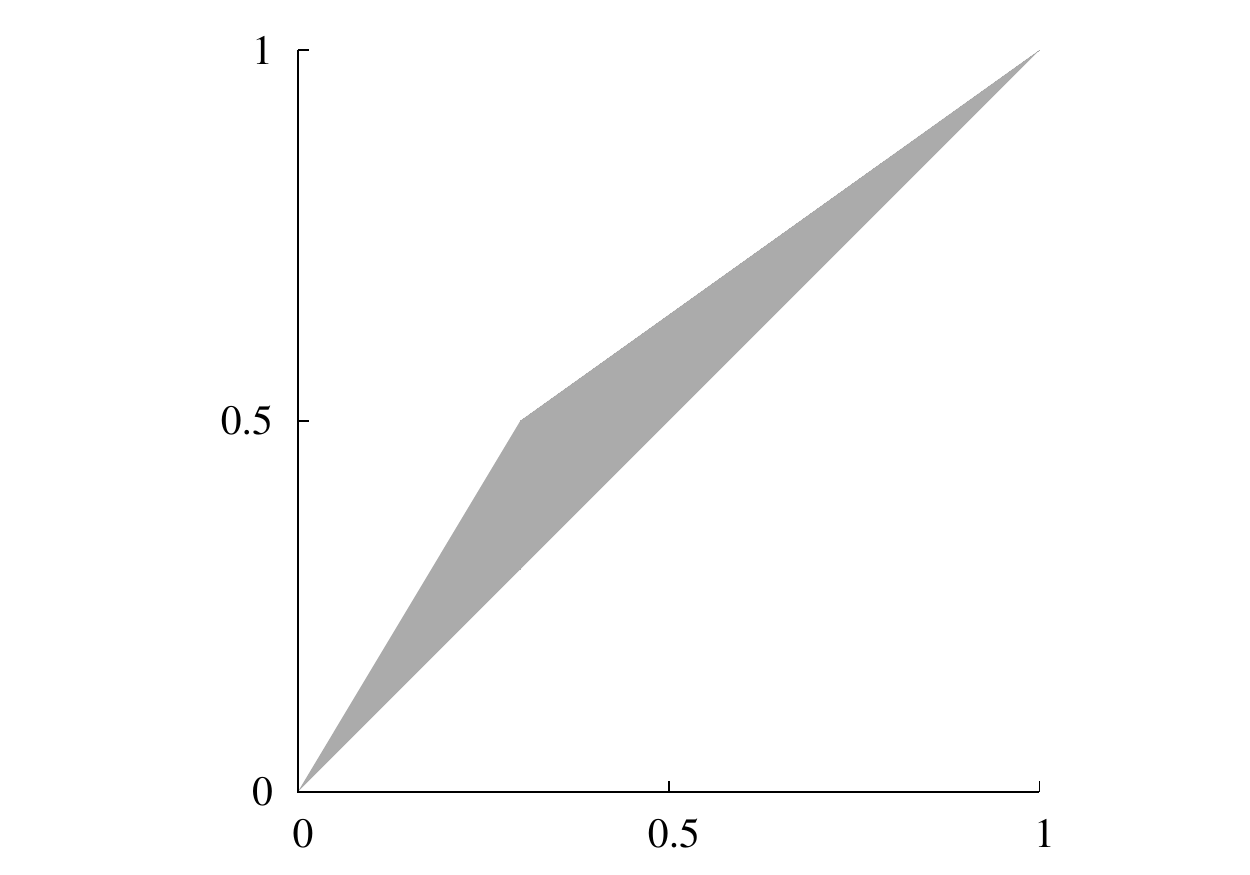}
	\put(-70,60){$\cR(P,Q_2)$}
	\put(-80,-4){$\varepsilon$}
	\put(-148,60){$\delta$}
	\end{center} 
	\caption{A formal definition of $(\varepsilon,\delta)$-mode collapse and its accompanying region representation 
	captures the intensity of mode collapse for generators $Q_1$ with mode collapse and 
			 $Q_2$ which does not have mode collapse, for a toy example distributions  $P$, $Q_1$, and $Q_2$ shown on the left. 
			The region of $(\varepsilon,\delta)$-mode collapse that is achievable is shown in  grey.  } 
	\label{fig:toy} 
\end{figure}

A key observation is that \emph{two pairs of distributions can have the same total variation distance while exhibiting very different mode collapse patterns.}
To see this, consider a toy example in Figure \ref{fig:toy}, with a uniform target  distribution $P=U([0,1])$
 over $[0,1]$. 
Now consider all generators at a fixed total variation distance of $0.2$ from $P$.
We compare the intensity of mode collapse for 
two extreme cases of such generators.
$Q_1=U([0.2,1])$ is uniform over $[0.2,1]$ and 
$Q_2 = 0.6 U([0,0.5]) + 1.4 U([0.5,1])$ is a mixture of two uniform distributions, 
as shown in Figure \ref{fig:toy}.  
They are designed to have the same total variations distance, i.e.~$d_{\rm TV}(P,Q_1)=d_{\rm TV}(P,Q_2)=0.2$, 
but $Q_1$ exhibits an extreme mode collapse as the whole probability mass in $[0,0.2]$ is lost, whereas 
$Q_2$ captures a more balanced deviation from $P$.  

Definition \ref{def:modecollapse} 
captures the fact that $Q_1$ has more mode collapse than $Q_2$, since  
the pair $(P,Q_1)$ exhibits $(\varepsilon=0,\delta=0.2)$-mode collapse, whereas the pair 
$(P,Q_2)$ exhibits only $(\varepsilon=0.12,\delta=0.2)$-mode collapse, for the same value of $\delta=0.2$. 
However, the appropriate way to precisely represent mode collapse (as we define it) 
is to visualize it through a two-dimensional region we call the {\em mode collapse region}. 
For a given pair $(P,Q)$, 
the corresponding mode collapse region $\cR(P,Q)$ is defined as the convex hull of 
the region of points $(\varepsilon,\delta)$ such that $(P,Q)$ exhibit $(\varepsilon,\delta)$-mode collapse, as shown in Figure \ref{fig:toy}. 
\begin{eqnarray}
	\cR(P,Q) &\triangleq& {\rm conv}\big(\, \big\{ \, (\varepsilon,\delta) \,\big|\,  \delta>\varepsilon \text{ and }(P,Q)\text{ has $(\varepsilon,\delta)$-mode collapse}  \, \big\} \, \big) \;,
	\label{eq:def_region}
\end{eqnarray}
where ${\rm conv}(\cdot)$ denotes the convex hull. 
This definition of region is fundamental in the sense that it is a sufficient statistic that captures 
the relations between $P$ and $Q$ for the purpose of hypothesis testing. 
This assertion is made precise in Section \ref{sec:region} by making a strong connection between the mode collapse region 
and the type I and type II errors in binary hypothesis testing. 
That connection allows us to prove a sharp result on 
how the loss, as seen by the discriminator, evolves under PacGAN in Section \ref{sec:proof}. 
For now, we can use this region representation of a given target-generator pair to 
detect the strength of mode collapse occurring for a given generator. 

Typically, we are interested in the presence of mode collapse with a small $\varepsilon$ and a 
much larger $\delta$;
this corresponds to a sharply-increasing slope near the origin $(0,0)$ in the mode collapse region. 
For example, the middle panel in Figure \ref{fig:toy} depicts the mode collapse region (shaded in gray) for a pair of distributions $(P,Q_1)$ that exhibit significant mode collapse; notice the sharply-increasing slope at $(0,0)$ 
of the upper boundary of the shaded grey region (in this example the slope is in fact infinite). 
The right panel in Figure \ref{fig:toy} illustrates the same region for a pair of distributions $(P,Q_2)$ that do not exhibit strong mode collapse, resulting a region with a much gentler slope at $(0,0)$ 
of the upper boundary of the shaded grey region. 


Similarly, if the generator assigns a large probability mass 
compared to the target distribution on a subset, 
we call it a {\em mode augmentation}, and give a formal definition below. 
\begin{definition}
	\label{def:modeaugmentation}
	A pair of a target distribution $P$ and a generator $Q$ has an $(\varepsilon,\delta)$-{\em mode augmentation} for some $0\leq \varepsilon < \delta\leq 1$ if there exists a set $S\subseteq \cX$ such that $Q(S)\geq \delta$ and $P(S) \leq \varepsilon$.
\end{definition} 

Note that we distinguish mode collapse and augmentation strictly here, for analytical purposes. 
In GAN literature, both collapse and augmentation contribute to the observed 
``mode collapse'' phenomenon, which loosely refers to the lack of diversity in the generated samples.

\subsection{Evolution of the region under product distributions}
\label{sec:productregion}
The toy example generators $Q_1$ and $Q_2$ from Figure \ref{fig:toy} could not be distinguished using only their total variation distances from $P$, despite exhibiting very different mode collapse properties. 
This suggests that the original GAN (with 0-1 loss) may be vulnerable to mode collapse. 
We prove in  Theorem \ref{thm:main2} that a discriminator that packs multiple samples 
together \emph{can} better distinguish mode-collapsing generators. 
Intuitively, $m$ packed samples are equivalent to 
a single sample 
drawn from the 
product distributions $P^m$ and $Q^m$. 
We show in this section that 
there is a fundamental connection between 
the strength of mode collapse of $(P,Q)$ and the loss as seen by the packed discriminator  $d_{\rm TV}(P^m,Q^m)$.

\begin{figure}[ht]
	\begin{center}
	\includegraphics[width=.33\textwidth]{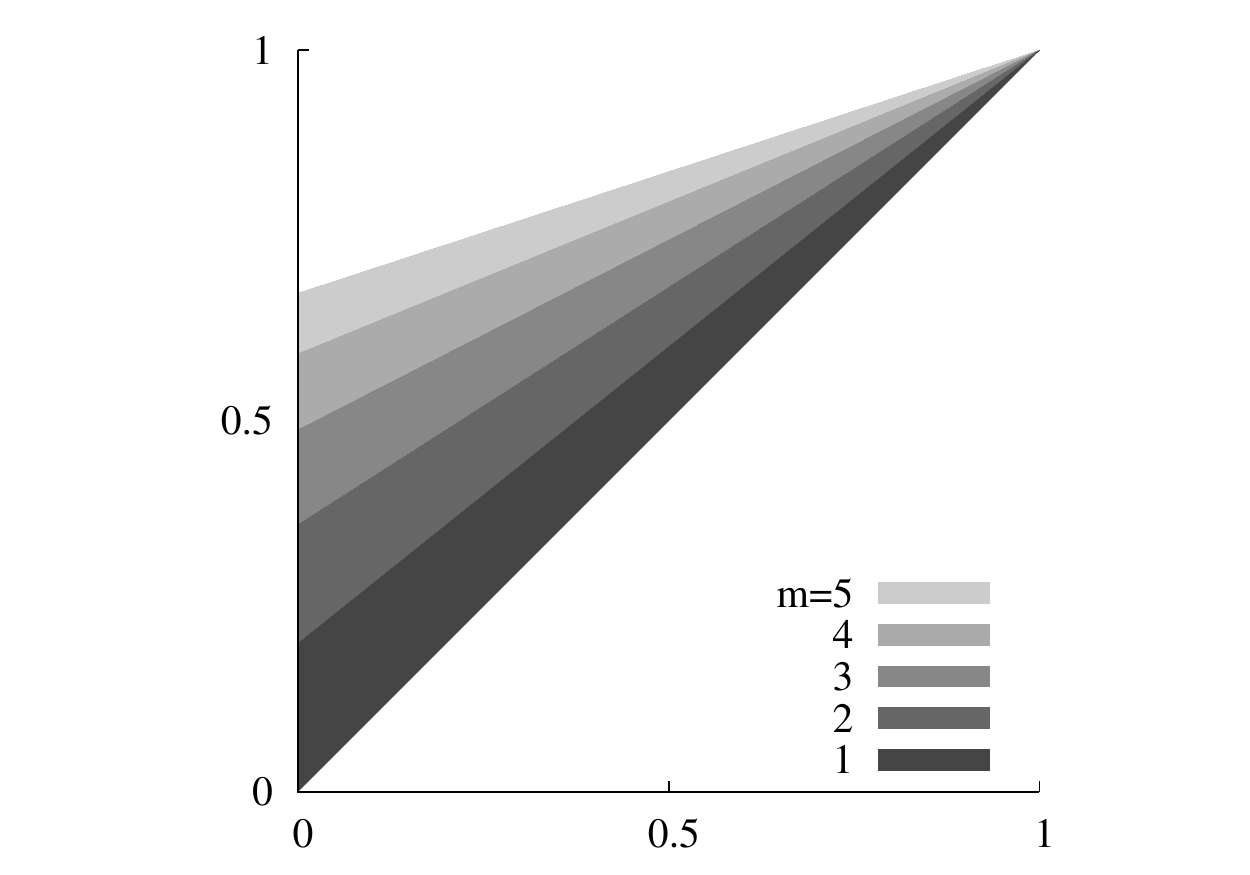}
	\put(-70,50){$\cR(P^m,Q_1^m)$}
	\put(-75,-4){$\varepsilon$}
	\put(-138,60){$\delta$}
	\includegraphics[width=.33\textwidth]{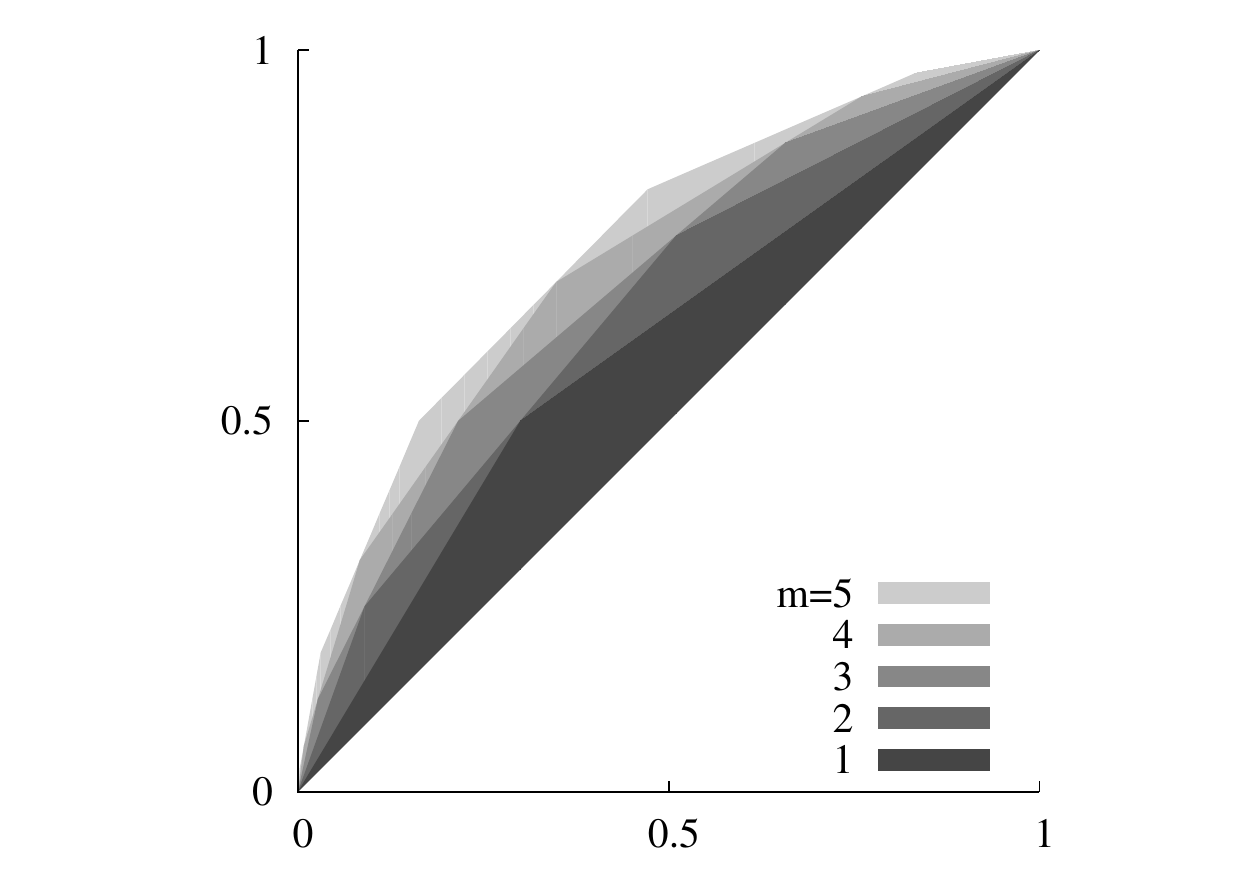}
	\put(-70,50){$\cR(P^m,Q_2^m)$}
	\put(-75,-4){$\varepsilon$}
	\put(-138,60){$\delta$}
	\includegraphics[width=.32\textwidth]{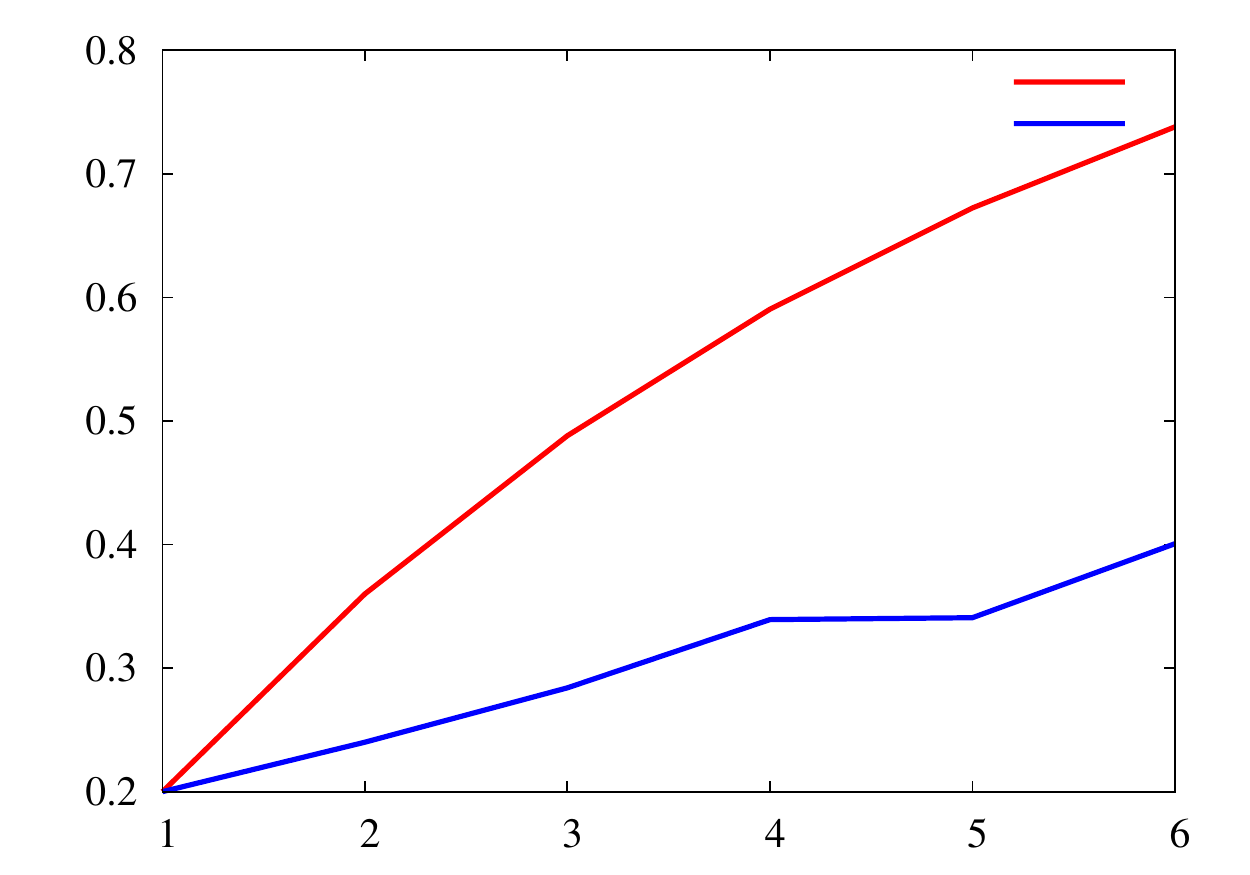}
	\put(-115,-4){degree of packing $m$}
	\put(-160,104){Total variation distance}
	\put(-76,95){\tiny$d_{\rm TV}(P^m,Q_1^m)$}
	\put(-76,88){\tiny$d_{\rm TV}(P^m,Q_2^m)$}
	\end{center}
	\caption{Evolution of the mode collapse region over the degree of packing $m$ for the two toy examples from Figure \ref{fig:toy}. 
		The region of the mode-collapsing 
		generator $Q_1$ expands faster than the non-mode-collapsing generator $Q_2$ when discriminator inputs are packed (at $m=1$ these examples have the same TV distances). This causes a discriminator to penalize mode collapse as desired.  }
	\label{fig:toy_evolve}
\end{figure}

 \bigskip
 \noindent
{\bf Intuition via toy examples.} 
Concretely, consider the  example from the previous section and 
recall that $P^m$ denote the product distribution resulting from  packing  together $m$ 
independent samples from $P$. 
Figure \ref{fig:toy_evolve} illustrates how the mode collapse region evolves over $m$, 
the degree of packing. 
This evolution highlights a key insight:
the region $\cR(P^m,Q_1^m)$ of a mode-collapsing generator 
expands much faster as $m$ increases
compared to 
the region $\cR(P^m,Q_2^m)$ of a non-mode-collapsing generator. 
This implies that the total variation distance of $(P,Q_1)$ increases more rapidly as we pack more samples, compared to 
$(P,Q_2)$.
This follows from the fact that 
the total variation distance between $P$ and the generator can be determined directly from the upper boundary of the mode collapse region (see Section \ref{sec:properties} for the precise relation). 
In particular, a larger mode collapse region  implies a larger total variation distance between $P$ and the generator, which is made precise in Section \ref{sec:properties}. 
The total variation distances $d_{\rm TV}(P^m,Q_1^m)$ and $d_{\rm TV}(P^m,Q_2^m)$, which were explicitly chosen to be equal at $m=1$ in our example, grow farther apart with increasing $m$, as illustrated in the right figure below. 
This implies that if we use a packed discriminator, the mode-collapsing generator $Q_1$ will be 
heavily penalized for having a larger loss, compared to the  non-mode-collapsing $Q_2$. 


\begin{figure}[ht]
	\begin{center}
	\includegraphics[width=.3\textwidth]{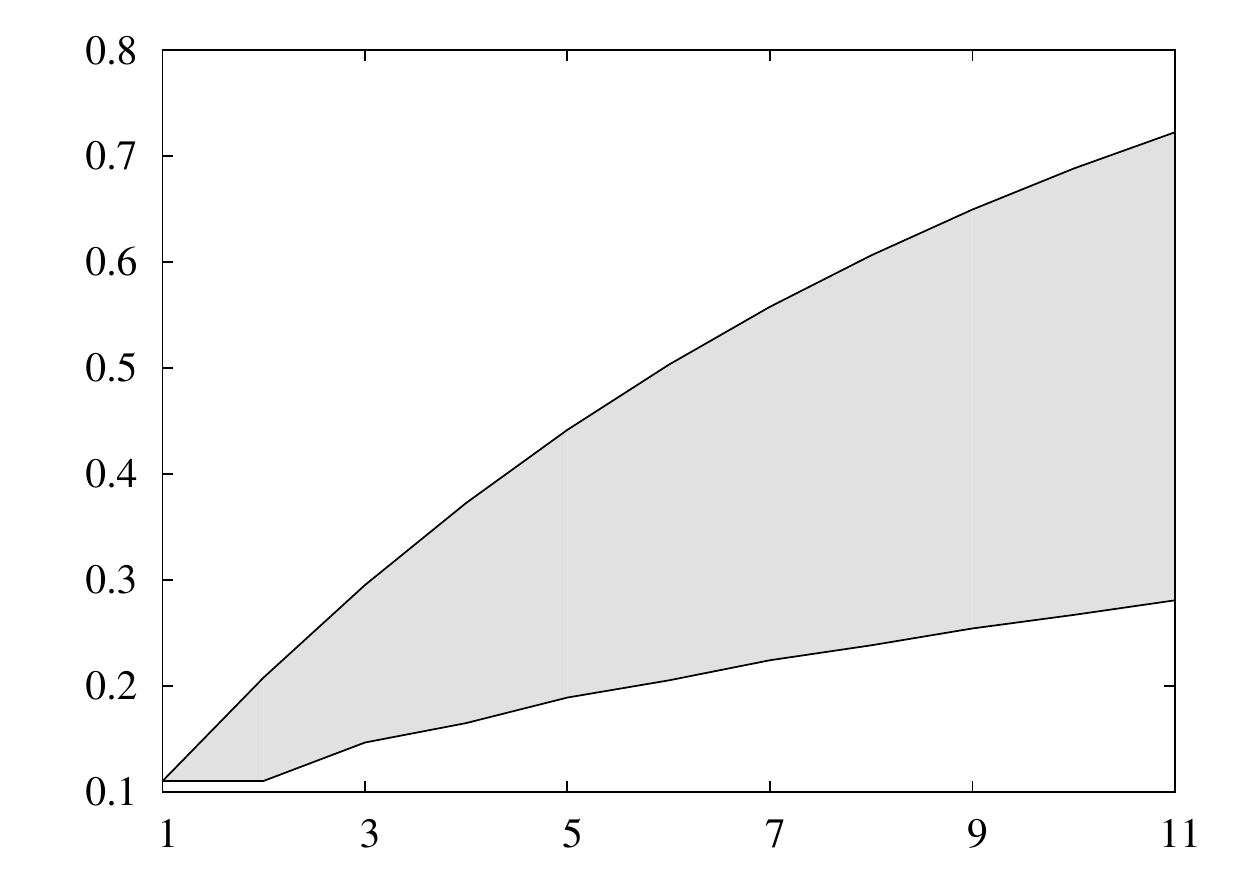}
	\put(-200,65){$d_{\rm TV}(P^m,Q^m)$}
	\put(-115,-7){degree of packing $m$}
	\includegraphics[width=.3\textwidth]{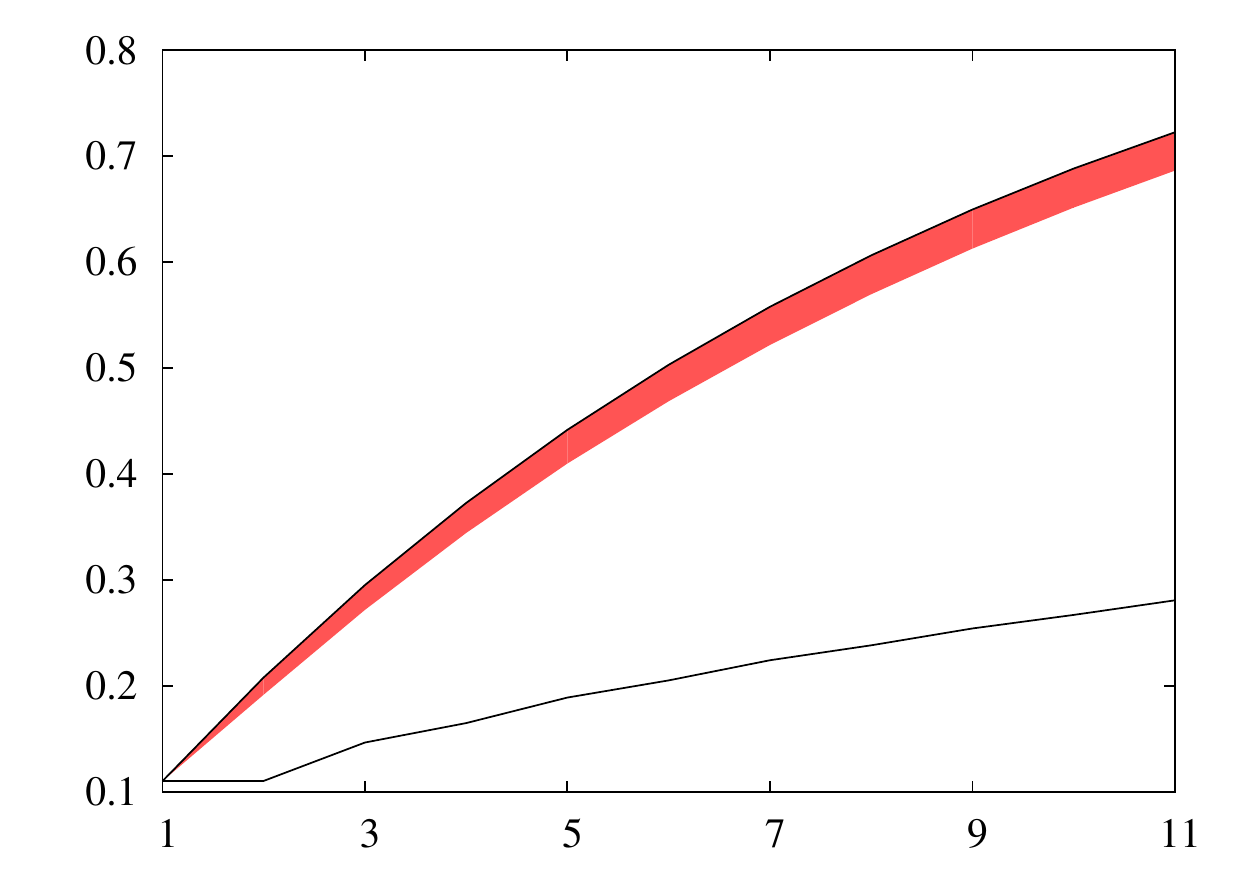}
	\put(-120,100){\small $(0.00,0.1)$-mode collapse}
	\put(-115,-7){degree of packing $m$}
	\includegraphics[width=0.3\textwidth]{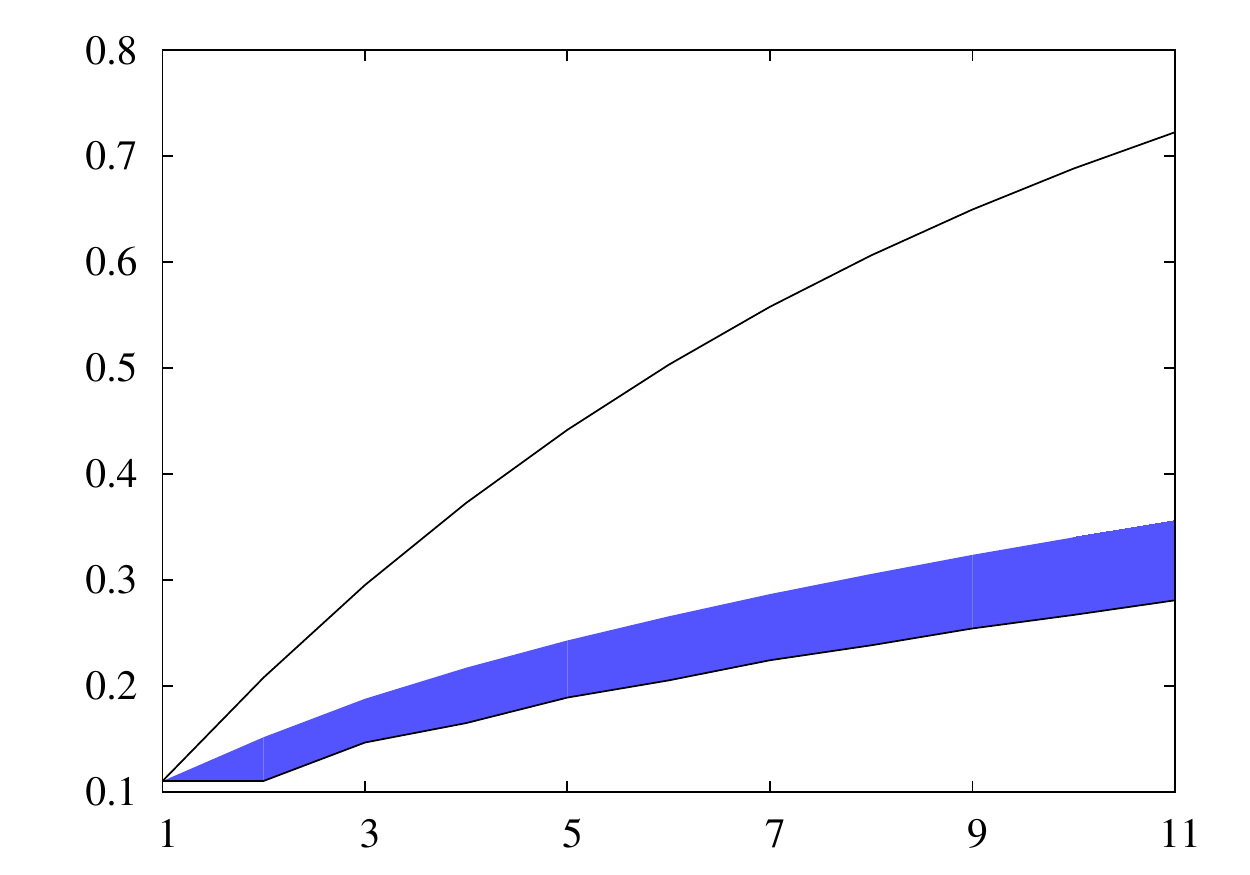}
	\put(-120,100){\small no $(0.07,0.1)$-mode collapse}
	\put(-115,-7){degree of packing $m$}
	\end{center}
	\caption{The range of $d_{\rm TV}(P^m,Q^m)$ achievable by pairs with $d_{\rm TV}(P,Q)=\tau$, for a choice of $\tau=0.11$, defined by the solutions of 
	the optimization \eqref{eq:opt_main1} provided in Theorem \ref{thm:main1} (left panel). 
	The range of $d_{\rm TV}(P^m,Q^m)$ achievable by those pairs that 	
	also have $(\varepsilon=0.00,\delta=0.1)$-mode collapse (middle panel). 
	 A similar range achievable by pairs of distributions that do not have 
	 $(\varepsilon=0.07,\delta=0.1)$-mode collapse or 
	  $(\varepsilon=0.07,\delta=0.1)$-mode augmentation (right panel).
	Pairs $(P,Q)$ with strong mode collapse occupy the top region (near the upper bound) and the pairs 
	with weak mode collapse occupy the bottom region (near the lower bound).
	}
	\label{fig:main1}
\end{figure}

 \bigskip
 \noindent
{\bf Evolution of total variation distances.}
In order to generalize the intuition from the above toy examples, 
we first analyze how the total variation evolves for the set  of all pairs $(P,Q)$ that have the same total variation distance $\tau$ when unpacked (i.e., when $m=1$). 
The solutions to the following optimization problems give the desired upper and lower bounds, respectively, on total variation distance for any distribution pair in this set with a packing degree of $m$: 
\begin{flalign}
	\label{eq:opt_main1}
	&&\min_{P,Q}  ~~d_{\rm TV}(P^m,Q^m) \phantom{a}&&  \max_{P,Q}  ~~d_{\rm TV}(P^m,Q^m) \phantom{aa}&&\\
	&&\text{subject to} ~~~ d_{\rm TV}(P,Q)=\tau && \text{subject to} ~~~ d_{\rm TV}(P,Q)=\tau \;, && \nonumber
\end{flalign}
where the maximization and minimization are over all probability measures $P$ and $Q$. 
We give  the exact solution in Theorem \ref{thm:main1}, which is illustrated pictorially in Figure \ref{fig:main1} (left). 

\begin{thm}
	\label{thm:main1}
	For all $0\leq\tau\leq1$ and a positive integer $m$,
	the solution to the maximization in \eqref{eq:opt_main1} is $1-(1-\tau)^m$, 
	and the solution to the minimization in  \eqref{eq:opt_main1} is 
	\begin{eqnarray}
		L(\tau,m) &  \triangleq &  \min_{ 0 \leq \alpha\leq 1 -\tau }  \; d_{\rm TV} \Big(\, P_{\rm inner}(\alpha)^{ m}, 		Q_{\rm inner}(\alpha,\tau)^{ m} \,\Big)\;, \label{eq:defL}
	\end{eqnarray}
	where $P_{\rm inner}(\alpha)^m$ and $Q_{\rm inner}(\alpha,\tau)^m$ are the $m$-th order product distributions of  binary random variables distributed as 
	\begin{eqnarray}
		P_{\rm inner}(\alpha) &=& \begin{bmatrix}1-\alpha, &\alpha\end{bmatrix} \;, \label{eq:pinner}\\
		Q_{\rm inner}(\alpha,\tau) &=& \begin{bmatrix} 1-\alpha - \tau, & \alpha+\tau \end{bmatrix}\;.\label{eq:qinner}
	\end{eqnarray}
\end{thm}
Although this is a simple statement that can be proved in several different ways, 
we introduce in Section \ref{sec:proof}  a novel geometric proof technique that critically relies on 
the proposed mode collapse region. 
This particular technique will allow us to generalize the proof to more complex problems involving mode collapse in Theorem \ref{thm:main2}, 
for which other techniques do not generalize. 
Note that 
the claim in Theorem \ref{thm:main1} has nothing to do with mode collapse. 
Still,  the mode collapse region definition (used here purely as a proof technique) provides 
a novel technique that seamlessly generalizes to prove more complex statements in the following. 

For any given value of $\tau$ and $m$, 
the bounds in Theorem \ref{thm:main1} are easy to evaluate numerically, as shown below in the left panel of Figure \ref{fig:main1}.   
Within this achievable range, some subset of pairs $(P,Q)$ have rapidly increasing total variation, occupying the upper part of the region (shown in red, middle panel of Figure \ref{fig:main1}), 
and some subset of pairs $(P,Q)$ have slowly increasing total variation, occupying the lower part as shown in blue in the right panel in Figure \ref{fig:main1}. 
In particular, the evolution of the mode-collapse region of a pair of $m$-th power distributions 
	$\cR(P^m,Q^m)$ is fundamentally connected to the  strength of mode collapse in the original pair $(P,Q)$. 
This means that for a mode-collapsed pair $(P,Q_1)$, the $m$th-power distribution will exhibit a different total variation distance evolution than a non-mode-collapsed pair $(P,Q_2)$. 
As such, these two pairs can be distinguished by a packed discriminator. 
	Making such a claim precise for a broad class of mode-collapsing and non-mode-collapsing generators is challenging, 
	as it depends on the target $P$ and the generator $Q$, each of which can be a  complex high dimensional distribution, 
	like natural images.
	The proposed region interpretation, endowed with the hypothesis testing interpretation and the data processing inequalities that come with it, 
	 is critical:
	it enables the abstraction of technical details and 
	provides a  simple and tight proof based on {\em geometric techniques} 
	on two-dimensional regions.

 \bigskip
 \noindent
{\bf Evolution of total variation distances with mode collapse.}
We analyze how the total variation evolves for the set  of all pairs $(P,Q)$ that have the same total variations distances $\tau$ when unpacked, with $m=1$, 
and have $(\varepsilon,\delta)$-mode collapse for some $0\leq\varepsilon< \delta\leq 1$. 
The solution of the following optimization problem gives the desired range of total variation distances: 
\begin{flalign}
	\label{eq:opt_main2}
	\min_{P,Q}  ~~d_{\rm TV}(P^m,Q^m)\phantom{AAAAAAAAA} &&  \max_{P,Q}  ~~d_{\rm TV}(P^m,Q^m) \phantom{AAAAAAAAAAA}\\
	\text{subject to} ~~~ d_{\rm TV}(P,Q)=\tau \phantom{AAAAAAAAA}&& \text{subject to} ~~~ d_{\rm TV}(P,Q)=\tau \phantom{AAAAAAAAAA}\; \nonumber \\
	  \text{$(P,Q)$ has $(\varepsilon,\delta)$-mode collapse} && \text{$(P,Q)$ has $(\varepsilon,\delta)$-mode collapse}\nonumber\;, 
\end{flalign}
where the maximization and minimization are over all probability measures $P$ and $Q$, and 
the mode collapse constraint  is defined in Definition \ref{def:modecollapse}. 
$(\varepsilon,\delta)$-mode collapsing pairs  have total variation at least $\delta-\varepsilon$ by definition, 
and when $\tau<\delta-\varepsilon$, the feasible set of the above optimization is empty.
Otherwise, the next theorem establishes that mode-collapsing pairs occupy the upper part of the total variation region; 
that is, total variation increases rapidly as we pack more samples together (Figure \ref{fig:main1}, middle panel). 
One implication is that distribution pairs $(P,Q)$ at the top of the total variation evolution region are those with the strongest mode collapse. 
Another implication is that a pair $(P,Q)$ with strong mode collapse (i.e., with larger $\delta$ and smaller $\varepsilon$ in the constraint) will 
be penalized more under packing, and hence a generator minimizing an approximation of $d_{\rm TV}(P^m,Q^m)$ 
will be unlikely to 
select a distribution that exhibits such strong mode collapse. 

\begin{figure}[ht]
	\begin{center}
		\includegraphics[width=0.33\textwidth]{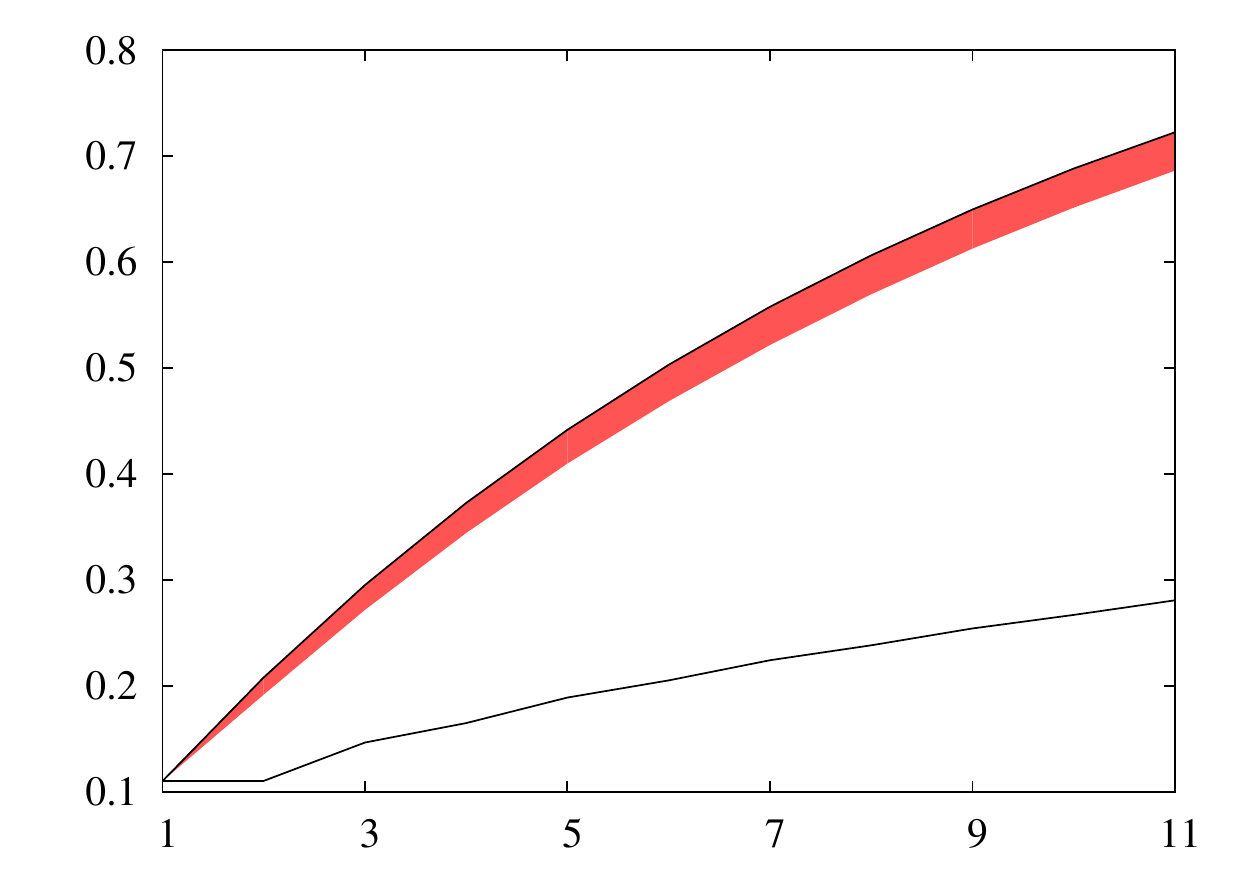}
		\put(-217,64){\small$d_{\rm TV}(P^m,Q^m)$}
		\put(-120,110){\small $(0.00,0.1)$-mode collapse}
		\includegraphics[width=0.33\textwidth]{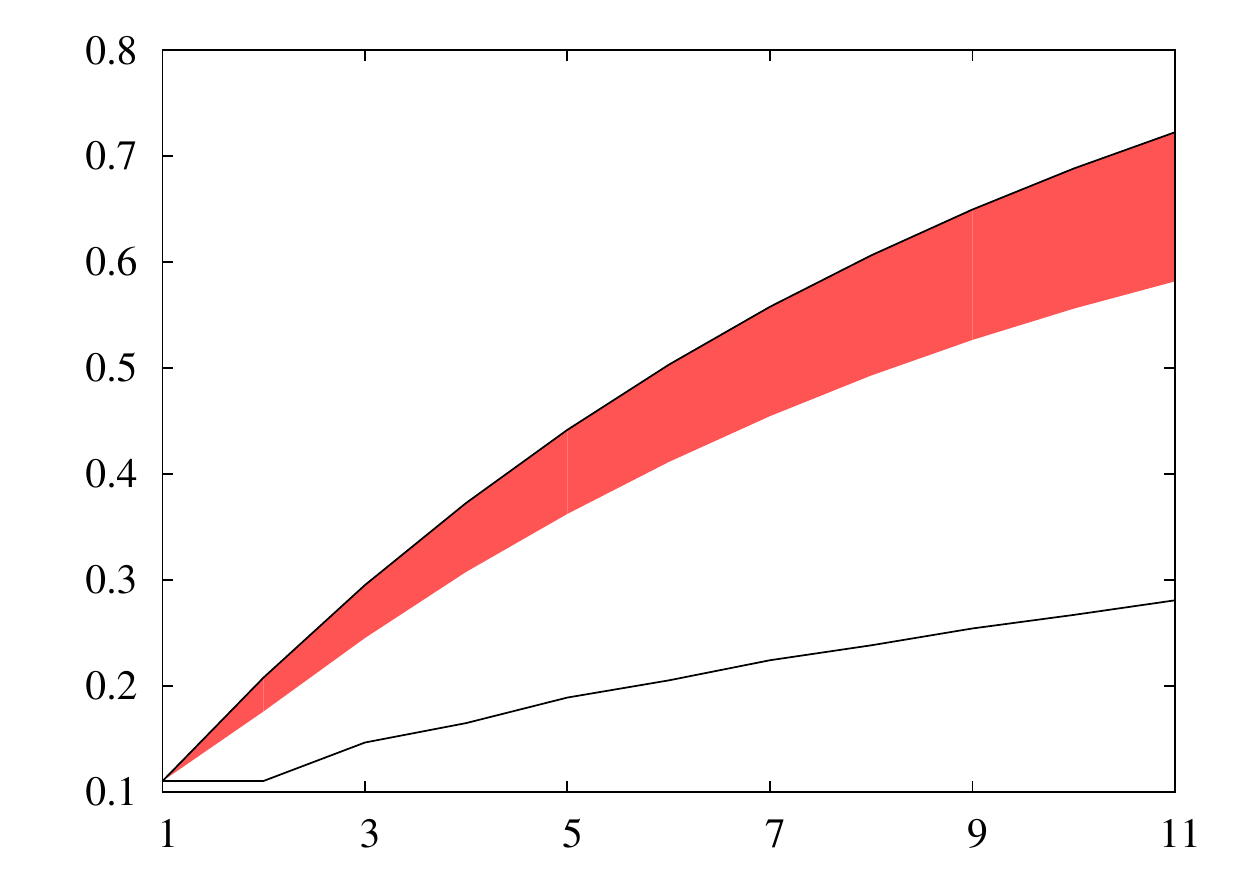}
		\put(-120,110){\small $(0.01,0.1)$-mode collapse}
		\includegraphics[width=0.33\textwidth]{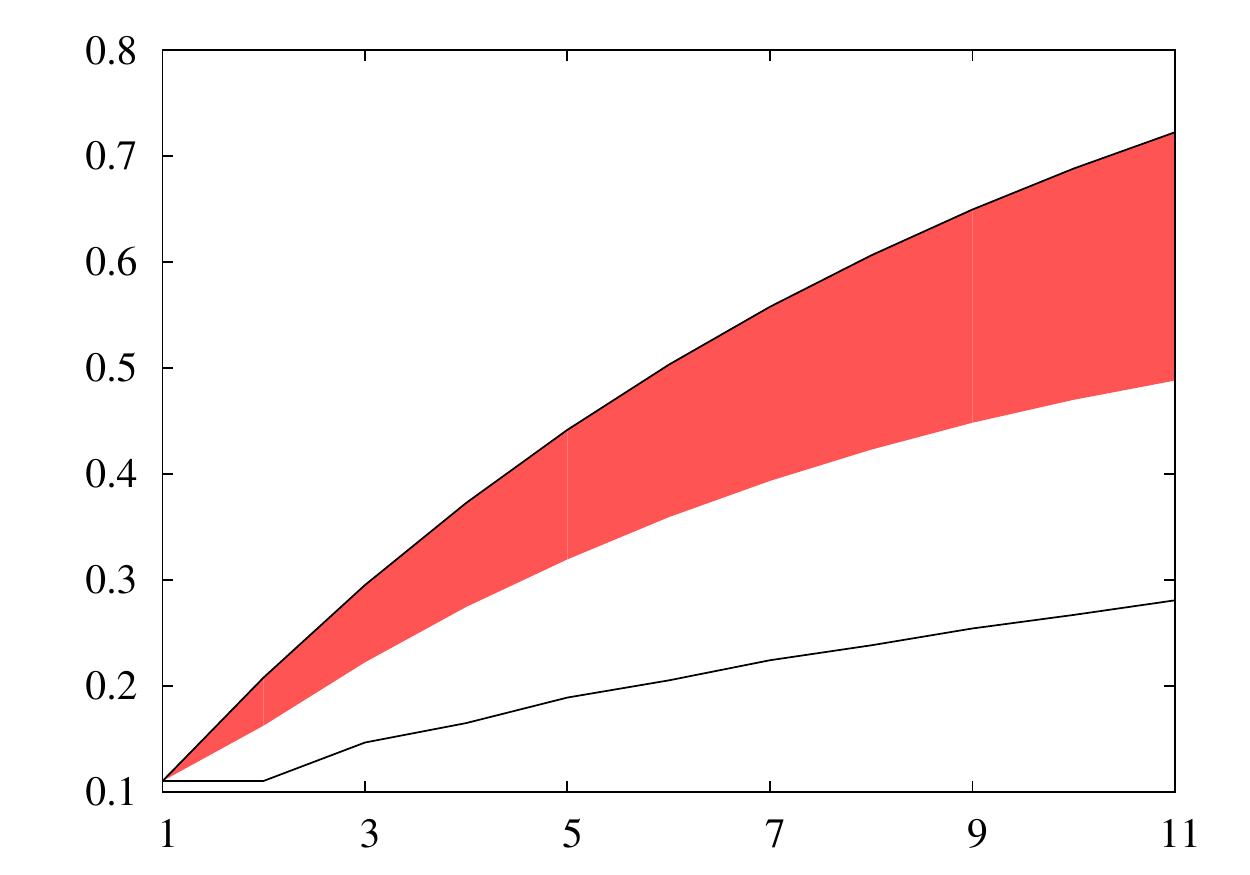}
		\put(-120,110){\small $(0.02,0.1)$-mode collapse}
		\\
		\vspace{0.3cm}
		\includegraphics[width=0.33\textwidth]{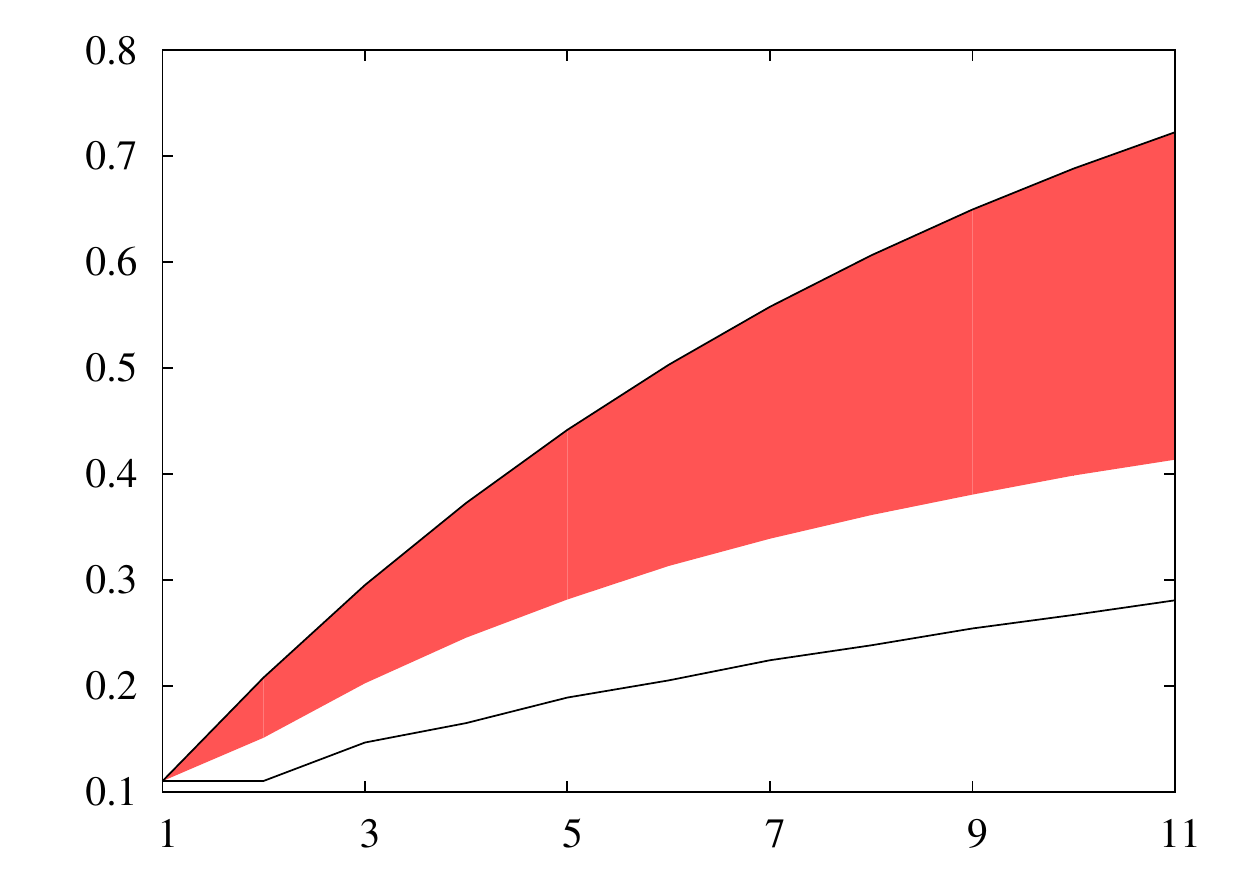}
		\put(-217,64){\small$d_{\rm TV}(P^m,Q^m)$}
		\put(-73,-3){\small $m$}
		\put(-120,110){\small $(0.03,0.1)$-mode collapse}
		\includegraphics[width=0.33\textwidth]{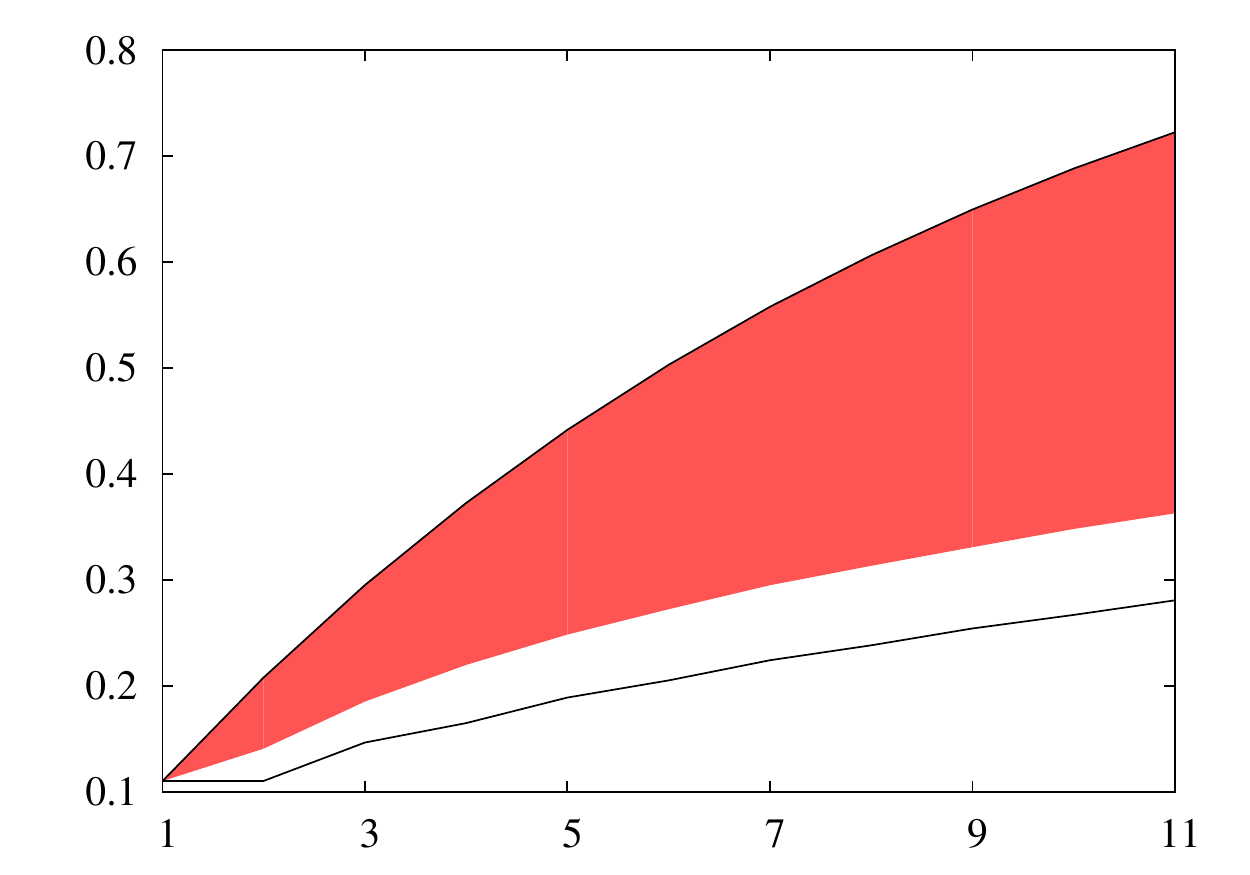}
		\put(-73,-3){\small $m$}
		\put(-120,110){\small $(0.04,0.1)$-mode collapse}
		\includegraphics[width=0.33\textwidth]{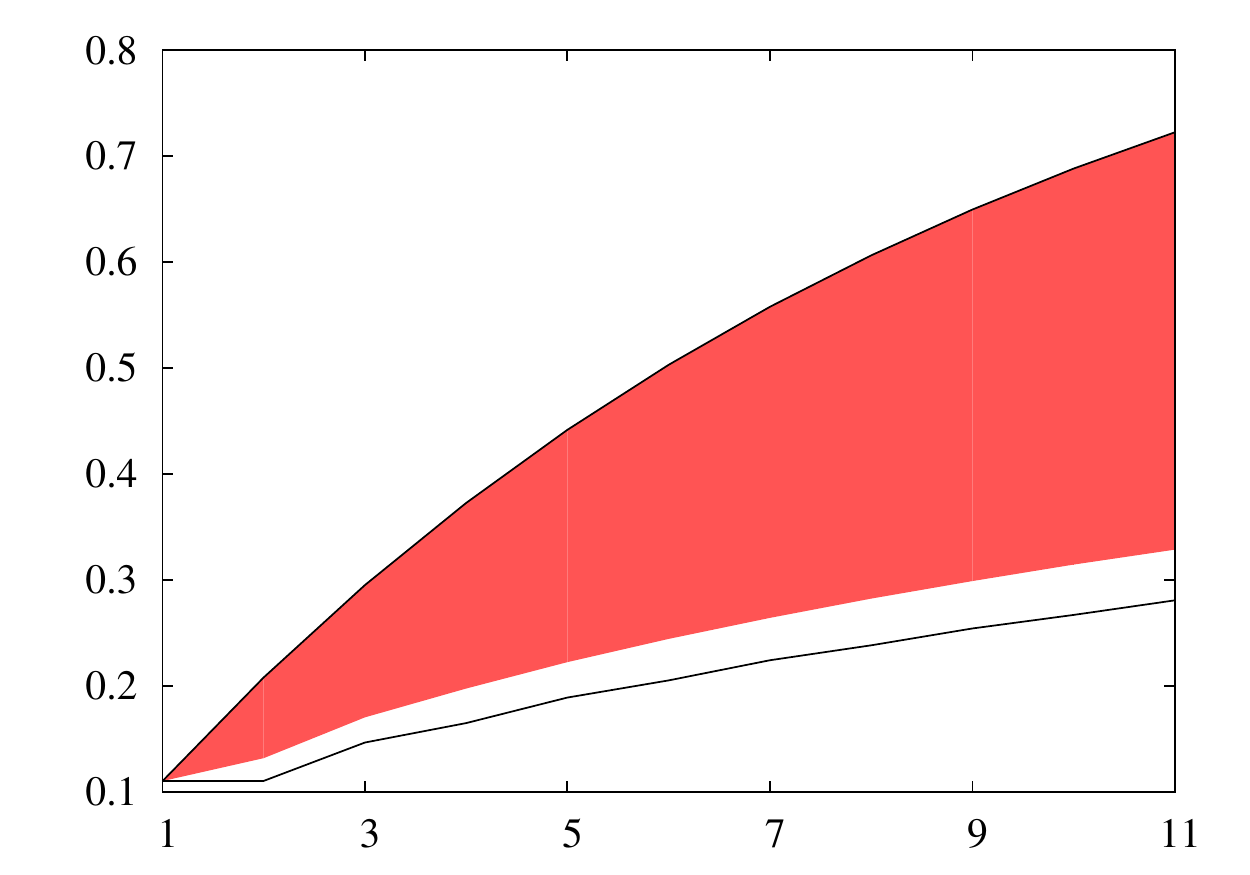}
		\put(-73,-3){\small $m$}
		\put(-120,110){\small $(0.05,0.1)$-mode collapse}
	\end{center}
	\caption{The evolution of total variation distance over the packing degree $m$ 
		for mode collapsing pairs is shown as a red band. 
		The upper and lower boundaries of the red band is 
		defined by the optimization \ref{eq:opt_main2} 
		and computed using Theorem \ref{thm:main2}. 
		For a fixed $d_{\rm TV}(P,Q)=\tau=0.11$ and 
		$(\varepsilon,\delta=0.1)$-mode collapse, 
		we show the evolution with different choices of $\varepsilon \in\{0.00,0.01,0.02,0.03,0.04,0.05\}$. 
		The black   solid lines show the maximum/minimum total variation in the optimization problem \eqref{eq:opt_main1} as a reference. 
		The family of pairs $(P,Q)$ with stronger mode collapse (i.e.~smaller $\varepsilon$ in the constraint), 
		occupy a smaller region at the top with higher total variation under packing, and hence is more penalized when training the generator. 
		}
	\label{fig:power2}
\end{figure}

\begin{thm}
	\label{thm:main2} 
	For all $0\leq\varepsilon<\delta\leq1$ and a positive integer $m$, 
	if $1\geq \tau \geq \delta - \varepsilon$
	then the solution to the maximization in \eqref{eq:opt_main2} is $1-(1-\tau)^m$, 
	and the solution to the minimization in  \eqref{eq:opt_main2} is 
	\begin{eqnarray}
		L_1(\varepsilon,\delta,\tau,m) &  \triangleq &  \min\Big\{\, \min_{ 0 \leq \alpha \leq 1-\frac{\tau\delta}{\delta-\varepsilon} }  \; d_{\rm TV} \Big(\, P_{\rm inner1}(\delta,\alpha)^{ m}, 		Q_{\rm inner1}(\varepsilon,\alpha,\tau)^{ m} \,\Big) \,,\, \nonumber\\
		&& \hspace{1.2cm}  \min_{1-\frac{\tau\delta}{\delta-\varepsilon}\leq  \alpha \leq 1-\tau } \; d_{\rm TV} \Big(\, P_{\rm inner2}(\alpha)^{ m}, 	Q_{\rm inner2}(\alpha,\tau)^{ m} \,\Big)\,\Big\}\;, \label{eq:defL1}
	\end{eqnarray}
	where $P_{\rm inner1}(\delta,\alpha)^m$, $Q_{\rm inner1}(\varepsilon,\alpha,\tau)^m$, $P_{\rm inner2}(\alpha)^m$, and $Q_{\rm inner2}(\alpha,\tau)^m$ 
	 are the $m$-th order product distributions of  discrete random variables distributed as 
	 \begin{eqnarray}
	P_{\rm inner1}(\delta,\alpha) &=& \begin{bmatrix}\delta, &1-\alpha-\delta,&\alpha\end{bmatrix}\;,\label{eq:pinner1}\\ 
	Q_{\rm inner1}(\varepsilon,\alpha,\tau) & =& \begin{bmatrix} \varepsilon, & 1-\alpha-\tau-\varepsilon,&\alpha+\tau \end{bmatrix}\;,\label{eq:qinner1}\\ 
	P_{\rm inner2}(\alpha) &=&  \begin{bmatrix}1-\alpha, &\alpha\end{bmatrix} \;, \label{eq:pinner2}\\ 
	Q_{\rm inner2}(\alpha,\tau) &=&  \begin{bmatrix} 1-\alpha-\tau , & \alpha+\tau \end{bmatrix}\;. \label{eq:qinner2}
	\end{eqnarray}
	If $\tau < \delta - \varepsilon$, then the optimization in \eqref{eq:opt_main2} has no solution and the feasible set is an empty set. 
\end{thm}
A proof of this theorem is provided in Section \ref{sec:proof2}, which critically relies on 
the proposed mode collapse region representation of the pair $(P,Q)$, and the celebrated result by Blackwell from \cite{Bla53}. 
The solutions in Theorem \ref{thm:main2} can be numerically evaluated 
for any given choices of $(\varepsilon,\delta,\tau)$ as we
show  in Figure \ref{fig:power2}. 

Analogous results to the above theorem can be shown for pairs $(P,Q)$ that exhibit $(\epsilon,\delta)$ mode augmentation (as opposed to mode collapse). 
These results are omitted for brevity, but the results and analysis are straightforward extensions of the proofs for mode collapse. 
This holds because total variation distance is a metric, and therefore symmetric. 


 \bigskip
 \noindent
{\bf Evolution of total variation distances without mode collapse.}
We next analyze how the total variation evolves for the set  of all pairs $(P,Q)$ that have the same total variations distances $\tau$ when unpacked, with $m=1$, 
and {\em do not} have $(\varepsilon,\delta)$-mode collapse for some $0\leq\varepsilon< \delta\leq 1$. 
Because of the symmetry of the total variation distance, 
mode augmentation in Definition \ref{def:modeaugmentation} 
is equally damaging as mode collapse, when it comes to how fast total variation distances evolve. 
Hence, we characterize this evolution for those family of pairs of distributions that do not have either mode collapse or augmentation. 
The solution of the following optimization problem gives the desired range of total variation distances: 
\begin{flalign}
	\label{eq:opt_main3}
	\min_{P,Q}  ~~d_{\rm TV}(P^m,Q^m) \phantom{aAAAAAAAAAA}&&  \max_{P,Q}  ~~d_{\rm TV}(P^m,Q^m) \phantom{aAAAAAAAAAA} \\
	\text{subject to} ~~~ d_{\rm TV}(P,Q)=\tau \phantom{AAAAAAAAAA}&& \text{subject to} ~~~ d_{\rm TV}(P,Q)=\tau  \phantom{AAAAAAAAAA} \; \nonumber \\
	 \text{$(P,Q)$ does not have $(\varepsilon,\delta)$-mode} && \text{$(P,Q)$ does not have $(\varepsilon,\delta)$-mode }\nonumber\; \\
	 ~~\text{ collapse or augmentation} && ~~\text{ collapse or augmentation} \nonumber\;,  
\end{flalign}
where the maximization and minimization are over all probability measures $P$ and $Q$, and 
the mode collapse and augmentation constraints  are defined in Definitions \ref{def:modecollapse} and \ref{def:modeaugmentation}, respectively. 

It is not possible to have $d_{\rm TV}(P,Q) > (\delta-\varepsilon)/(\delta+\varepsilon)$ and $\delta+\varepsilon\leq 1$, 
and satisfy the mode collapse and mode augmentation constraints  (see Section \ref{sec:proof3} for a proof). 
Similarly, it is not possible to have 
$d_{\rm TV}(P,Q) > (\delta-\varepsilon)/(2-\delta-\varepsilon)$ and $\delta+\varepsilon\geq 1$, 
and satisfy the constraints. 
Hence, the feasible set is empty when $\tau > \max\{(\delta - \varepsilon)/(\delta+\varepsilon),(\delta-\varepsilon)/(2-\delta-\varepsilon)\}$. 
On the other hand, when $\tau\leq\delta-\varepsilon$, 
no pairs with total variation distance $\tau$ can have $(\varepsilon,\delta)$-mode collapse. 
In this case, the optimization reduces to the simpler one in \eqref{eq:opt_main1} with no mode collapse constraints. 
Non-trivial solution exists in the middle regime, i.e.~$ \delta -\varepsilon \leq \tau \leq \max\{(\delta -\varepsilon) / (\delta+\varepsilon),(\delta -\varepsilon)/(2-\delta-\varepsilon)\} $. 
The lower bound for this regime, given in equation \eqref{eq:defL2}, is the same as the lower bound in \eqref{eq:defL}, except it optimizes over a different range of $\alpha$ values. For a wide range of parameters $\varepsilon$, $\delta$, and $\tau$, 
those lower bounds will be the same, and even if they differ for some parameters, they differ slightly. 
This implies that the pairs $(P,Q)$ with weak mode collapse will occupy the bottom part of the evolution of 
the total variation distances  (see Figure \ref{fig:main1} right panel), 
and also will be penalized less under packing. 
Hence a generator minimizing (approximate) $d_{\rm TV}(P^m,Q^m)$ is likely to 
generate distributions with  weak mode collapse. 

\begin{figure}[ht]
	\begin{center}
		\includegraphics[width=0.33\textwidth]{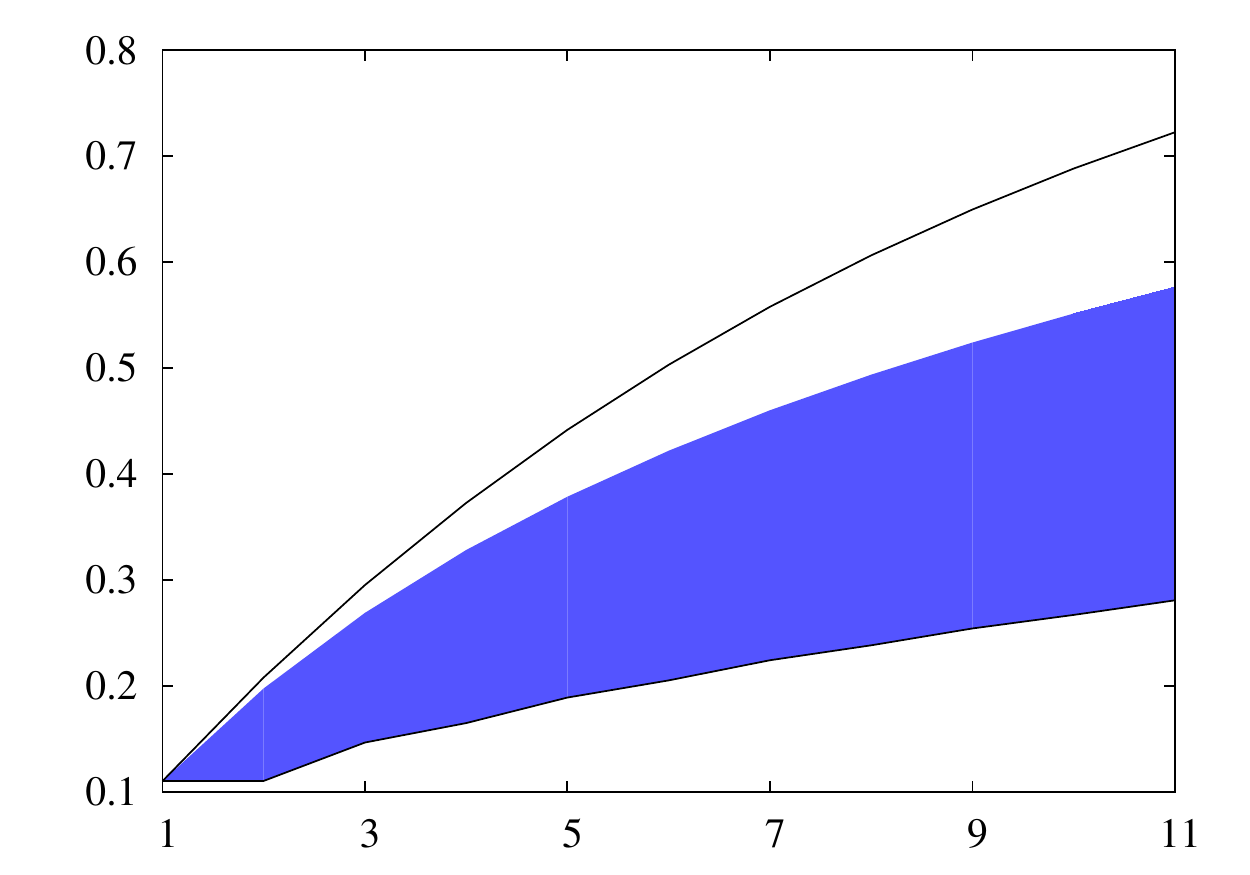}
		\put(-20,130){No $(\epsilon,\delta)$ mode collapse or augmentation}
		\put(-217,64){\small$d_{\rm TV}(P^m,Q^m)$}
		\put(-115,110){\small $(\epsilon,\delta)=(0.03,0.1)$}
		\includegraphics[width=0.33\textwidth]{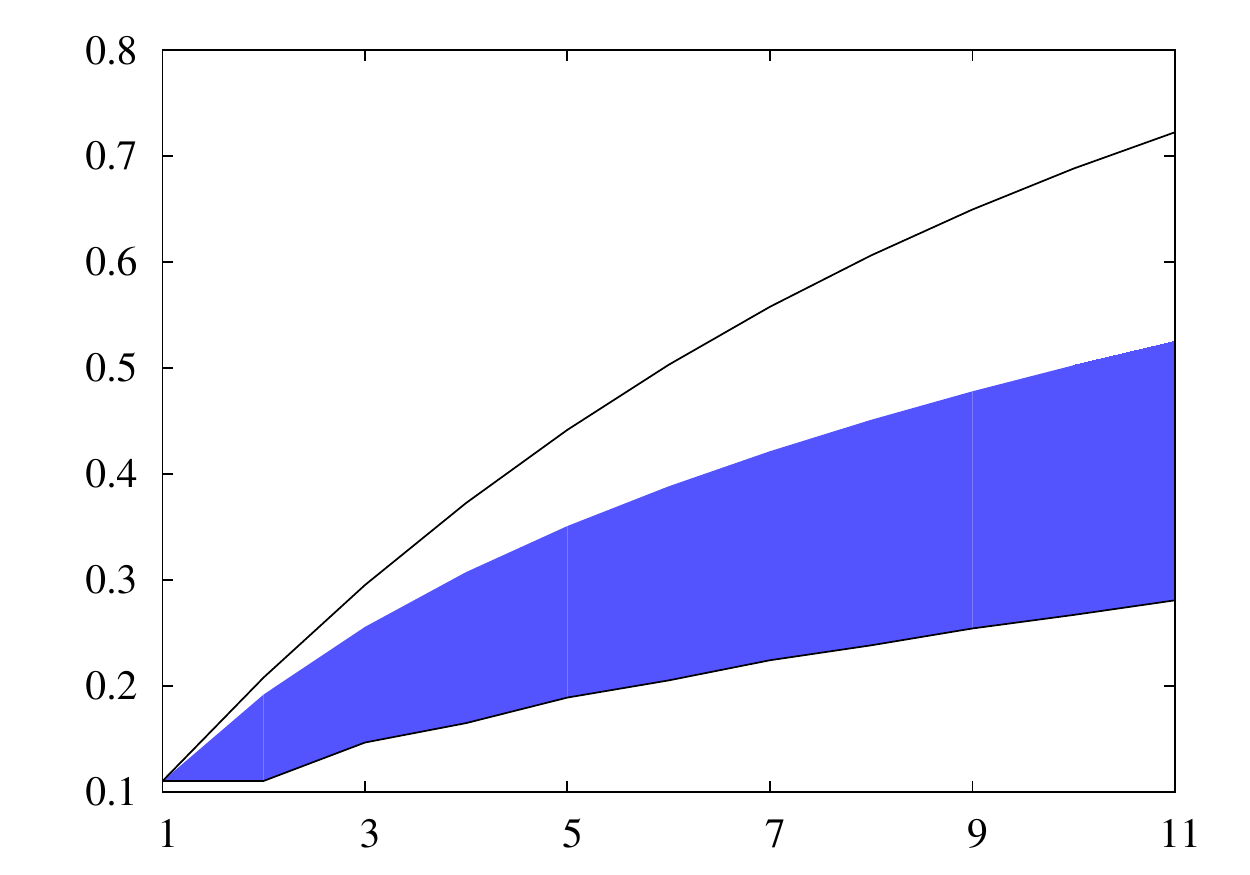}
		\put(-115,110){\small $(\epsilon,\delta)=(0.04,0.1)$}
		\includegraphics[width=0.33\textwidth]{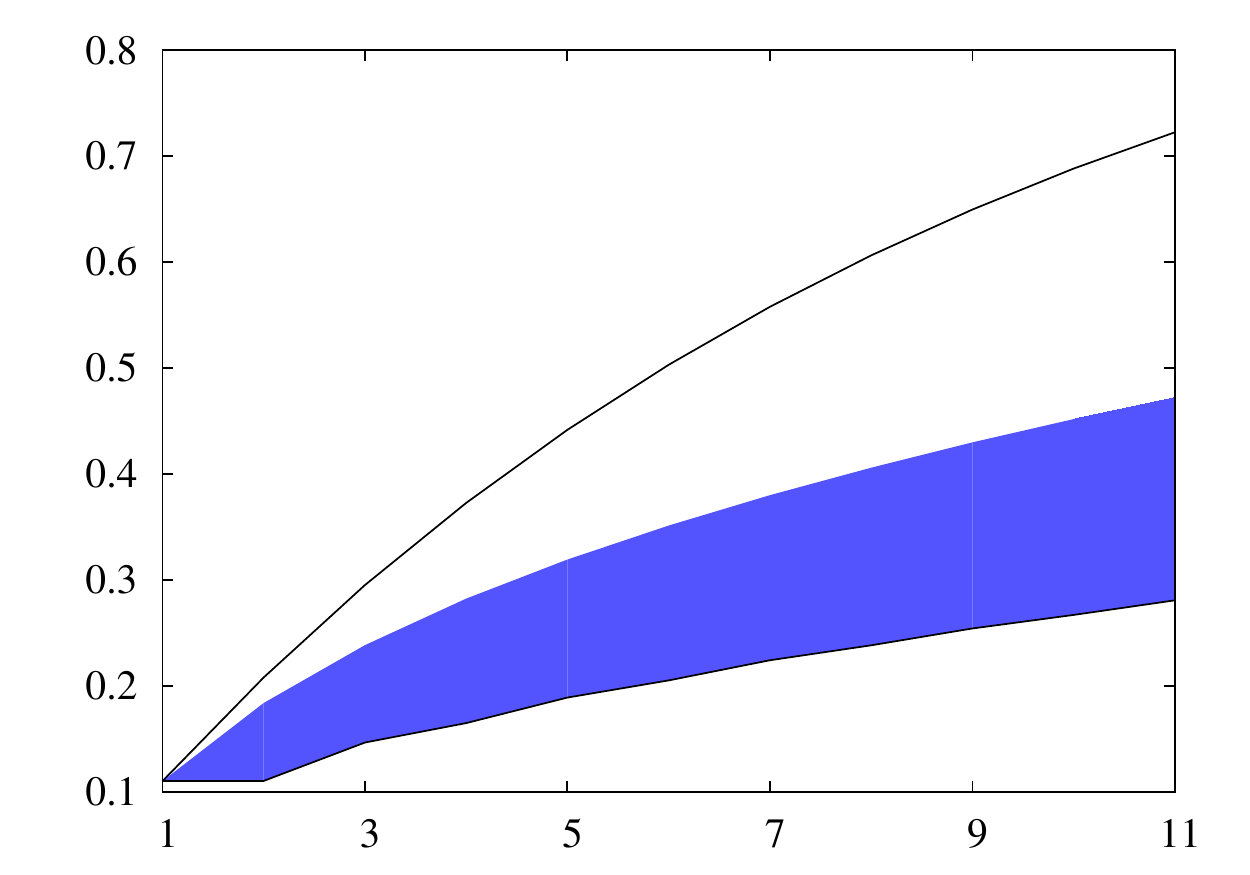}
		\put(-115,110){\small $(\epsilon,\delta)=(0.05,0.1)$}
		\\
		\vspace{0.3cm}
		\includegraphics[width=0.33\textwidth]{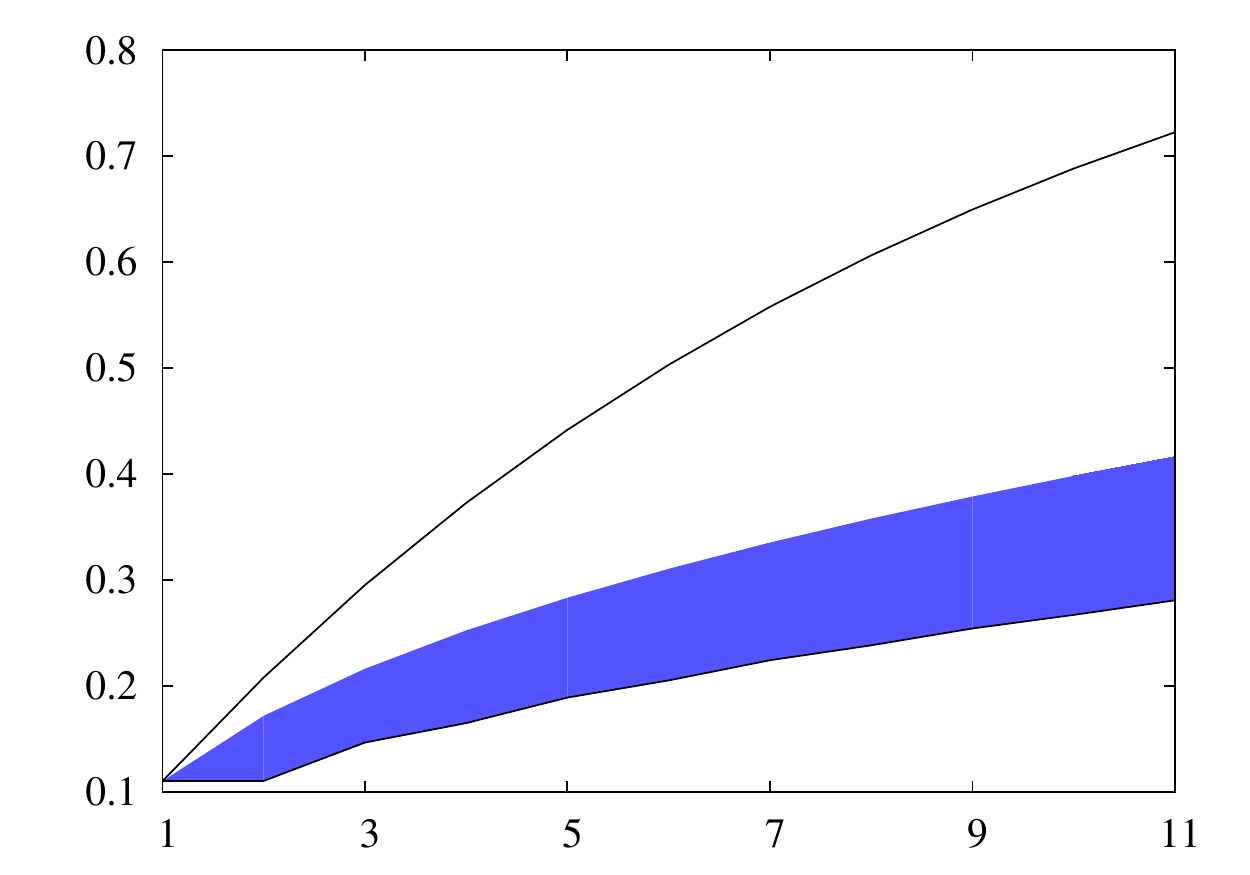}
		\put(-217,64){\small$d_{\rm TV}(P^m,Q^m)$}
		\put(-115,110){\small $(\epsilon,\delta)=(0.06,0.1)$}
		\put(-73,-3){\small $m$}
		\includegraphics[width=0.33\textwidth]{graphs/power_e2_007_white}
		\put(-115,110){\small $(\epsilon,\delta)=(0.07,0.1)$}
		\put(-73,-3){\small $m$}
		\includegraphics[width=0.33\textwidth]{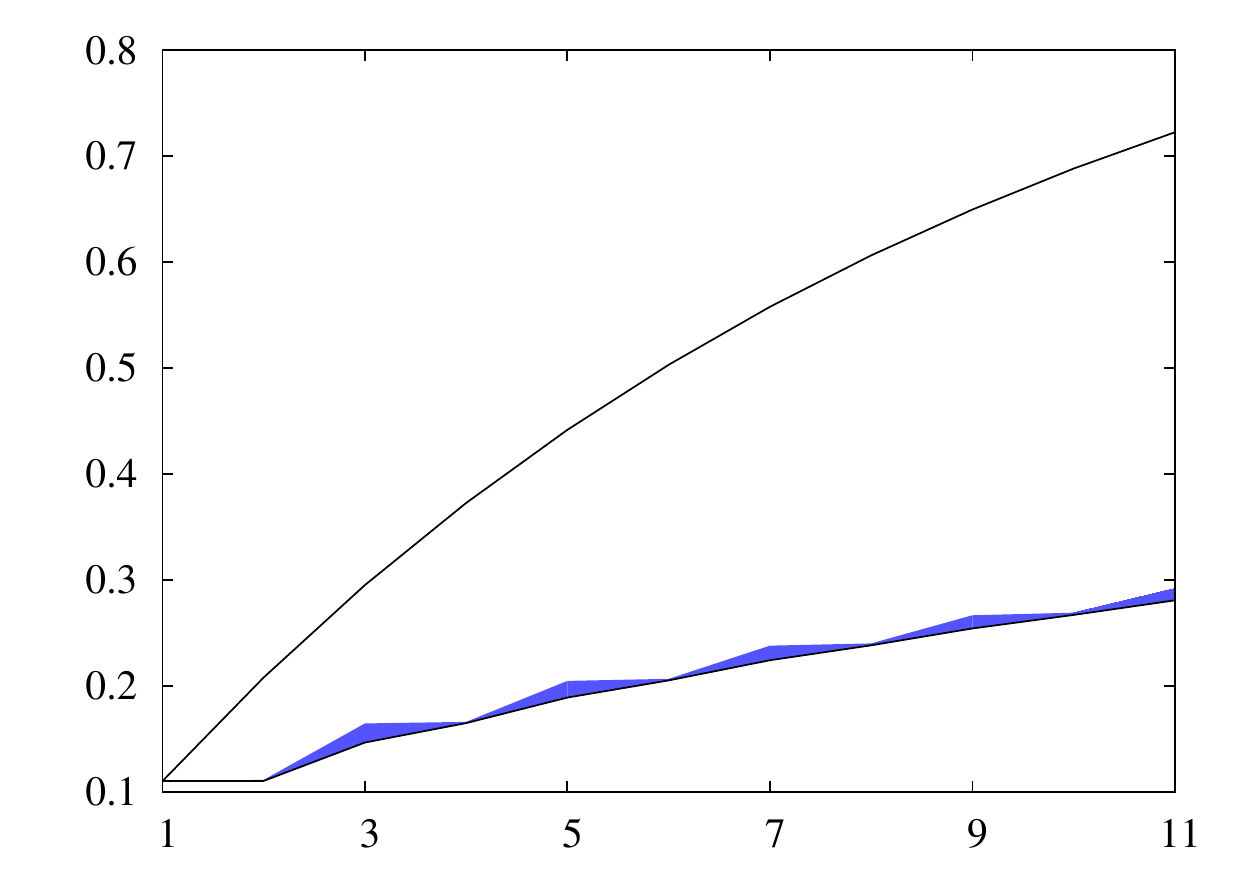}
		\put(-115,110){\small $(\epsilon,\delta)=(0.08,0.1)$}
		\put(-73,-3){\small $m$}
	\end{center}
	\caption{The evolution of total variation distance over the packing degree $m$ 
		for  pairs with no mode collapse/augmentation is shown as a blue band, 
		as defined by the optimization \eqref{eq:opt_main3} 
		and computed using Theorem \ref{thm:main3}. 
		For a fixed $d_{\rm TV}(P,Q)=\tau=0.11$ and 
		the lack of $(\varepsilon,\delta=0.1)$-mode collapse/augmentation constraints, 
		we show the evolution with different choices of $\varepsilon \in\{0.03,0.04,0.05,0.06,0.07,0.08\}$. 
		The black solid lines show the maximum/minimum total variation in the optimization \eqref{eq:opt_main1} as a reference. 
		The family of pairs $(P,Q)$ with weaker mode collapse (i.e.~larger $\varepsilon$ in the constraint), 
		occupies a smaller region at the bottom with smaller total variation under packing, and hence is less penalized when training the generator. 
	}
	\label{fig:power3}
\end{figure}

\begin{thm}
	\label{thm:main3} 
	For all $0\leq\varepsilon<\delta\leq1$ and a positive integer $m$, 
	if $0\leq \tau < \delta-\varepsilon$, then the maximum and the minimum of \eqref{eq:opt_main3}
	are the same as those of the optimization \eqref{eq:opt_main1} provided in Theorem \ref{thm:main1}. 

	If $\delta+\varepsilon\leq1$ and $ \delta -\varepsilon \leq \tau \leq (\delta -\varepsilon) / (\delta+\varepsilon) $
	then the solution to the maximization in \eqref{eq:opt_main3} is 
	\begin{eqnarray}
		U_1({\epsilon,\delta,\tau,m})   &\triangleq &  
		\max_{  \alpha + \beta \leq 1-\tau, \frac{\varepsilon \tau}{\delta-\varepsilon}  \leq \alpha,\beta }  \; \;
		d_{\rm TV} \Big(\, P_{\rm outer1}(\varepsilon,\delta,\alpha,\beta,\tau)^{ m}, 	Q_{\rm outer1}(\varepsilon,\delta,\alpha,\beta,\tau)^{ m} \,\Big)\;, 
		\label{eq:defU1}
	\end{eqnarray} 
	where $P_{\rm outer1}(\varepsilon,\delta,\alpha,\beta,\tau)^m$ and $Q_{\rm outer1}(\varepsilon,\delta,\alpha,\beta,\tau)^m$ are the $m$-th order product distributions of  
	discrete random variables distributed as 
	\begin{eqnarray}
		P_{\rm outer1}(\varepsilon,\delta,\alpha,\beta,\tau) &=& 
		\begin{bmatrix} \frac{\alpha(\delta-\varepsilon)-\varepsilon\tau}{\alpha-\varepsilon},&\frac{\alpha(\alpha+\tau-\delta)}{\alpha-\varepsilon},&
			1-\tau-\alpha-\beta,&\beta,&0 \end{bmatrix} \text{, and} \label{eq:pouter1}\\
		Q_{\rm outer1}(\varepsilon,\delta,\alpha,\beta,\tau) &=& 
		\begin{bmatrix} 0,& \alpha, & 1-\tau-\alpha-\beta,& 
		\frac{\beta(\beta+\tau-\delta)}{\beta-\varepsilon},& \frac{\beta(\delta-\varepsilon)-\varepsilon\tau}{\beta-\varepsilon} \end{bmatrix}\;. \label{eq:qouter1}
	\end{eqnarray}
	The solution to the minimization in  \eqref{eq:opt_main3} is 
	\begin{eqnarray}
		L_2(\tau,m) &  \triangleq &  \min_{ \frac{\varepsilon \tau}{\delta-\varepsilon} \leq \alpha\leq 1-\frac{\delta \tau}{\delta-\varepsilon}  }  \; d_{\rm TV} \Big(\, P_{\rm inner}(\alpha)^{ m}, 		Q_{\rm inner}(\alpha,\tau)^{ m} \,\Big)\;, 
		\label{eq:defL2}
	\end{eqnarray}
	where $P_{\rm inner}(\alpha)$ and $Q_{\rm inner}(\alpha,\tau)$ are defined as in Theorem \ref{thm:main1}. 
	
	If $\delta+\varepsilon>1$ and $ \delta -\varepsilon \leq \tau \leq (\delta -\varepsilon) / (2-\delta-\varepsilon) $
	then the solution to the maximization in \eqref{eq:opt_main3} is 
	\begin{eqnarray}
		U_2({\epsilon,\delta,\tau,m})   &\triangleq &  
		\max_{  \alpha + \beta \leq 1-\tau, \frac{(1-\delta) \tau}{\delta-\varepsilon}  \leq \alpha,\beta }  \; \;
		d_{\rm TV} \Big(\, P_{\rm outer2}(\varepsilon,\delta,\alpha,\beta,\tau)^{ m}, 	Q_{\rm outer2}(\varepsilon,\delta,\alpha,\beta,\tau)^{ m} \,\Big)\;, 
		\label{eq:defU2}
	\end{eqnarray} 
	where $P_{\rm outer2}(\varepsilon,\delta,\alpha,\beta,\tau)^m$ and $Q_{\rm outer2}(\varepsilon,\delta,\alpha,\beta,\tau)^m$ are the $m$-th order product distributions of  
	discrete random variables distributed as 
	\begin{eqnarray}
		P_{\rm outer2}(\varepsilon,\delta,\alpha,\beta,\tau) &=& 
		\begin{bmatrix} \frac{\alpha(\delta-\varepsilon)-(1-\delta) \tau}{\alpha-(1-\delta)},&\frac{\alpha(\alpha+\tau-(1-\varepsilon))}{\alpha-(1-\delta)},&
			1-\tau-\alpha-\beta,&\beta,&0 \end{bmatrix} \text{, and} \label{eq:pouter2}\\
		Q_{\rm outer2}(\varepsilon,\delta,\alpha,\beta,\tau) &=& 
		\begin{bmatrix} 0,& \alpha, & 1-\tau-\alpha-\beta,& 
		\frac{\beta(\beta+\tau-(1-\varepsilon))}{\beta-(1-\delta)},& \frac{\beta(\delta-\varepsilon)-(1-\delta)\tau}{\beta-(1-\delta)} \end{bmatrix}\;. 
		\label{eq:qouter2}
	\end{eqnarray}
	The solution to the minimization in  \eqref{eq:opt_main3} is 
	\begin{eqnarray}
		L_3(\tau,m) &  \triangleq &  
		\min_{  \frac{(1-\delta) \tau}{\delta-\varepsilon}  \leq \alpha \leq1-\frac{(1-\varepsilon) \tau}{\delta-\varepsilon}   }  
		\; d_{\rm TV} \Big(\, P_{\rm inner}(\alpha)^{ m}, 		Q_{\rm inner}(\alpha,\tau)^{ m} \,\Big)\;, 
		\label{eq:defL3}
	\end{eqnarray}
	where $P_{\rm inner}(\alpha)$ and $Q_{\rm inner}(\alpha,\tau)$ are defined as in Theorem \ref{thm:main1}.

	If $\tau > \max\{ (\delta - \varepsilon)/(\delta+\varepsilon), (\delta-\varepsilon)/(2-\delta-\varepsilon)\}$, then the optimization in \eqref{eq:opt_main3} has no solution and the feasible set is an empty set. 
\end{thm}
A proof of this theorem is provided in Section \ref{sec:proof3}, which also critically relies on 
the proposed mode collapse region representation of the pair $(P,Q)$ and the celebrated result by Blackwell from \cite{Bla53}. 
The solutions in Theorem \ref{thm:main3} can be numerically evaluated 
for any given choices of $(\varepsilon,\delta,\tau)$ as we
show  in Figure \ref{fig:power3}.

\begin{figure}[h]
	\begin{center}
		\includegraphics[width=0.6\textwidth]{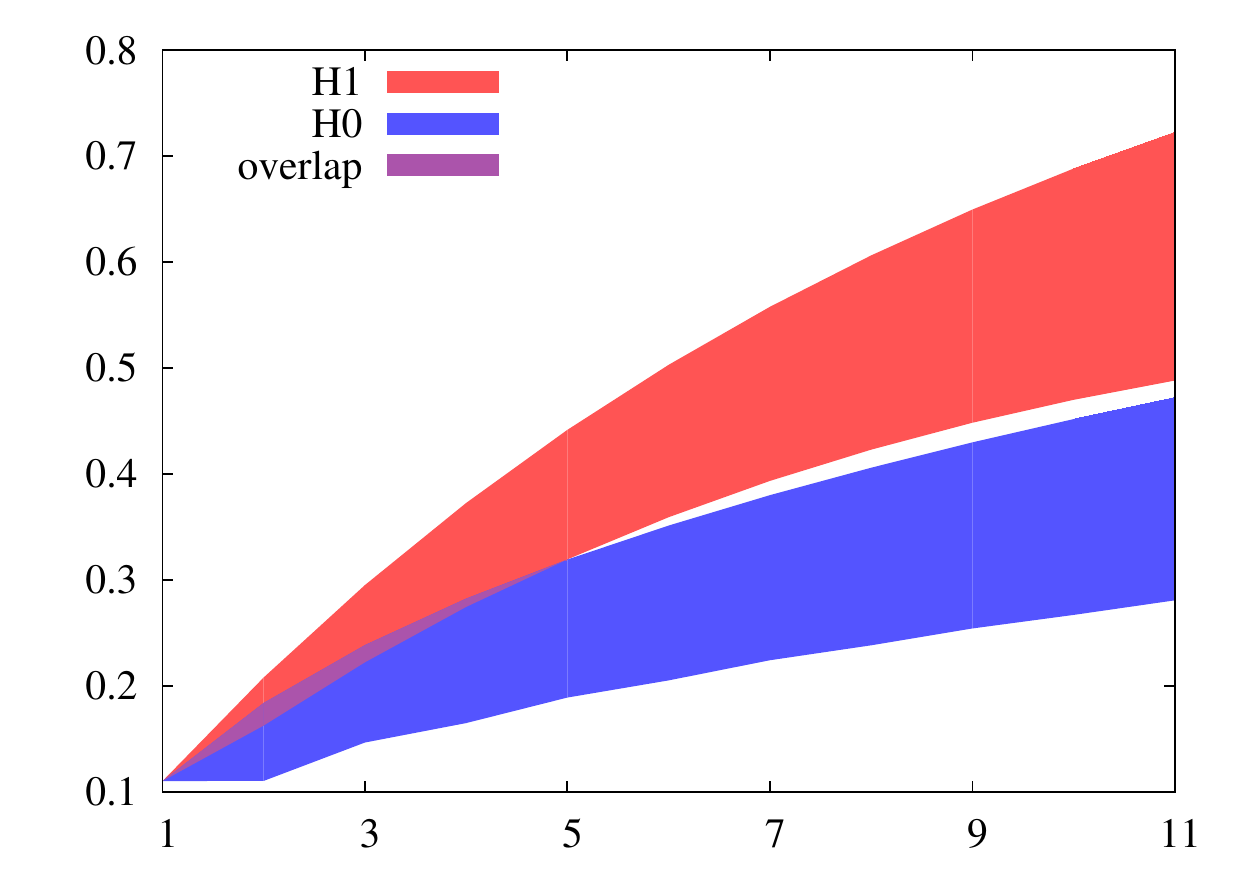}
		\put(-329,115){\small$d_{\rm TV}(P^m,Q^m)$}
		\put(-141,-3){\small $m$}
	\end{center}
	\caption{Evolution of achievable total variation distances 
	$d_{\rm TV}(P^m,Q^m)$ over packing size $m$  
	for two families of the target-generator pairs $H_0(0.05,0.1,0.11)$ and $H_1(0.02,0.1,0.11)$. The mode-collapsing $H_1$ is penalized significantly 
	by the discriminator (only with $m>1$) 
		and the two families can be strictly separated with packing for $m>5$. 
	}
	\label{fig:power5}
\end{figure}

\bigskip\noindent
{\bf The benefit of packing degree $m$.} 
We give a practitioner the choice of the degree $m$ of packing, namely how many samples to jointly pack together. 
There is a natural trade-off between computational complexity (which increases gracefully with $m$) and 
the additional distinguishability, which we illustrate via an example. 
Consider the goal of differentiating two families of target-generator pairs, one with mode collapse and one without:
\begin{align}
	&H_0(\varepsilon,\delta,\tau) \triangleq \{(P,Q)|(P,Q) \text{ without }(\varepsilon,\delta)\text{-mode collapse or  augmentation,} \text{ and } d_{\rm TV}(P,Q)=\tau \}\;,\nonumber\\
	&H_1(\varepsilon,\delta,\tau) \triangleq \{(P,Q)|(P,Q) \text{ with }(\varepsilon,\delta)\text{-mode collapse} \text{ and } d_{\rm TV}(P,Q)=\tau \} \;. \label{eq:defH}
\end{align} 
As both families have the same total variation distances, they cannot be distinguished by an unpacked discriminator.  
However, a packed discriminator that uses $m$ samples jointly 
can differentiate those two classes and even separate them entirely for a certain choices of parameters, 
as illustrated in Figure \ref{fig:power5}.  
In red, we show the achievable $d_{\rm TV}(P^m,Q^m)$ for $H_1(\varepsilon=0.02,\delta=0.1,\tau=0.11)$ (the bounds in Theorem \eqref{thm:main2}). 
In blue is shown a similar region for $H_0(\varepsilon=0.05,\delta=0.1,\tau=0.11)$ (the bounds in Theorem \eqref{thm:main3}). 
Although the two families are strictly separated (one with $\varepsilon=0.02$ and another with $\varepsilon=0.05$), 
a non-packed discriminator cannot differentiate those two families as the total variation is the same for both. 
However, as you pack mode samples, the packed discriminator becomes more powerful in differentiating the two hypothesized families. 
For instance, for $m\geq5$, the total variation distance completely separates the two families. 

In general, the overlap between those regions depends on the specific choice of parameters, 
but the overall trend is universal: 
packing separates generators with mode collapse from those without. 
Further, as the degree of packing increases, a packed discriminator increasingly penalizes generators with mode collapse and 
rewards generators that exhibit less mode collapse. 
Even if we consider complementary sets  $H_0$ and $H_1$ with the same $\varepsilon$ and $\delta$ 
(such that the union covers the whole space of pairs of $(P,Q)$ with the same total variation distance), 
the least penalized pairs will be those with least mode collapse, which fall within the blue region of the bottom right panel in Figure \ref{fig:power3}.
This is consistent with the empirical observations in Tables \ref{tbl:veegan1} and \ref{tbl:unrolled}, where 
increasing the degree of packing captures more modes. 

\bigskip\noindent
{\bf Jensen-Shannon divergence.} 
Our theoretical analysis focused on $0$-$1$ loss, 
as our current analysis technique gives exact solutions to the optimization problems 
\eqref{eq:opt_main1}, \eqref{eq:opt_main2}, and \eqref{eq:opt_main3} 
if the metric is total variation distance. 
This follows from the fact that 
we can 
provide tight inner and outer regions to the family of mode collapse regions $\cR(P,Q)$ that have the same total variation distances as 
$d_{\rm TV}(P,Q)$
as shown in Section \ref{sec:proof}.

In practice, $0$-$1$ loss is never used, as it is not differentiable. 
A popular choice of a loss function is the cross entropy loss in \eqref{eq:crossentropy}, 
which gives a metric of Jensen-Shannon (JS) divergence, as shown in the beginning 
of Section \ref{sec:theory}. 
However, the same proof techniques used to show Theorems \ref{thm:main2} and \ref{thm:main3} give loose bounds on JS divergence. 
In particular, this gap prevents us from sharply characterizing the full effect of packing degree $m$ on the JS divergence of a pair of distributions.
Nonetheless, we find that empirically, packing seems to reduce mode collapse even under a cross entropy loss. 
It is an interesting open question to find solutions to the optimization problems 
\eqref{eq:opt_main1}, \eqref{eq:opt_main2}, and \eqref{eq:opt_main3}, when the metric is the (more common) Jensen-Shannon divergence.

\begin{figure}[h]
	\begin{center}
	\includegraphics[width=.45\textwidth]{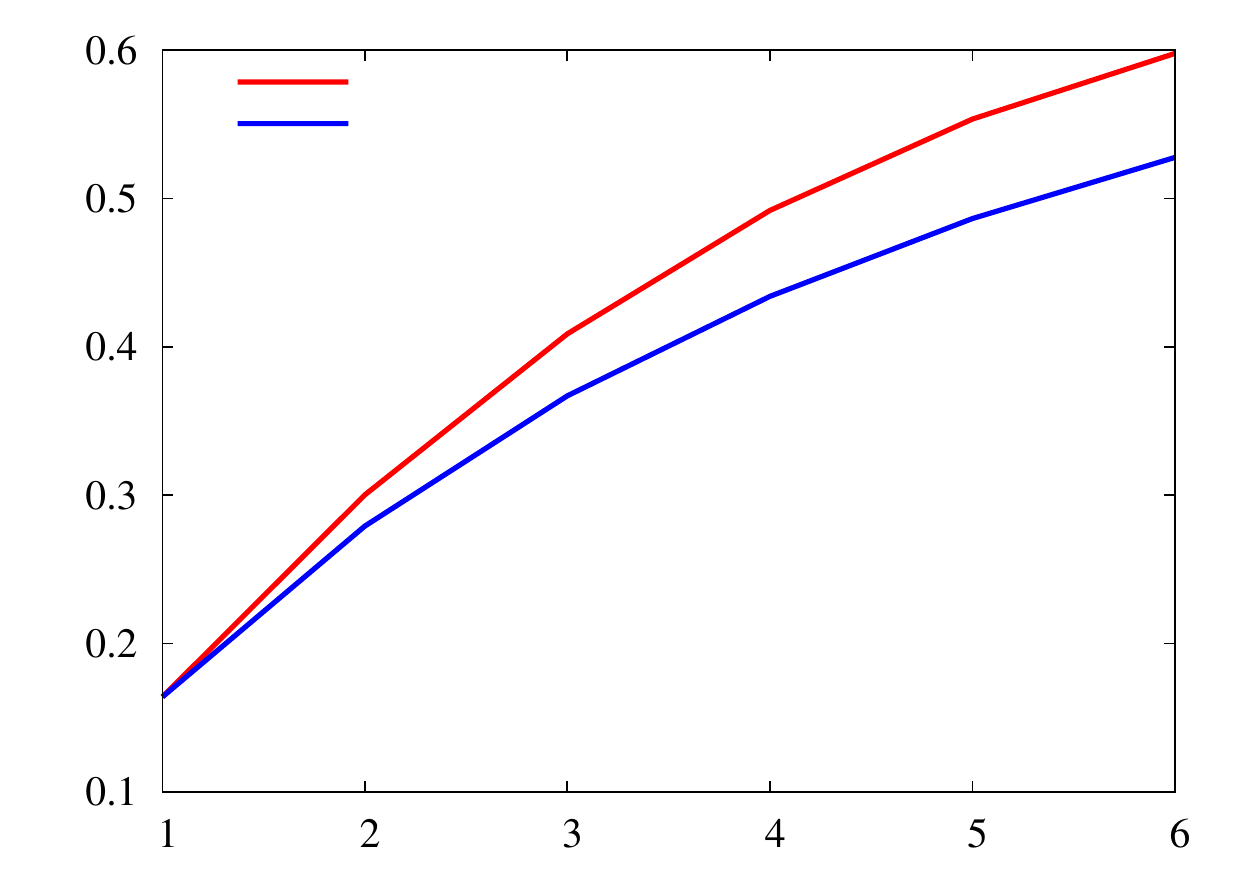}
	\put(-145,-6){degree of packing $m$}
	\put(-230,145){Jensen-Shannon divergence}
	\put(-150,133){\tiny$d_{\rm JS}(P^m,Q_1^m)$}
	\put(-150,126){\tiny$d_{\rm JS}(P^m,Q_2^m)$} 
	\end{center}
	\caption{Jensen-Shannon divergence increases faster as a function of packing degree $m$ for a mode collapsing generator $Q_1$, compared to a non mode collapsing generator $Q_2$.}
	\label{fig:toy_js}
\end{figure}

Although our proposed analysis technique does not provide a tight analysis for JS divergence, 
we can still analyze a toy example similar to the one in Section \ref{sec:productregion}. 
Consider a toy example 
with a uniform target  distribution $P=U([0,1])$ over $[0,1]$,  
a mode collapsing generator $Q_1=U([0.4,1])$, and 
a non mode collapsing generator 
$Q_2 = 0.285\, U([0,0.77815]) + 3.479\, U([0.77815,1])$.   
They are designed to have the same Jensen-Shannon divergence, i.e.~$d_{\rm JS}(P,Q_1)=d_{\rm JS}(P,Q_2)=0.1639$,  
but $Q_1$ exhibits an extreme mode collapse as the whole probability mass in $[0,0.4]$ is lost, whereas 
$Q_2$ captures a more balanced deviation from $P$.  Figure \ref{fig:toy_js} shows that 
the mode collapsing $Q_1$ have large JS divergence (and hence penalized more) under packing, compared to the non mode collapsing $Q_2$. 

\subsection{Operational interpretation of mode collapse via hypothesis testing region} 
\label{sec:region}
So far, all the definitions and theoretical results have been explained without explicitly using the {\em mode collapse region}. 
The main contribution of introducing the region definition is that it provides a new  proof technique   
  based on the geometric properties of these two-dimensional regions. 
Concretely, we show that the proposed mode collapse region is equivalent to a similar notion in binary hypothesis testing. 
This allows us to bring powerful mathematical tools from this mature area in statistics and information theory---in particular, 
the {\em data processing inequalities} 
originating from the seminal work of Blackwell \cite{Bla53}. 
We make this connection precise, 
which gives insights on how to interpret the mode collapse region, 
and list the properties and techniques 
which dramatically simplify the proof, while providing the tight 
results in Section \ref{sec:proof}. 

\subsubsection{Equivalence between the mode collapse region and the ROC curve}

There is a simple one-to-one correspondence between mode collapse region as we define it in Section \ref{sec:definition}  (e.g.~Figure \ref{fig:toy}) and 
the ROC curve studied in binary hypothesis testing. 
In the classical testing context,  there are two hypotheses, $h=0$ or $h=1$, 
and we make observations via some stochastic experiment in which our observations 
depend on the hypothesis. Let $X$ denote this observation. 
One way to visualize such an experiment is using 
a two-dimensional region defined by the corresponding type I and type II errors. 
This was, for example, used to prove strong composition theorems in the context of differential privacy in 	\cite{KOV17}, 
and subsequently to identify the optimal  differentially private mechanisms under 
local privacy \cite{KOV14} and multi-party communications \cite{KOV15}. 
Concretely, an ROC curve of a binary hypothesis testing is 
obtained by plotting the largest achievable true positive rate (TPR), i.e.~$1-$probability of missed detection, or equivalently $1-$ type II error,  
on the vertical axis 
against the false positive rate (FPR), i.e~probability of false alarm or equivalently type I error, on the horizontal axis.

We can map this binary hypothesis testing setup directly to the GAN context. 
Suppose the null hypothesis $h=0$ denotes observations being drawn from the generated distribution $Q$,
and the alternate hypothesis $h=1$ denotes observations being drawn from the true distribution $P$.
Given a sample $X$ from this experiment, 
suppose we make a decision on whether the sample came from $P$ or $Q$ based on a rejection region $S_{\rm reject}$, such that 
we reject the null hypothesis if $X \in S_{\rm reject}$.  
FPR (i.e.~Type I error) is when the null hypothesis is true but rejected, which happens with $\prob ( X\in S_{\rm reject} | h=0)$, 
and TPR (i.e.~1-type II error) is when the null hypothesis is false and rejected, which happens with $\prob(X\in S_{\rm reject} | h=1)$.
Sweeping through the achievable pairs $(\prob(X\in S_{\rm reject} | h=1),\prob ( X\in S_{\rm reject} | h=0))$ for all possible rejection sets, 
this defines a two dimensional convex region 
that we call {\em hypothesis testing region}. 
The upper boundary of this convex set is the ROC curve. 
An example of ROC curves for the two toy examples $(P,Q_1)$ and $(P,Q_2)$ from Figure \ref{fig:toy} are shown below in Figure \ref{fig:toy_equivalence}. 

\begin{figure} [ht]
	\begin{center} 
	\includegraphics[width=.36\textwidth]{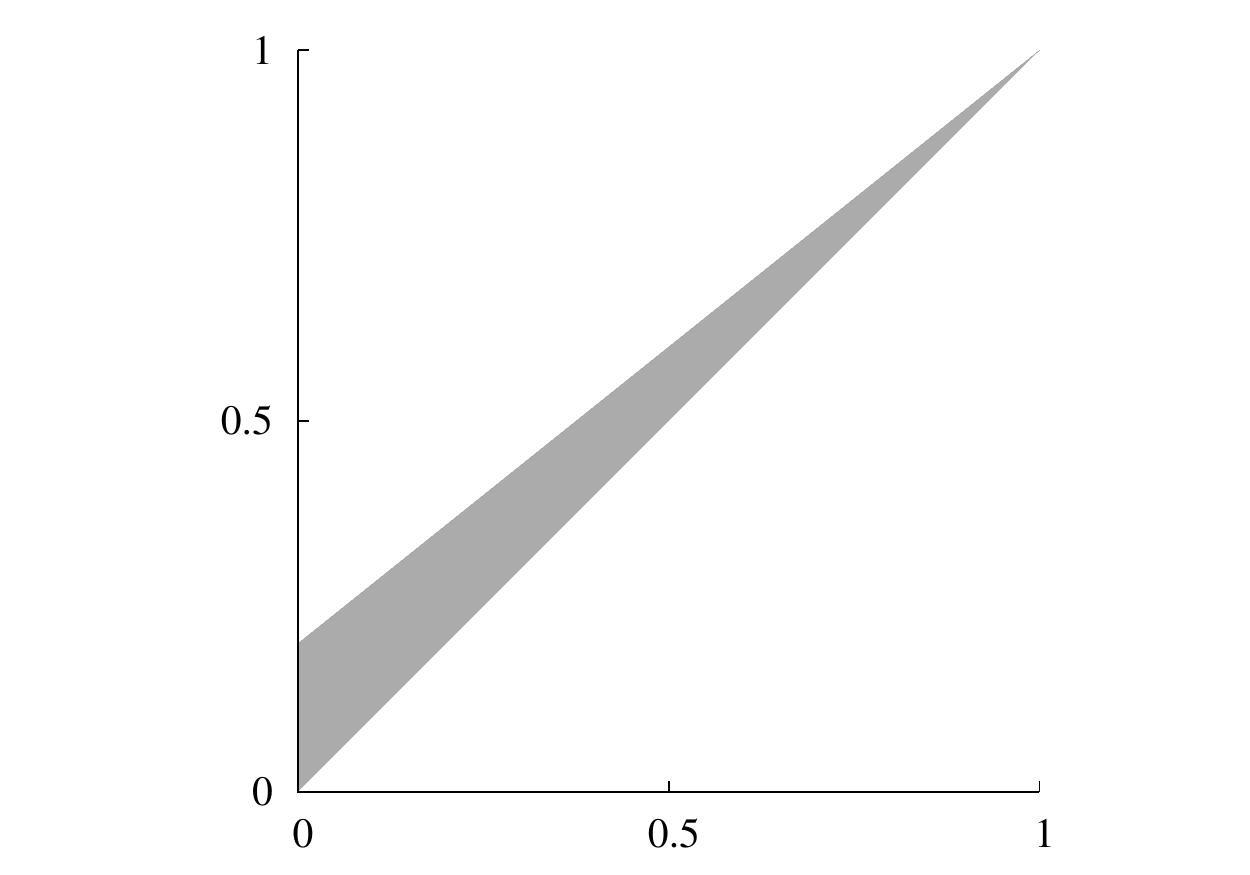}
	\put(-70,60){$\cR(P,Q_1)$}
	\put(-80,-4){$\varepsilon$}
	\put(-148,60){$\delta$}
	\put(-170,128){Mode collapse region for $(P,Q_1)$}
	\includegraphics[width=.36\textwidth]{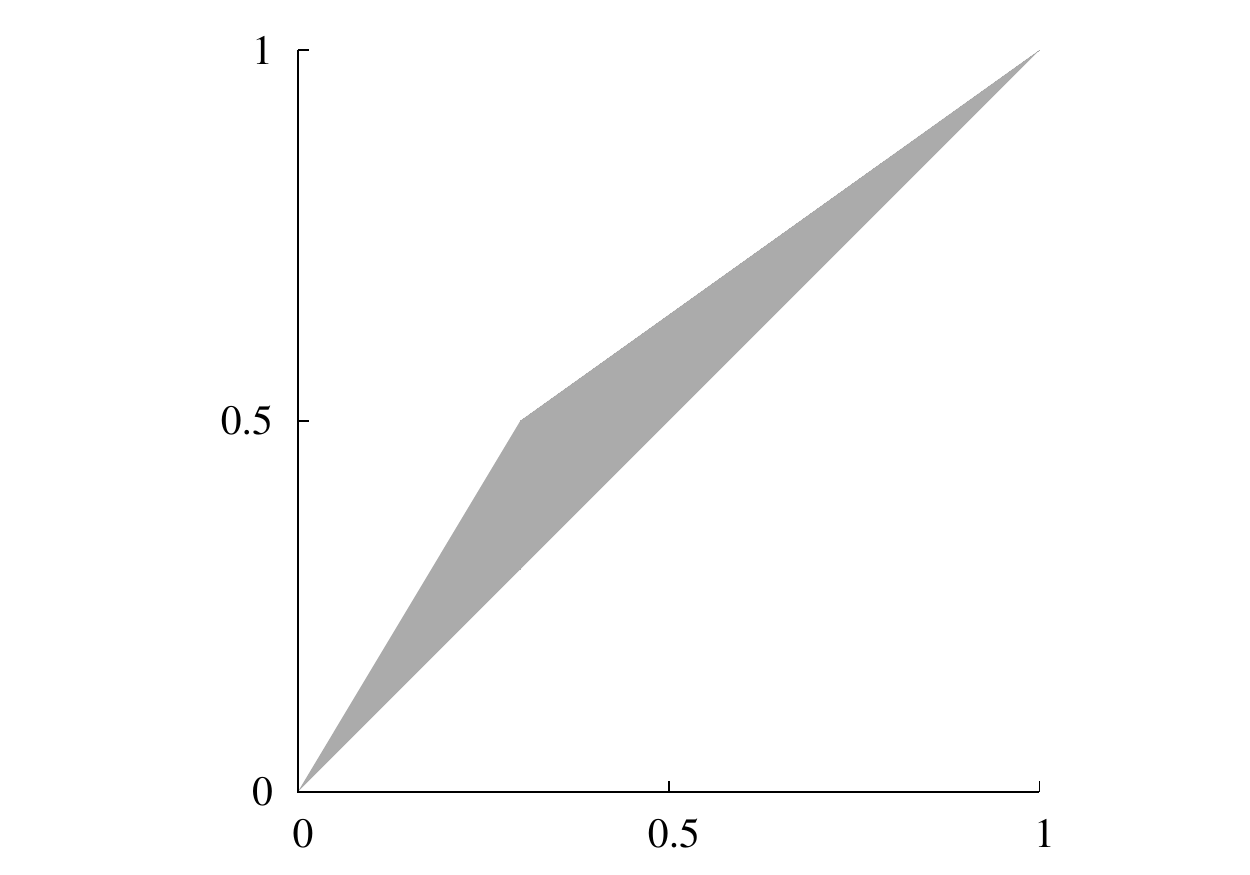}
	\put(-70,60){$\cR(P,Q_2)$}
	\put(-80,-4){$\varepsilon$}
	\put(-148,60){$\delta$}
	\put(-140,128){Mode collapse region for $(P,Q_2)$}
	\\
	\vspace{0.3cm}
	\includegraphics[width=.36\textwidth]{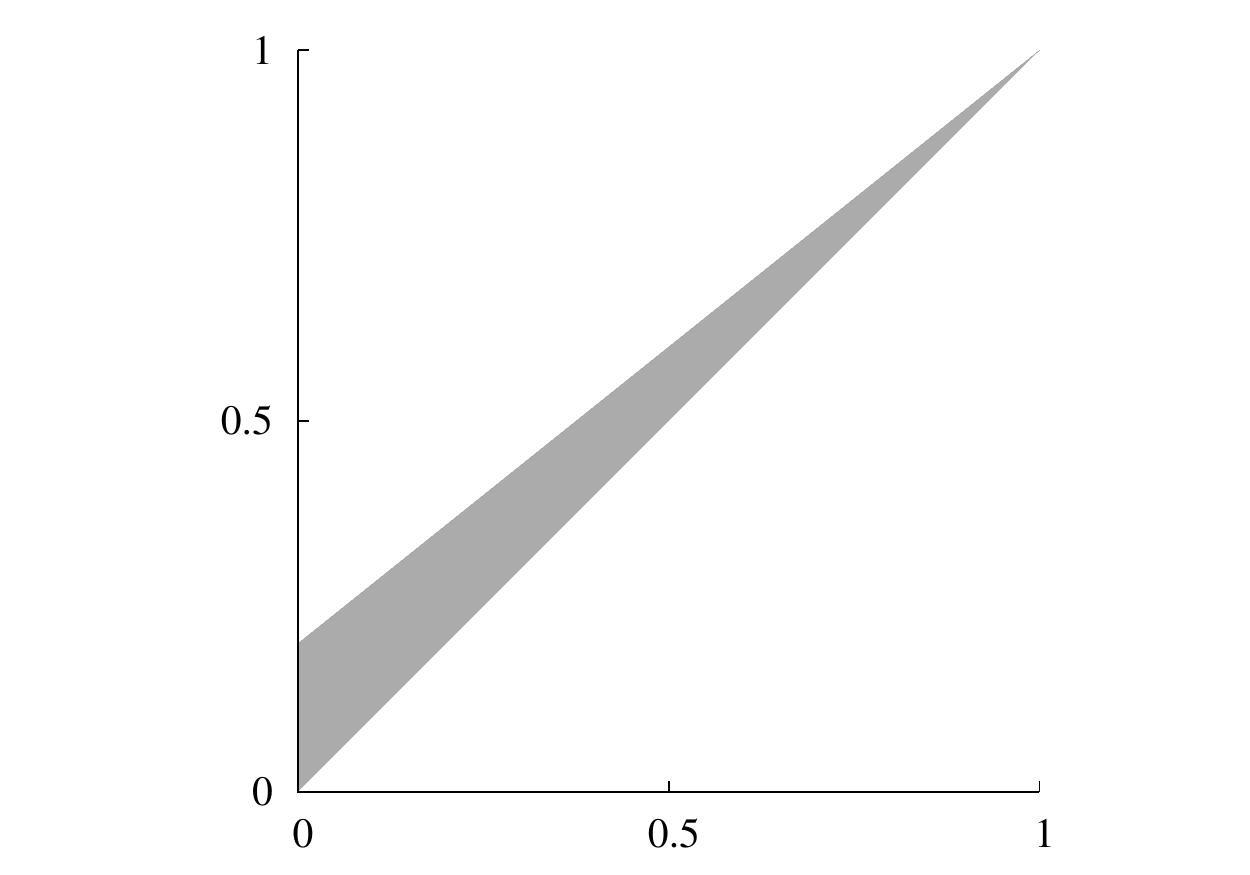}
	\put(-128,-9){False Positive Rate (FPR)}
	\put(-227,64){True Positive Rate}
	\put(-178,128){Hypothesis testing region for $(P,Q_1)$}
	\includegraphics[width=.36\textwidth]{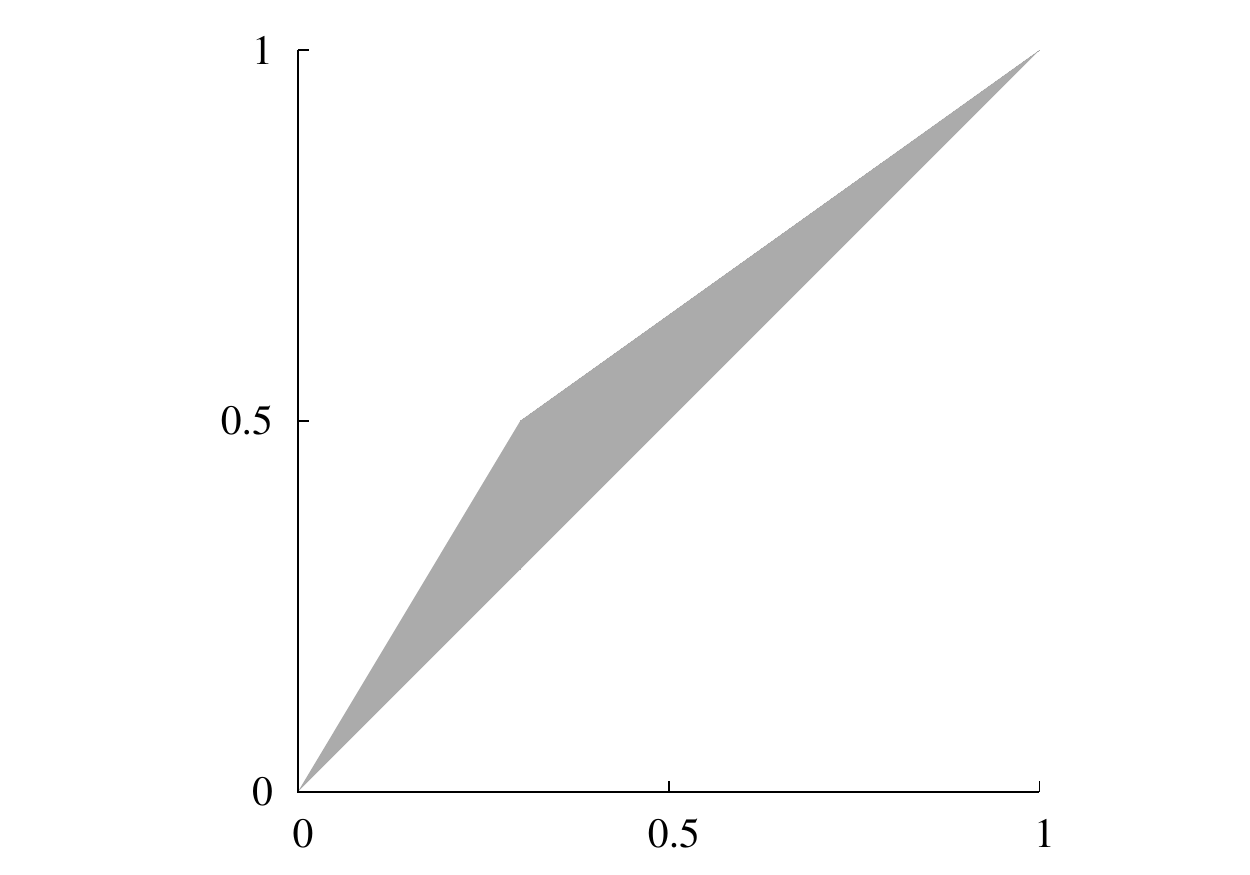}
	\put(-128,-9){False Positive Rate (FPR)}
	\put(-227,64){True Positive Rate}
	\put(-140,128){Hypothesis testing region for $(P,Q_2)$}
	\end{center} 
	\caption{The hypothesis testing region of $(P,Q)$ (bottom row) 
	is the same as  the mode collapse region (top row). 
	We omit the region below the $y=x$ axis in the hypothesis testing region as it is symmetric.  
	The regions for mode collapsing toy example in Figure \ref{fig:toy} $(P,Q_1)$ are shown on the left and 
	the regions for the non mode collapsing example $(P,Q_2)$ are shown on the right. 
	} 
	\label{fig:toy_equivalence} 
\end{figure} 

In defining the region, we allow stochastic decisions, such that if a point $(x,y)$ and another point $(x',y')$ 
are achievable TPR and FPR, then any convex combination of those points are also achievable by 
randomly choosing between those two rejection sets. 
Hence, the resulting hypothesis testing region is always a convex set by definition. 
We also show only the region above the 45-degree line passing through $(0,0)$ and $(1,1)$, as the other region is symmetric and redundant. 
For a given pair $(P,Q)$, there is a very simple relation between its mode collapse region and hypothesis testing region. 

\begin{remark}[Equivalence]
	\label{rem:hyp}
	For a pair of target $P$ and generator $Q$, 
	the hypothesis testing region is 
	the same as the mode collapse region. 
\end{remark}

This follows immediately from the 
definition of mode collapse region in Definition \ref{def:modecollapse}.  
If there exists a set $S$ such that $P(S) = \delta$ and $ Q(S)= \varepsilon$, then 
for the choice of $S_{\rm reject}=S$ in the binary hypothesis testing, 
then the point $(\prob(X\in S_{\rm reject}|h=0)=\varepsilon,\prob(X\in S_{\rm reject}|h=1)=\delta)$ in the hypothesis testing region is achievable.  
The converse is also true, in the case we make deterministic decisions on $S_{\rm reject}$. 
As the mode collapse region is defined as a convex hull of all achievable points, 
the points in the hypothesis testing region that require randomized decisions can also be covered. 


For example, the hypothesis testing regions of the toy examples from Figure \ref{fig:toy} are shown below in Figure \ref{fig:toy_equivalence}. 
This simple relation allows us to tap into the rich analysis tools known for hypothesis testing regions and ROC curves.
We list such properties of mode collapse regions derived from this relation in the next section. 
The proof of all the remarks follow from the equivalence to binary hypothesis testing and 
corresponding existing results from \cite{Bla53} and \cite{KOV17}.

%
%



\subsubsection{Properties of the mode collapse region} 
\label{sec:properties}

\begin{figure}[ht]
	\begin{center}
	\includegraphics[width=.36\textwidth]{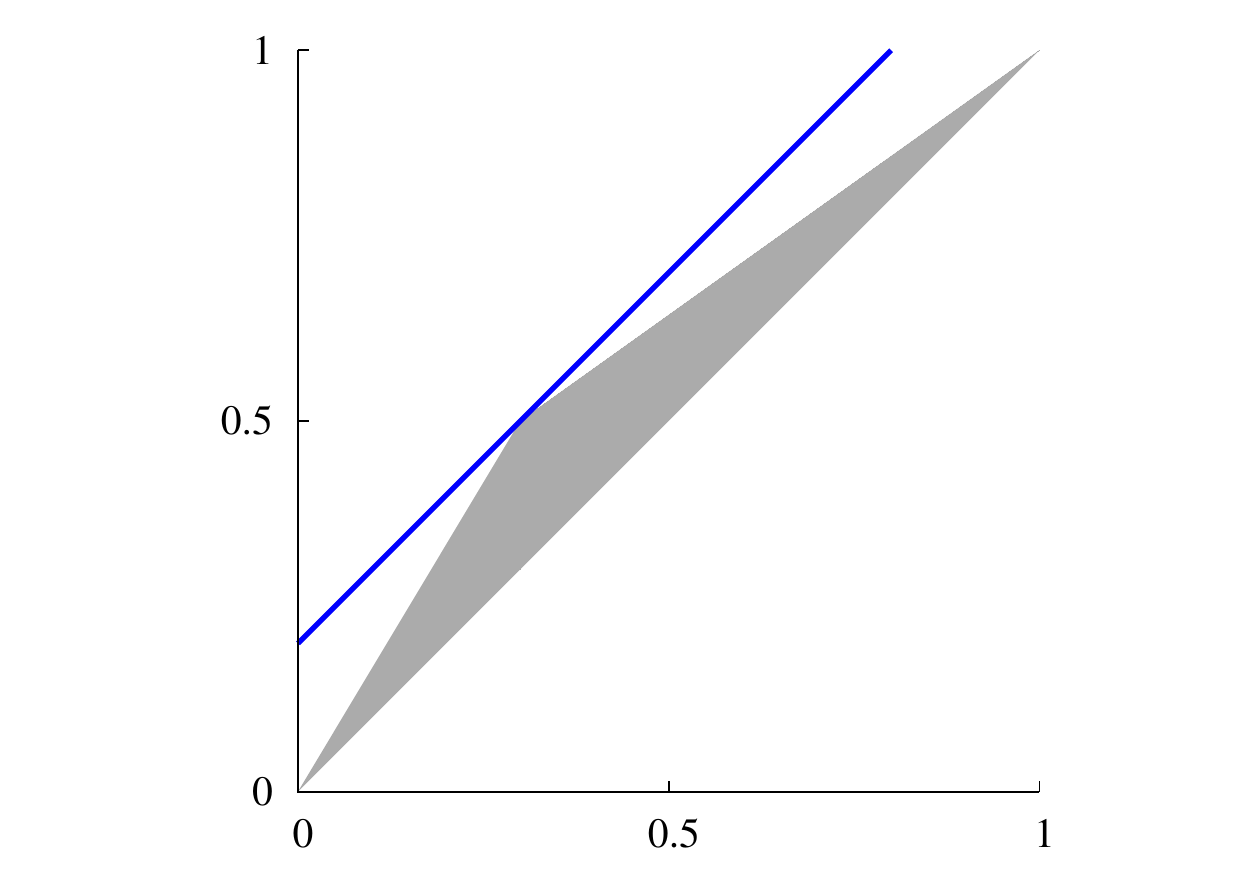}
	\put(-138,18){$\Big\Updownarrow$}
	\put(-188,18){$d_{\rm TV}(P,Q_2)$}
	\put(-70,60){$\cR(P,Q_2)$}
	\put(-80,-4){$\varepsilon$}
	\put(-148,60){$\delta$}
	\put(-80,113){slope $=1$}
	\end{center}
	\caption{Total variation distance is one among many properties of $(P,Q_2)$ that can be directly read off of the region $\cR(P,Q)$.}
	\label{fig:region}
\end{figure}

Given the equivalence between the mode collapse region and the binary hypothesis testing region, 
several important properties follow as corollaries. 
First, the hypothesis testing region is a sufficient statistic for the purpose of binary hypothesis testing 
from a pair of distributions $(P,Q)$. 
This implies, among other things, that all $f$-divergences can be derived from the region. 
In particular, for the purpose of GAN with $0$-$1$ loss, 
we can define total variation as a geometric property of the region, 
which is crucial to proving our main results. 
\begin{remark}[Total variation distance]
	\label{rem:tv}
	The total variation distance between $P$ and $Q$ 
	is the intersection between the vertical axis and 
	the tangent line to the upper boundary of $\cR(P,Q)$ that has a slope of one, 
	as shown in Figure \ref{fig:region}. 
\end{remark}

This follows from the equivalence of the mode collapse region (Remark \ref{rem:hyp}) and the hypothesis testing region. 
This geometric definition of total variation allows us to enumerate over all pairs $(P,Q)$ that have the same total variation $\tau$ in our proof, 
via enumerating over all regions that touch the line that has a unit slope and a shift $\tau$ (see Figure \ref{fig:proof1_1}). 

The major strength of the region perspective, as originally studied by Blackwell \cite{Bla53}, is in providing a comparison of stochastic experiments.
In our GAN context, 
consider comparing two pairs of target distributions and generators $(P,Q)$ and $(P',Q')$ as follows. 
First, a hypothesis $h$ is drawn, choosing whether to produce samples from the true distribution, in which case we say $h=1$, 
or to produce samples from the generator, in which case we say $h=0$. 
Conditioned on this hypothesis $h$, we use $X$ to denote a random variable that is drawn from the first pair $(P,Q)$ such that 
$f_{X|h}(x|1)=P(x)$ and $f_{X|h}(x|0)=Q(x)$. 
Similarly, we use $X'$ to denote a random sample from the second pair, where 
$f_{X'|h}(x|1)=P'(x)$ and $f_{X'|h}(x|0)=Q'(x)$. 
Note that the  conditional distributions are well-defined for both $X$ and $X'$, but there is no coupling defined between them. 
Suppose $h$ is independently drawn from the uniform distribution. 

\begin{definition}
	For a given coupling between $X$ and $X'$, 
	we say $X$ {\em dominates} $X'$ if they form a Markov chain $h\text{--}X\text{--}X'$. 
\end{definition}

The {\em data processing inequality} in the following remark shows that if we further {\em process} the output samples from the pair $(P,Q)$
then the further processed samples can only have less mode collapse. 
Processing output of stochastic experiments has the effect of smoothing out the distributions, and mode collapse, which corresponds to a {\em peak} in the pair of distributions,  
are smoothed out in the processing down the Markov chain. 

\begin{remark}[Data processing inequality]
	\label{rem:dp0}
	The following data processing inequality holds for the mode collapse region. 
	For two coupled target-generator pairs $(P,Q)$ and $(P',Q')$, if 
	$X$ {\em dominates} another pair $X'$,
	then 
	$$\cR(P',Q') \;\; \subseteq \;\; \cR(P,Q)\;.$$
\end{remark}
This is expected, and follows directly from the equivalence of the mode collapse region (Remark \ref{rem:hyp}) and the hypothesis testing region, 
and corresponding data processing inequality of hypothesis  testing region in \cite{KOV17}.  
What is perhaps surprising is that the reverse is also true. 

\begin{remark}[Reverse data processing inequality]
	\label{rem:dp}
	The following reverse data processing inequality holds for the mode collapse region. 
	For two paired marginal distributions  $X$ and $X'$, if 
	$$\cR(P',Q') \;\; \subseteq \;\; \cR(P,Q)\;,$$
	then there exists a coupling of the random samples from $X$ and $X'$ such that $X$ dominates $X'$, i.e.~they form a Markov chain 
	$h\text{--}X\text{--}X'$.
\end{remark}
This follows from the equivalence between the mode collapse region and the hypothesis testing region 
(Remark \ref{rem:hyp}) and Blackwell's celebrated result on comparisons of stochastic experiments \cite{Bla53} (see \cite{KOV17} for a simpler version of the statement). 
This region interpretation, and the accompanying (reverse) data processing inequality, abstracts away all the details about $P$ and $Q$, enabling us to
use geometric analysis tools to prove our results. 
In proving our main results, we will mainly rely on the following remark, which is the corollary of the Remarks \ref{rem:dp0} and \ref{rem:dp}. 

\begin{remark}
	\label{rem:blackwell}
	For all positive integers $m$, the dominance of regions are preserved under taking 
	$m$-th order product distributions, i.e.~if $\cR(P',Q')  \subseteq  \cR(P,Q)$, then 	$\cR((P')^m,(Q')^m)  \subseteq  \cR(P^m,Q^m)$.
\end{remark}

%
%
%

\section{Proofs of the main results}
\label{sec:proof}
In this section, we showcase how the region interpretation provides a new proof technique that is 
simple and tight. 
This transforms the measure-theoretic problem into a geometric one in a simple 2D compact plane, 
 facilitating the proof of otherwise-challenging results. 

\subsection{Proof of Theorem \ref{thm:main1}}
\label{sec:proof1}

Note that although the original optimization \eqref{eq:opt_main1} has nothing to do with mode collapse, 
we use the mode collapse region to represent the pairs $(P,Q)$ to be optimized over. 
This allows us to use simple geometric techniques  to enumerate over all possible pairs $(P,Q)$ that 
have the same total variation distance $\tau$. 

\begin{figure}[ht]
	\begin{center}
		\includegraphics[width=.35\textwidth]{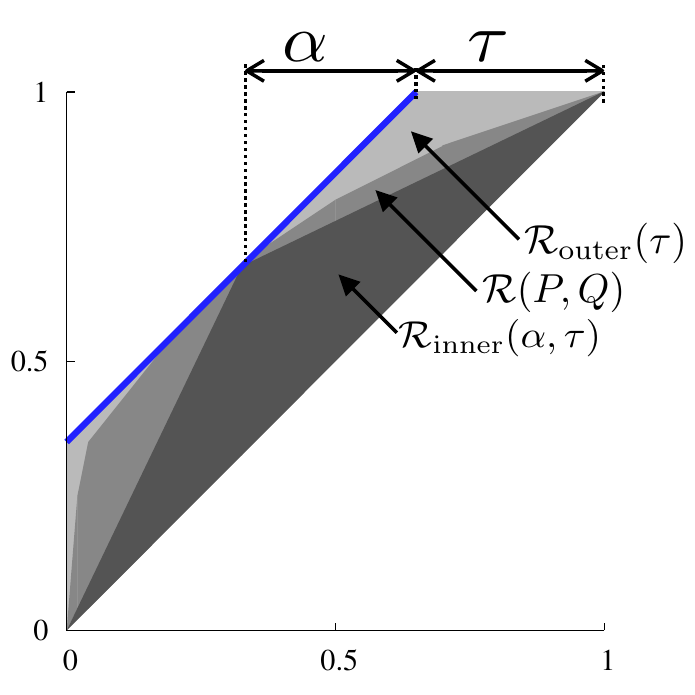}
		\put(-90,-6){$\varepsilon$}
		\put(-175,75){$\delta$}
	\end{center}
	\caption{For any pair $(P,Q)$ with total variation distance $\tau$, there exists an $\alpha$ such that 
	 the corresponding region $\cR(P,Q)$ is sandwiched 
	between $\cR_{\rm inner}(\alpha,\tau)$ and $\cR_{\rm outer}(\tau)$. 
	  }
	\label{fig:proof1_1}
\end{figure}
\begin{figure}[ht]
	\begin{center}
		\includegraphics[width=.45\textwidth]{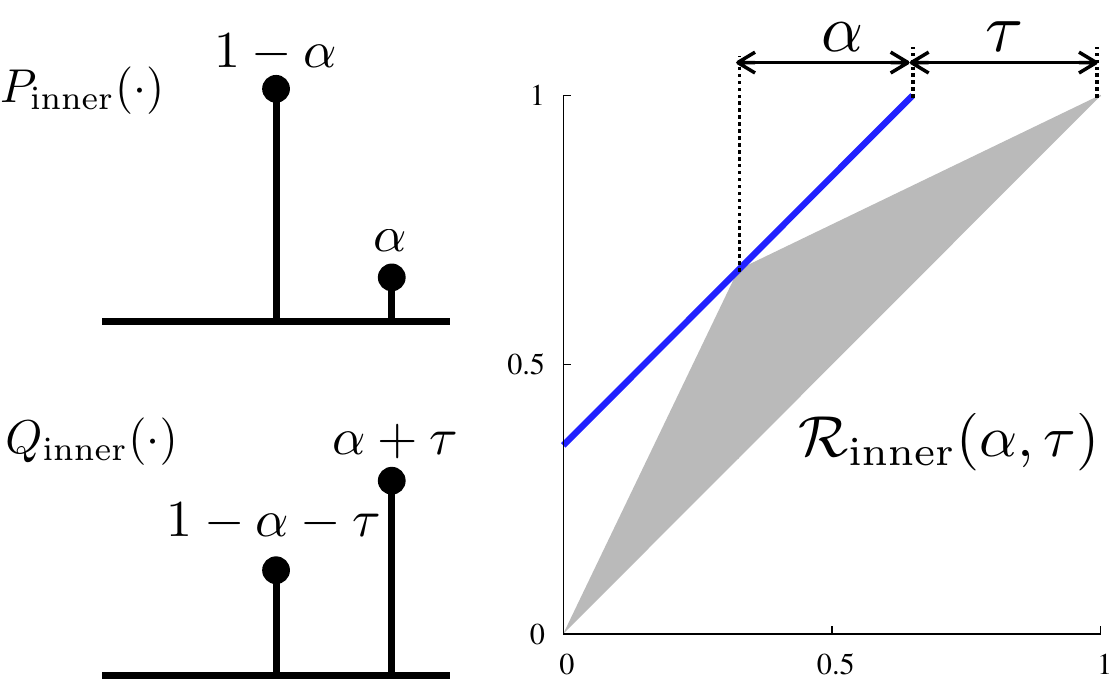}
		\put(-60,-6){$\varepsilon$}
		\put(-112,69){$\delta$}
		\hspace{0.8cm}
		\includegraphics[width=.45\textwidth]{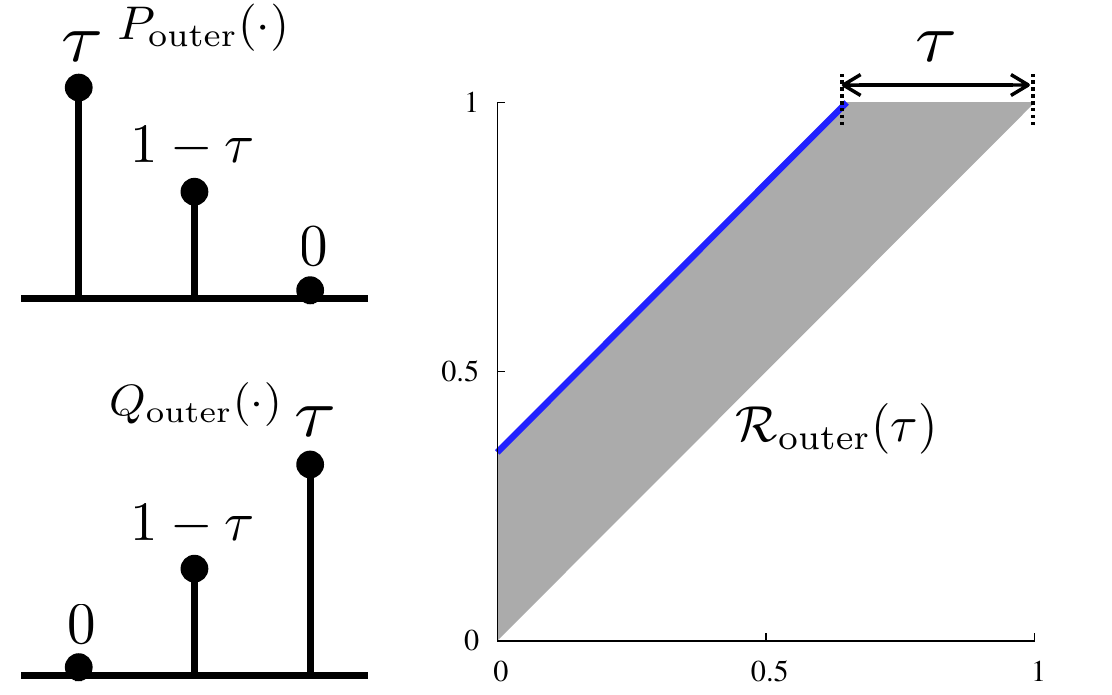}
		\put(-70,-6){$\varepsilon$}
		\put(-127,69){$\delta$}
	\end{center}
	\caption{Canonical pairs of distributions corresponding to $\cR_{\rm inner}(\alpha,\tau)$ and $\cR_{\rm outer}(\tau)$. 
	  }
	\label{fig:proof1_2}
\end{figure}

By Remark \ref{rem:tv}, all pairs $(P,Q)$ that have total variation $\tau$ must 
have a mode collapse region $\cR(P,Q)$ that is tangent to the blue line in Figure \ref{fig:proof1_1}. 
Let us denote a point where $\cR(P,Q)$ meets the blue line by 
the point $(1-\alpha-\tau,1-\alpha)$ in the 2D plane, parametrized by 
$\alpha \in [0,1-\tau]$. 
Then, for any such $(P,Q)$, we can  sandwich the region $\cR(P,Q)$ 
between two regions $\cR_{\rm inner}$ and $\cR_{\rm outer}$:  
\begin{eqnarray}
	\cR_{\rm inner}(\alpha,\tau)  \;\; \subseteq\;\;  \cR(P,Q) \;\; \subseteq\;\; \cR_{\rm outer}(\tau) \;,\label{eq:proof1_1}
\end{eqnarray}
which  
are illustrated in Figure \ref{fig:proof1_2}.  
Now, we wish to understand how these inner and outer regions evolve under product distributions. 
This endeavor is complicated by the fact that there can be infinite pairs of distributions that have the same region $\cR(P,Q)$.
However, note that if two pairs of distributions have the same region $\cR(P,Q)=\cR(P',Q')$, then their product distributions will also have the same region $\cR(P^m,Q^m)=\cR((P')^m,(Q')^m)$. 
As such, we can focus on the simplest, \emph{canonical} pair of distributions, whose support set has the minimum cardinality over all pairs of distributions with region $\cR(P,Q)$.

For a given $\alpha$, we denote the pairs of canonical distributions  
achieving these exact inner and outer regions as  in Figure \ref{fig:proof1_2}:
let  $(P_{\rm inner}(\alpha) , Q_{\rm inner}(\alpha,\tau))$ 
be as defined in \eqref{eq:pinner} and  \eqref{eq:qinner}, 
and let $(P_{\rm outer}(\tau) , Q_{\rm outer}(\tau))$ 
be defined as below.   
Since the outer region has three sides (except for the universal 45-degree line), 
we only need alphabet size of three to find the canonical probability distributions corresponding to the outer region.
By the same reasoning, the inner region requires only a binary alphabet.
Precise probability mass functions on these discrete alphabets can be found easily 
from the shape of the regions and the equivalence to the hypothesis testing region explained in Section \ref{sec:region}. 

By the preservation of dominance under product distributions in Remark \ref{rem:blackwell}, it follows from the dominance in \eqref{eq:proof1_1} that 
for any $(P,Q)$ there exists an $\alpha$ such that 
\begin{eqnarray}
	\cR(P_{\rm inner}(\alpha)^m , Q_{\rm inner}(\alpha,\tau)^m) \;\;\subseteq\;\;  \cR(P^m,Q^m) \;\; \subseteq \;\;  \cR(P_{\rm outer} (\tau)^m, Q_{\rm outer}(\tau)^m)\;.
	\label{eq:proof1_2}
\end{eqnarray}
Due to the data processing inequality of mode collapse region in Remark \ref{rem:dp}, 
it follows that dominance of region implies dominance of total variation distances: 
\begin{eqnarray}
	\min_{ 0 \leq \alpha \leq 1-\tau  } d_{\rm TV}(P_{\rm inner}(\alpha)^m,Q_{\rm inner}(\alpha,\tau)^m ) &\leq 
	\;\; d_{\rm TV}(P^m,Q^m) \;\; \leq &  d_{\rm TV}(P_{\rm outer}(\tau)^m,Q_{\rm outer}(\tau)^m)\;.
\end{eqnarray}
The RHS and LHS of the above inequalities can be completely characterized by taking the $m$-th power of those canonical pairs of distributions.  
For the upper bound, all mass except for $(1-\tau)^m$ is nonzero only on one of the pairs, which gives 
$d_{\rm TV}(P_{\rm outer}^m,Q_{\rm outer}^m) = 1-(1-\tau)^m$. 
For the lower bound, writing out the total variation gives $L(\tau,m)$  in \eqref{eq:defL}. 
This finishes the proof of Theorem \ref{thm:main1}. 

\subsection{Proof of Theorem \ref{thm:main2}}
\label{sec:proof2}

In optimization \eqref{eq:opt_main2}, we consider only those pairs with $(\varepsilon,\delta)$-mode collapse. 
It is simple to see that the outer bound does not change. 
We only need a new inner bound. 
Let us denote a point where $\cR(P,Q)$ meets the blue line by 
the point $(1-\alpha-\tau,1-\alpha)$ in the 2D plane, parametrized by 
$\alpha \in [0,1-\tau]$. 
We consider the case where $\alpha<1-(\tau\delta/(\delta-\varepsilon))$ for now, 
and treat the case when $\alpha$ is larger separately, as the analyses are similar 
but require a different canonical pair of distributions $(P,Q)$ 
for the inner bound.
The additional constraint that $(P,Q)$ has $(\varepsilon,\delta)$-mode collapse translates into a geometric constraint that 
we need to consider all regions $\cR(P,Q)$ that include the orange solid circle at point $(\varepsilon,\delta)$. 
Then, for any such $(P,Q)$, we can  sandwich the region $\cR(P,Q)$ 
between two regions $\cR_{\rm inner1}$ and $\cR_{\rm outer}$:  
\begin{eqnarray}
	\cR_{\rm inner1}(\varepsilon,\delta,\alpha,\tau)  \;\; \subseteq\;\;  \cR(P,Q) \;\; \subseteq\;\; \cR_{\rm outer}(\tau) \;,\label{eq:proof2_1}
\end{eqnarray}

\begin{figure}[ht]
	\begin{center}
		\includegraphics[width=.35\textwidth]{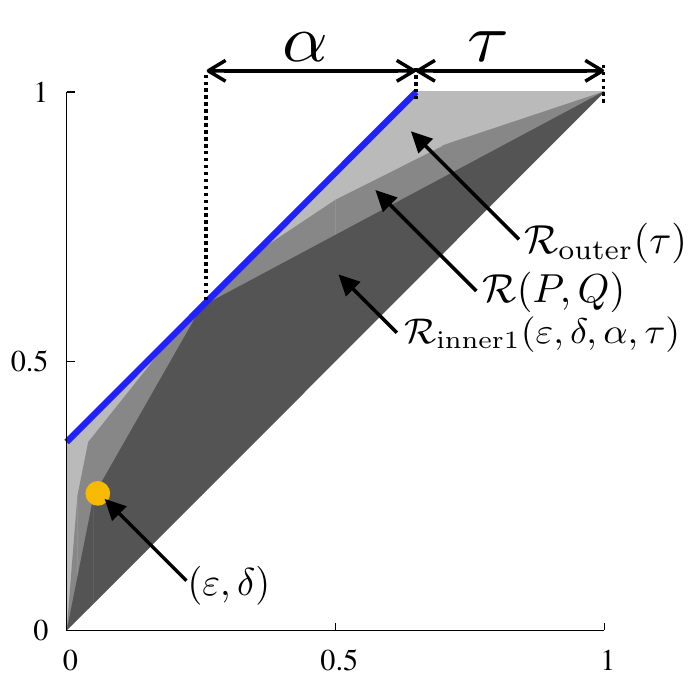}
		\put(-90,-6){$\varepsilon$}
		\put(-175,75){$\delta$}
	\end{center}
	\caption{For any pair $(P,Q)$ with $(\varepsilon,\delta)$-mode collapse, the corresponding region $\cR(P,Q)$ is sandwiched 
	between $\cR_{\rm inner1}(\varepsilon,\delta,\alpha,\tau)$ and $\cR_{\rm outer}(\tau)$. 
	  }
	\label{fig:proof2_1}
\end{figure}
\begin{figure}[ht]
	\begin{center}
		\includegraphics[width=.45\textwidth]{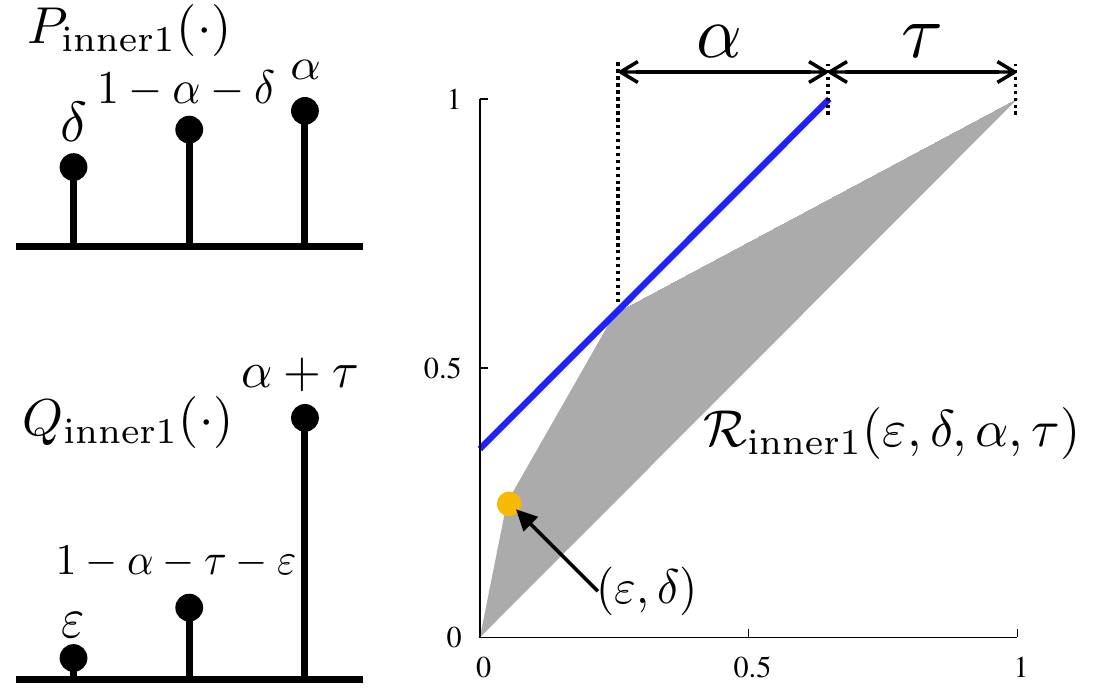}
		\put(-74,-6){$\varepsilon$}
		\put(-128,69){$\delta$}
		\hspace{0.8cm}
		\includegraphics[width=.45\textwidth]{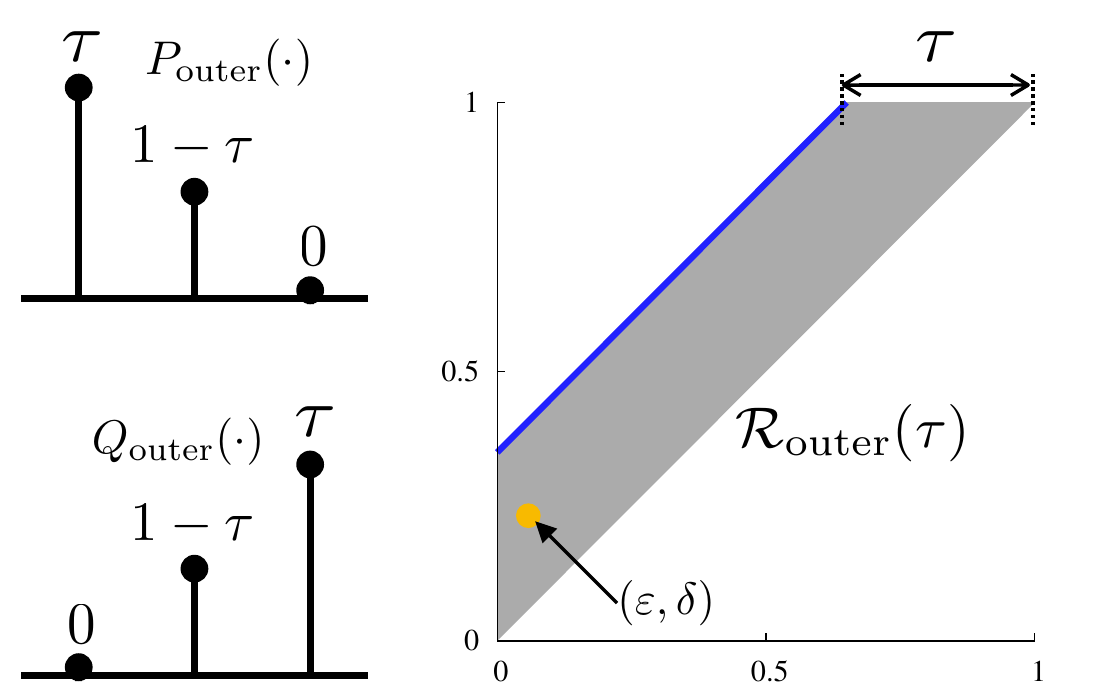}
		\put(-70,-6){$\varepsilon$}
		\put(-127,69){$\delta$}
	\end{center}
	\caption{Canonical pairs of distributions corresponding to $\cR_{\rm inner}(\varepsilon,\delta,\tau,\alpha)$ and $\cR_{\rm outer}(\tau)$. 
	  }
	\label{fig:proof2_2}
\end{figure}

Let $(P_{\rm inner1}(\delta,\alpha) , Q_{\rm inner1}(\varepsilon,\alpha,\tau))$ 
defined in \eqref{eq:pinner1} and  \eqref{eq:qinner1}, 
and $(P_{\rm outer}(\tau) , Q_{\rm outer}(\tau))$ defined in Section \ref{sec:proof1} 
denote the pairs of canonical distributions  
achieving the inner and outer regions exactly as shown in Figure \ref{fig:proof2_2}. 
By the preservation of dominance under product distributions in Remark \ref{rem:blackwell}, it follows from the dominance in \eqref{eq:proof2_1} that 
for any $(P,Q)$ there exists an $\alpha$ such that 
\begin{eqnarray}
	\cR(P_{\rm inner1}(\delta,\alpha)^m , Q_{\rm inner1}(\varepsilon,\delta,\alpha,\tau)^m) \;\;\subseteq\;\;  \cR(P^m,Q^m) \;\; \subseteq \;\;  \cR(P_{\rm outer} (\tau)^m, Q_{\rm outer}(\tau)^m)\;.
	\label{eq:proof2_2}
\end{eqnarray}
Due to the data processing inequality of mode collapse region in Remark \ref{rem:dp}, 
it follows that dominance of region implies dominance of total variation distances: 
\begin{align}
	\min_{ 0 \leq \alpha \leq 1-\frac{\tau\delta}{\delta-\varepsilon}   } d_{\rm TV}(P_{\rm inner1}(\delta,\alpha)^m,Q_{\rm inner1}(\varepsilon,\delta,\alpha,\tau)^m ) &\leq 
	\;\; d_{\rm TV}(P^m,Q^m) \;\; \leq &  d_{\rm TV}(P_{\rm outer}(\tau)^m,Q_{\rm outer}(\tau)^m)\;.
\end{align}
The RHS and LHS of the above inequalities can be completely characterized by taking the $m$-th power of those canonical pairs of distributions.  
For the upper bound, all mass except for $(1-\tau)^m$ is nonzero only on one of the pairs, which gives 
$d_{\rm TV}(P_{\rm outer}^m,Q_{\rm outer}^m) = 1-(1-\tau)^m$. 
For the lower bound, writing out the total variation gives $L_1(\varepsilon,\delta,\tau,m)$  in \eqref{eq:defL1}.

For $\alpha > 1-(\tau\delta/(\delta-\varepsilon))$, we need to consider a different class of canonical distributions for the inner region, shown below. 
The inner region $\cR_{\rm inner2}(\alpha,\tau)$ and corresponding canonical distributions 
$P_{\rm inner2}(\alpha)$ and $Q_{\rm inner2}(\alpha,\tau)$ defined in \eqref{eq:pinner2} and \eqref{eq:qinner2} are shown below. 
We take the smaller one between the total variation distance resulting from these two cases. 
Note that $\alpha\leq 1-\tau$ by definition. 
This finishes the proof of Theorem \ref{thm:main2}. 
\begin{figure}[ht]
	\begin{center}
		\includegraphics[width=.45\textwidth]{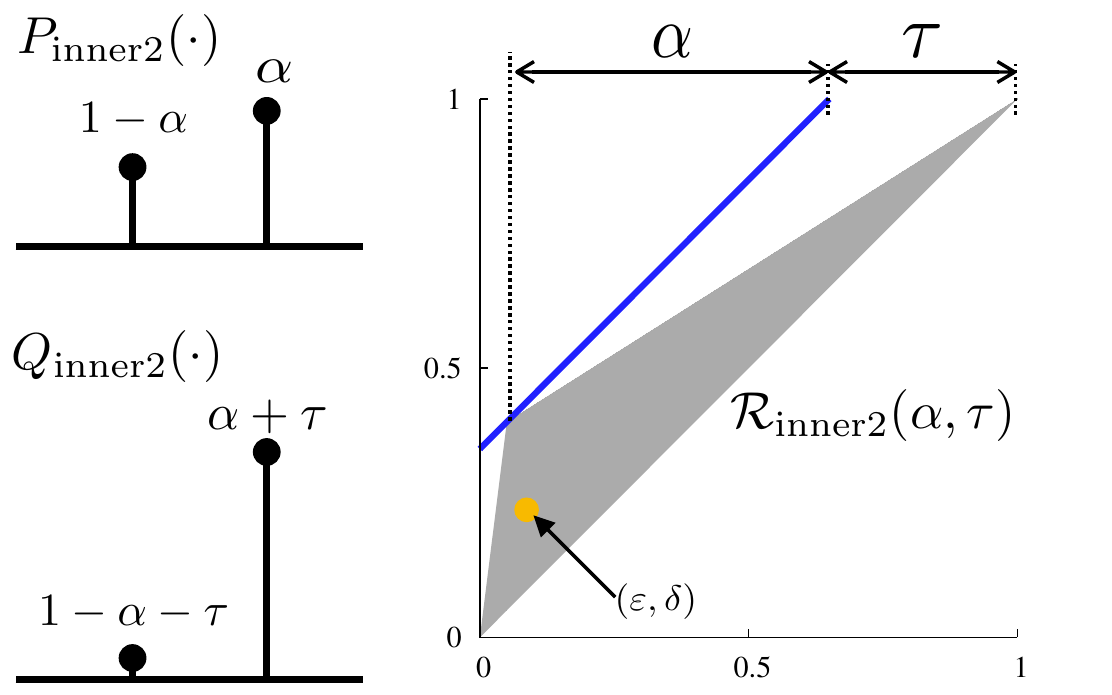}
		\put(-74,-6){$\varepsilon$}
		\put(-128,69){$\delta$}
	\end{center}
	\caption{When $\alpha > 1-(\tau\delta/(\delta-\varepsilon))$, this shows a canonical pair of distributions corresponding to $\cR_{\rm inner}(\varepsilon,\delta,\tau,\alpha)$ for the mode-collapsing scenario $H_1(\varepsilon,\delta,\tau)$. 
	  }
	\label{fig:proof2_1}
\end{figure} 

%
%

\subsection{Proof of Theorem \ref{thm:main3}}
\label{sec:proof3}
When $\tau<\delta-\varepsilon$, all pairs $(P,Q)$ with $d_{\rm TV}(P,Q)=\tau$ 
cannot have $(\varepsilon,\delta)$-mode collapse, and 
the optimization of \eqref{eq:opt_main3} reduces to that of \eqref{eq:opt_main1} without any mode collapse constraints. 

\begin{figure}[ht]
	\begin{center}
		\includegraphics[width=.3\textwidth]{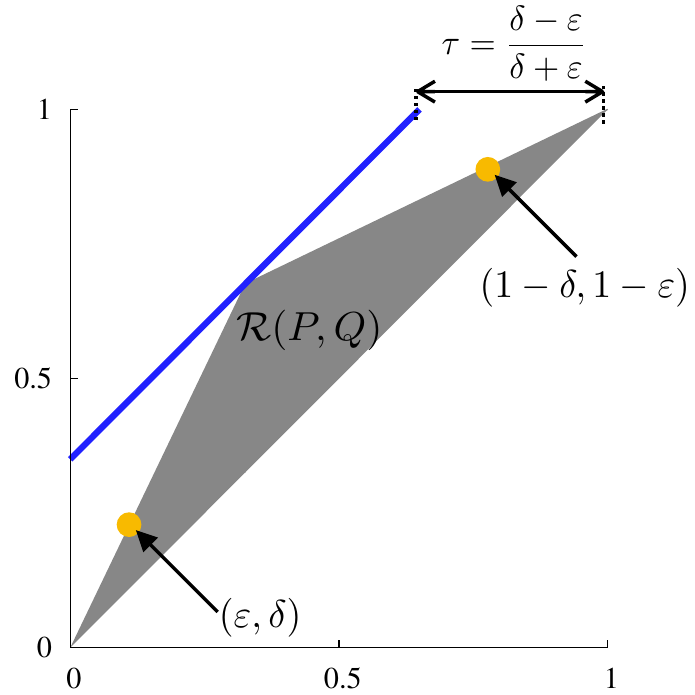}
		\put(-100,150){when $\varepsilon+\delta\leq1$}
		\hspace{0.5cm}
		\includegraphics[width=.3\textwidth]{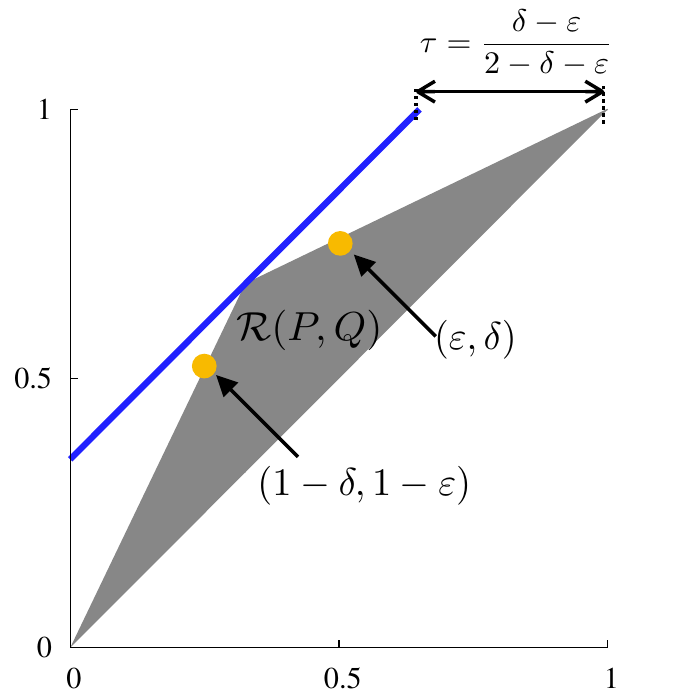}
		\put(-100,150){when $\varepsilon+\delta > 1$}
	\end{center}
	\caption{When $\delta+\varepsilon\leq1$ and 
	$\tau = (\delta-\varepsilon)/(\delta+\varepsilon)$ (i.e.~$(1-\tau)/2: (1+\tau)/2=\varepsilon:\delta $), a triangle mode collapse region that touches both 
	points $(\varepsilon,\delta)$ and $(1-\delta,1-\varepsilon)$ at two of its edges  
		also touches the 45-degree line with a  $\tau$ shift at a vertex (left). When $\delta+\varepsilon>1$, the same happens when 
		$\tau = (\delta-\varepsilon)/(2-\delta-\varepsilon)$ (i.e.~$(1-\tau)/2: (1+\tau)/2=(1-\delta):(1-\varepsilon)$). 
		Hence, if $\tau>\max\{ (\delta-\varepsilon)/(\delta+\varepsilon),(\delta-\varepsilon)/(2-\delta-\varepsilon)\}$, then 
		the triangle region that does not include both orange points cannot touch the blue 45-degree line. 
		 }
	\label{fig:proof3_1}
\end{figure}

When $\delta+\varepsilon\leq1$ and $\tau>(\delta-\varepsilon)/(\delta+\varepsilon)$, 
no convex region $\cR(P,Q)$ can touch the 45-degree line at $\tau$ as shown below, and the feasible set is empty.
This follows from the fact that a triangle region passing through both $(\varepsilon,\delta)$ 
and $(1-\delta,1-\varepsilon)$ will have a total variation distance of $(\delta-\varepsilon)/(\delta+\varepsilon)$. 
Note that no $(\varepsilon,\delta)$ mode augmentation constraint translates into
the region not including the point $(1-\delta,1-\varepsilon)$.
We can see easily from Figure \ref{fig:proof3_1} that any total variation beyond that will require violating either the 
no-mode-collapse constraint or the no-mode-augmentation constraint. 
Similarly, when $\delta+\varepsilon>1$ and $\tau>(\delta-\varepsilon)/(2-\delta-\varepsilon)$, the feasible set is also empty. 
These two can be unified as $\tau>\max\{ (\delta-\varepsilon)/(\delta+\varepsilon),(\delta-\varepsilon)/(2-\delta-\varepsilon)\}$.

Suppose $\delta+\varepsilon\leq1$, 
and consider the intermediate regime when $\delta-\varepsilon \leq \tau \leq (\delta-\varepsilon)/(\delta+\varepsilon)$.  
In optimization \eqref{eq:opt_main3}, we consider only those pairs with no 
$(\varepsilon,\delta)$-mode collapse or $(\varepsilon,\delta)$-mode augmentation. 
It is simple to see that the inner bound does not change from optimization in \eqref{eq:opt_main1}. 
Let us denote a point where $\cR(P,Q)$ meets the blue line by 
the point $(1-\alpha'-\tau,1-\alpha')$ in the 2D plane, parametrized by 
$\alpha' \in [0,1-\tau]$. The $\cR(\alpha',\tau)$ defined in Figure \ref{fig:proof1_2} works in this case also. 
We only need a new outer bound. 

\begin{figure}[ht]
	\begin{center}
		\includegraphics[width=.35\textwidth]{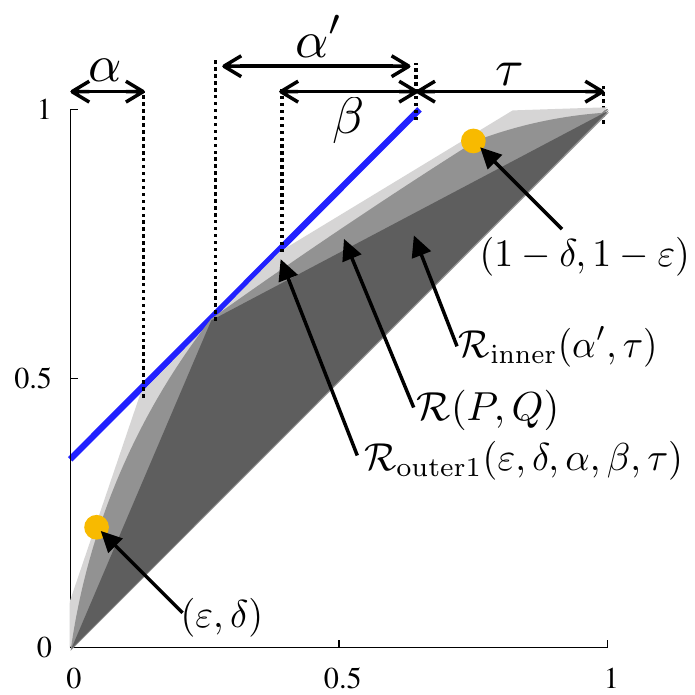}
		\put(-90,-6){$\varepsilon$}
		\put(-175,75){$\delta$}
		\includegraphics[width=.65\textwidth]{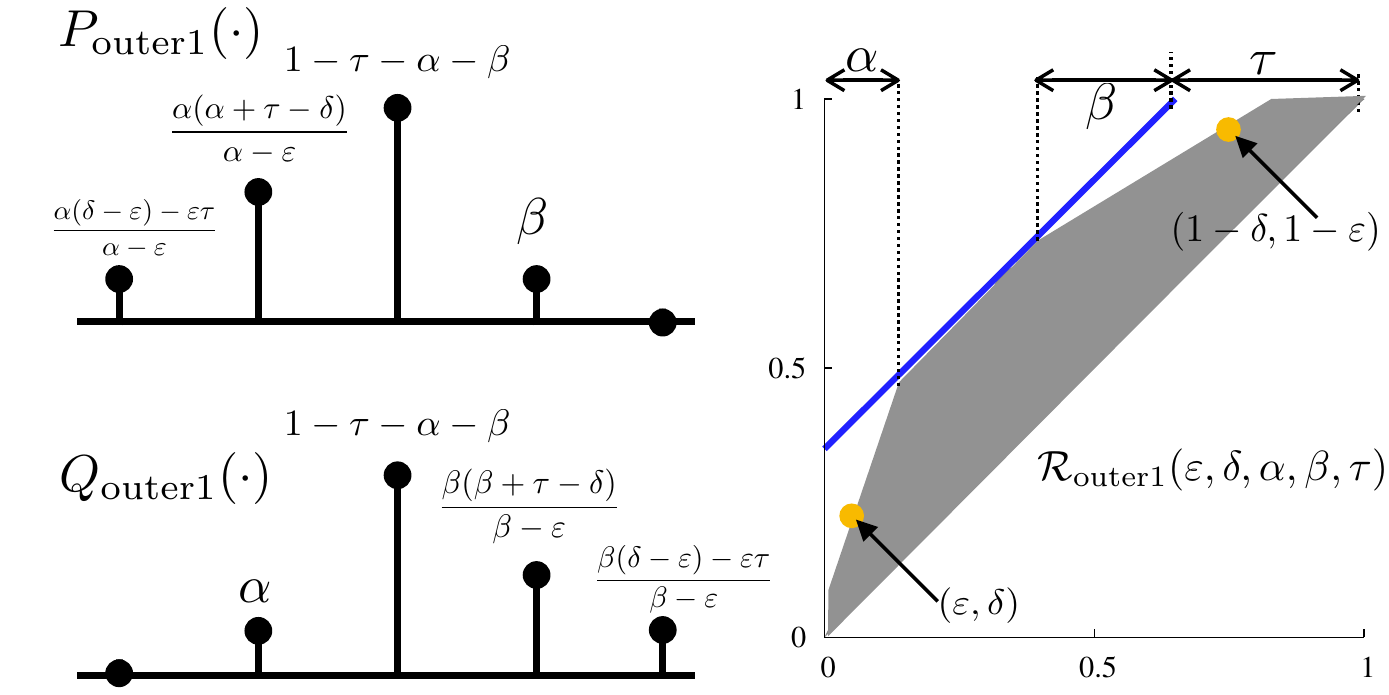}
		\put(-70,-6){$\varepsilon$}
		\put(-135,74){$\delta$}
	\end{center}
	\caption{For any pair $(P,Q)$ with no $(\varepsilon,\delta)$-mode collapse or no $(\varepsilon,\delta)$-mode augmentation, 
	the corresponding region $\cR(P,Q)$ is sandwiched 
	between $\cR_{\rm inner}(\alpha',\tau)$ and $\cR_{\rm outer1}(\varepsilon,\delta,\alpha,\beta,\tau)$ (left). 
	A canonical pair of distributions corresponding to $\cR_{\rm outer1}(\varepsilon,\delta,\alpha,\beta,\tau)$ (middle and right). 
	  }
	\label{fig:proof3_2}
\end{figure}

We construct an outer bound region, according to the following rule. 
We fit a hexagon where one edge is the 45-degree line passing through the origin, 
one edge is the vertical axis, 
one edge is the horizontal line passing through $(1,1)$, 
one edge is the 45-degree line with shift $\tau$ shown in blue in Figure \ref{fig:proof3_2}, 
and the remaining two edges include the two orange points, respectively, 
at $(\varepsilon,\delta)$ and $(1-\delta,1-\varepsilon)$. 
For any $\cR(P,Q)$ satisfying the constraints in \eqref{eq:opt_main3}, 
there exists at least one such hexagon that includes $\cR(P,Q)$. 
We parametrize the hexagon by $\alpha$ and $\beta$, 
where $(\alpha,\tau+\alpha)$ denotes the left-most point where the hexagon meets the blue line, 
and $(1-\tau-\beta,1-\beta)$ denotes the right-most point where the hexagon meets the blue line.

The additional constraint that $(P,Q)$ has no $(\varepsilon,\delta)$-mode collapse or 
$(\varepsilon,\delta)$-mode augmentation 
translates into a geometric constraint that 
we need to consider all regions $\cR(P,Q)$ that does not include the orange solid circle at point $(\varepsilon,\delta)$ and $(1-\delta,1-\varepsilon)$. 
Then, for any such $(P,Q)$, we can  sandwich the region $\cR(P,Q)$ 
between two regions $\cR_{\rm inner}$ and $\cR_{\rm outer1}$:  
\begin{eqnarray}
	\cR_{\rm inner}(\alpha',\tau) \;\;\subseteq\;\; \cR(P,Q) \;\;\subseteq\;\; \cR_{\rm outrer1}(\varepsilon,\delta,\alpha,\beta,\tau)\;,
	\label{eq:proof3_1}
\end{eqnarray}
where $\cR_{\rm inner}(\alpha,\tau)$ is defined as in Figure \ref{fig:proof1_2}. 

Let $(P_{\rm inner}(\alpha') , Q_{\rm inner}(\alpha',\tau))$ 
defined in \eqref{eq:pinner} and  \eqref{eq:qinner}, 
and $(P_{\rm outer1}(\varepsilon,\delta,\alpha,\beta,\tau) , Q_{\rm outer1}(\varepsilon,\delta,\alpha,\beta,\tau))$
denote the pairs of canonical distributions  
achieving the inner and outer regions exactly as shown in Figure \ref{fig:proof3_2}. 

By the preservation of dominance under product distributions in Remark \ref{rem:blackwell}, it follows from the dominance in \eqref{eq:proof3_1} that 
for any $(P,Q)$ there exist  $\alpha'$, $\alpha$, and $\beta$ such that 
\begin{eqnarray}
	\cR(P_{\rm inner}(\alpha')^m , Q_{\rm inner}(\alpha',\tau)^m) \;\;\subseteq\;\;  \cR(P^m,Q^m) \;\; \subseteq \;\;  \cR(P_{\rm outer1} (\varepsilon,\delta,\alpha,\beta,\tau)^m, Q_{\rm outer1}(\varepsilon,\delta,\alpha,\beta,\tau)^m)\;.
	\label{eq:proof3_2}
\end{eqnarray}
Due to the data processing inequality of mode collapse region in Remark \ref{rem:dp}, 
it follows that dominance of region implies dominance of total variation distances: 
\begin{align}
	&\min_{ \frac{\varepsilon\tau}{\delta-\varepsilon} \leq \alpha' \leq 1-\frac{\tau\delta}{\delta-\varepsilon}   } d_{\rm TV}(P_{\rm inner}(\alpha')^m,Q_{\rm inner}(\alpha',\tau)^m ) \;\;\leq \;\; 
	 d_{\rm TV}(P^m,Q^m)  \nonumber\\
	 &\hspace{4cm}\leq\;\; \max_{\alpha,\beta\geq \frac{\varepsilon\tau}{\delta-\varepsilon}, \alpha+\beta\leq 1-\tau} \;\; d_{\rm TV}(P_{\rm outer1}(\varepsilon,\delta,\alpha,\beta,\tau)^m,Q_{\rm outer1}(\varepsilon,\delta,\alpha,\beta,\tau)^m)\;.
\end{align}
The RHS and LHS of the above inequalities can be completely characterized by taking the $m$-th power 
of those canonical pairs of distributions, and then taking the respective minimum over $\alpha'$ and 
maximum over $\alpha$ and $\beta$. 
For the upper bound, this gives 
$U_1({\epsilon,\delta,\tau,m}) $ in \eqref{eq:defU1}, and 
for the lower bound this gives $L_2(\tau,m)$ in \eqref{eq:defL2}. 

\begin{figure}[ht]
	\begin{center}
		\includegraphics[width=.65\textwidth]{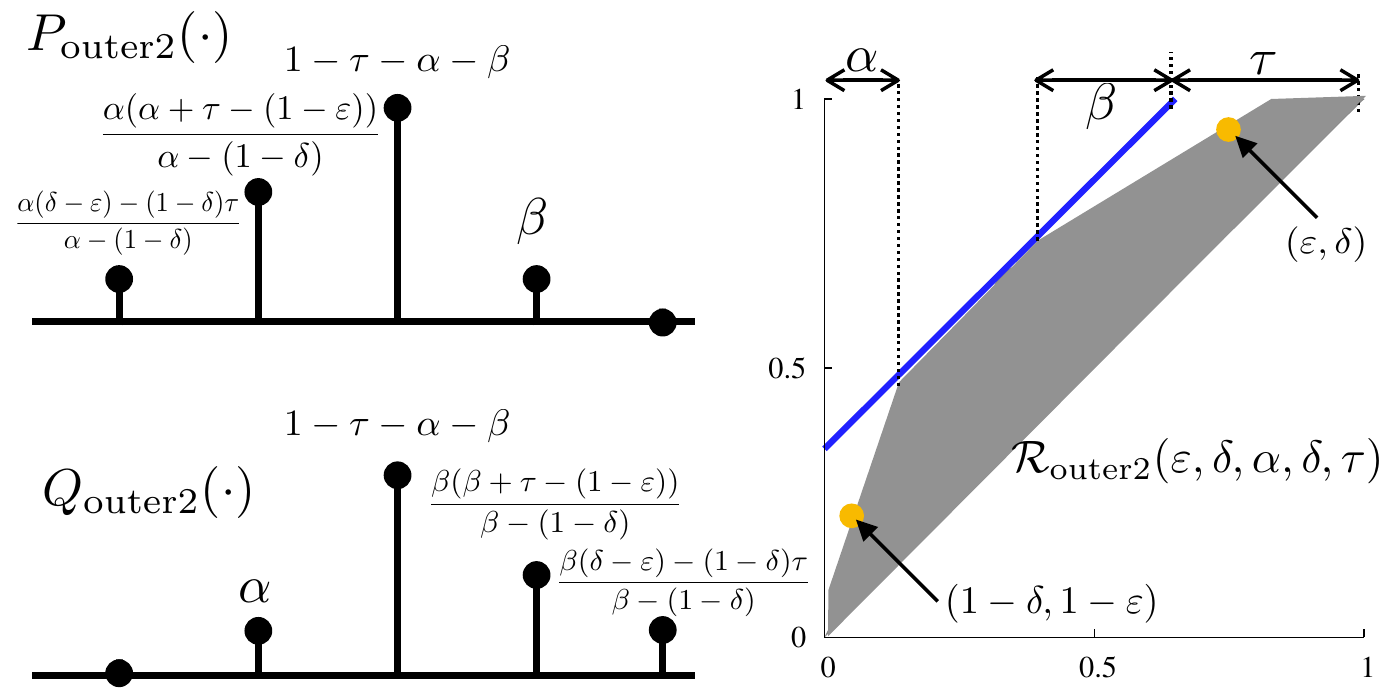}
		\put(-70,-6){$\varepsilon$}
		\put(-135,74){$\delta$}
	\end{center}
	\caption{
	A canonical pair of distributions corresponding to $\cR_{\rm outer2}(\varepsilon,\delta,\alpha,\beta,\tau)$. 
	  }
	\label{fig:proof3_3}
\end{figure}

Now, suppose $\delta+\varepsilon > 1$, 
and consider the intermediate regime when $\delta-\varepsilon \leq \tau \leq (\delta-\varepsilon)/(2-\delta-\varepsilon)$.  
We have a different outer bound $\cR_{\rm outer2}(\varepsilon,\delta,\alpha,\delta,\tau)$ as the role of $(\varepsilon,\delta)$ 
and $(1-\delta,1-\varepsilon)$ have switched. 
A similar analysis gives 
\begin{align}
	 d_{\rm TV}(P^m,Q^m)  
	 \;\; \leq\;\; \max_{\alpha,\beta\geq \frac{(1-\delta)\tau}{\delta-\varepsilon}, \alpha+\beta\leq 1-\tau} \;\; 
	 d_{\rm TV}(P_{\rm outer2}(\varepsilon,\delta,\alpha,\beta,\tau)^m,Q_{\rm outer2}(\varepsilon,\delta,\alpha,\beta,\tau)^m)\;,
\end{align}
where the canonical distributions are shown in Figure \ref{fig:proof3_3} and defined in 
\eqref{eq:pouter2} and \eqref{eq:qouter2}.
This gives $U_2({\epsilon,\delta,\tau,m}) $ in \eqref{eq:defU2}.
For the lower bound we only need to change the range of $\alpha$ we minimize over, which gives $L_3(\tau,m)$ in \eqref{eq:defL3}. 

\input{related.tex}

\section{Discussion}
\label{sec:discussion}
In this work, we propose a packing framework that theoretically and empirically mitigates mode collapse with low  overhead. 
Our analysis leads to several interesting open questions, 
including how to apply these analysis techniques to more general classes of loss functions 
such as  
 Jensen-Shannon divergence
and Wasserstein distances. 
This will complete the understanding of the superiority of our approach 
observed in experiments with JS divergence in Section \ref{sec:exp} and 
with Wasserstein distance in Section \ref{app:WGAN}.
Another important question is what packing architecture to use. 
For instance, a framework that provides permutation invariance may give better results 
such as 
graph neural networks
\cite{defferrard2016convolutional,thekumparampil2018attention,kipf2016semi}
 or deep sets \cite{zaheer2017deep}. 

\section*{Acknowledgement}
The authors would like to thank Sreeram Kannan  and Alex Dimakis for the initial discussions that lead to the inception of the packing idea,  
and Vyas Sekar for valuable discussions about GANs.
We thank  Srivastava Akash, Luke Metz, Tu Nguyen, and  Yingyu Liang 
for providing insights and/or the implementation details on their proposed architectures for VEEGAN \cite{SVR17}, Unrolled GAN \cite{MPP16}, D2GAN \cite{NLV17}, and MIX+GAN \cite{AGL17}. 

This work is supported by NSF awards CNS-1527754, CCF-1553452, and CCF-1705007, and RI-1815535  and Google Faculty Research Award.
This work used the Extreme Science and Engineering Discovery Environment (XSEDE), which is supported by National Science Foundation grant number OCI-1053575.  Specifically, it used the Bridges system, which is supported by NSF award number ACI-1445606, at the Pittsburgh Supercomputing Center (PSC).
This work is partially supported by 
the generous research credits on AWS cloud computing resources from Amazon. 


%
%
%
%
%
%
%
%
%
%
%
%
%
%
%

%

%
%
\bibliographystyle{plain}

\bibliography{_gan}



\end{document}

%% file: introduction.tex
\section{Introduction}
\label{sec:int}
Generative adversarial networks (GANs) are an innovative technique for training generative models 
to produce realistic examples from a data distribution \cite{GPM14}. 
Suppose we are given $N$ i.i.d. samples $X_1,\ldots, X_N$ from an unknown probability distribution $P$ over some high-dimensional space $\mathbb R^p$ (e.g., images).
The goal of generative modeling is to learn a model that enables us to produce samples from $P$ that are not in the training data.
Classical approaches to this problem typically search over a parametric family (e.g., a Gaussian mixture), 
and fit parameters to maximize the likelihood of the observed data. 
Such likelihood-based methods suffer from the curse of dimensionality in real-world datasets, such as images.  
Deep neural network-based generative models were proposed to cope with this problem \cite{KW13,Hin10,GPM14}. 
However, these modern generative models can be difficult to train, in large part because it is challenging to evaluate their likelihoods.  
Generative adversarial networks made a breakthrough in training such models, 
with an innovative training method that 
uses a minimax formulation whose solution 
is approximated by iteratively training two competing neural networks---hence the name ``adversarial networks".


GANs have attracted a great deal of interest recently. 
They are able to \emph{generate} realistic, crisp, and original examples of images \cite{GPM14,DCF15} and text \cite{yu2017seqgan}.
This is useful in image and video processing (e.g.~frame prediction \cite{VPT16}, 
image super-resolution \cite{LTH16}, and image-to-image translation \cite{IZZ16}), 
as well as dialogue systems or chatbots---applications where one may need realistic but artificially generated data.
Further, they implicitly learn a \emph{latent, low-dimensional representation} of arbitrary high-dimensional data.
Such embeddings have been hugely successful in the area of natural language processing (e.g.~word2vec \cite{MSC13}). 
GANs have the potential to provide such an unsupervised solution to learning representations that capture semantics of the domain to 
arbitrary data structures and applications.  
This can be used in various applications, such as 
image manipulation \cite{kingma2018glow} 
and defending against adversarial examples \cite{ilyas2017robust}.



\paragraph{Primer on GANs.}
Neural-network-based generative models are trained to map a (typically lower dimensional) random variable 
 $Z \in \mathbb R^d$ from a standard distribution (e.g.~spherical Gaussian) 
 to a domain of interest, like images. 
In this context, a \emph{generator} is a function $G: \mathbb R^d \rightarrow \mathbb R^p$,  
which is chosen from a rich class of parametric functions like deep neural networks.
In unsupervised 
generative modeling, 
one of the goals is to train the parameters of such a generator 
from unlabelled training data drawn independently from some real world dataset (such as celebrity faces in CelebA \cite{LLW15} or 
natural images from CIFAR-100 \cite{KH09}), 
in order to produce examples that are realistic but different from the training data.  

A breakthrough in training such generative models was achieved by the innovative idea of GANs  \cite{GPM14}. 
GANs train two neural networks: one for the generator $G(Z)$ and the other for a discriminator $D(X)$. 
These two neural networks play a dynamic minimax game against each other. 
An analogy provides the intuition behind this idea. 
The generator is acting as a forger trying to make fake coins (i.e., samples), 
and the discriminator is trying to detect which coins are fake and which are real. 
If these two parties are allowed to play against each other long enough, eventually 
both will become good. 
In particular, the generator will learn to 
produce coins that are indistinguishable from real coins (but preferably different from the training coins he was given). 

Concretely, we search for (the parameters of) neural networks $G$ and $D$ that optimize the following type of minimax objective: 
\begin{eqnarray}
G^* &\in& \arg \min_G \;\;  \max_D \;\; V(G,D) \nonumber \\
	&=& \arg \min_G \;\; \max_D \;\; \mathbb E_{X\sim P} [\log (D(X))] + \mathbb E_{Z\sim P_Z}[\log(1-D(G(Z)))] \;,
\label{eq:gan}
\end{eqnarray}
where $P$ is the distribution of the real data, and $P_Z$ is the distribution of the input code vector $Z$. 
Here $D$ is a function that tries to distinguish between real data and generated samples, whereas $G$ is the mapping from the latent space to the data space. 
Critically, \cite{GPM14} shows that the global optimum of \eqref{eq:gan} is achieved if and only if $P = Q$, where $Q$ is the generated distribution of $G(Z)$.
We refer to Section \ref{sec:theory} for a detailed discussion of this minimax formulation. 
The solution to the minimax problem \eqref{eq:gan} can be approximated by iteratively training two ``competing" neural networks, the generator $G$ and discriminator $D$. 
Each model can be updated individually by backpropagating the gradient of the loss function to each model's parameters.

\paragraph{Mode Collapse in GANs.}
One major challenge in training GAN is a phenomenon known as {\em mode collapse}, which collectively refers to the lack of diversity in 
generated samples.
One manifestation of mode collapse is the observation that GANs commonly miss some of the modes when trained on multimodal distributions. 
For instance, when trained on hand-written digits with ten modes, the generator might fail to produce some of the digits \cite{SGZ16}. 
Similarly, in tasks that translate a caption into an image, generators have been shown to generate series of nearly-identical images \cite{reed2016generative}. 
Mode collapse is believed to be related to the training instability of GANs---another major challenge in GANs. 

Several approaches have been proposed to fight mode collapse, e.g.~\cite{DKD16,DBP16,SVR17,SGZ16,MPP16,CLJ16,SW17,NLV17}. 
We discuss prior work on mode collapse in detail in Section \ref{sec:related}. 
Proposed solutions rely on modified architectures  \cite{DKD16,DBP16,SVR17,SGZ16}, loss functions \cite{CLJ16,ACB17}, and optimization algorithms \cite{MPP16}.
Although each of these proposed methods is empirically shown to help mitigate mode collapse, 
it is not well understood how the proposed changes  relate to mode collapse. 
Previously-proposed heuristics fall short of providing rigorous explanations on why they achieve empirical gains, 
especially when those gains are sensitive to architecture hyperparameters.  

\paragraph{Our Contributions.}
In this work, we examine GANs through the lens of \emph{binary hypothesis testing}.  
By viewing the discriminator as performing a binary hypothesis test on samples (i.e., whether they were drawn from distribution $P$ or $Q$), 
we can apply insights from classical  hypothesis testing literature to the analysis of GANs. 
In particular, this hypothesis-testing viewpoint provides a fresh perspective and understanding of GANs 
that leads to the following contributions:
\begin{enumerate}
	\item The first contribution is conceptual: we propose a formal mathematical definition of mode collapse that abstracts away the geometric properties of the underlying data distributions (see Section \ref{sec:definition}). 
	This definition is closely related to the notions of false alarm and missed detection in binary hypothesis testing (see Section \ref{sec:region}). 
	Given this definition, we provide a new interpretation of the pair of distributions $(P,Q)$ as 
	a two-dimensional  region called the \emph{mode collapse region}, where $P$ is the true data distribution and $Q$ the generated one.  
	The mode collapse region provides new insights on 
	how to reason about the relationship between those two distributions 
	(see Section \ref{sec:definition}). 

%
	\item The second  contribution is analytical:  through the lens of hypothesis testing and mode collapse regions, we show
	 that 
	if the discriminator is allowed to see samples from the $m$-th order product distributions 
	$P^m$ and $Q^m$ instead of the usual target distribution $P$ and generator distribution $Q$, 
	then the corresponding loss when training the generator naturally penalizes generator distributions with strong mode collapse (see Section \ref{sec:productregion}). 
	Hence, a generator trained with this type of discriminator will be encouraged to choose a distribution that exhibits less mode collapse. 
	The {\em region} interpretation of mode collapse and corresponding data processing inequalities 
	provide the analysis tools that 
	allows us to prove strong and sharp results with simple proofs (see Section \ref{sec:proof}). 
	 This follows a long tradition in information theory literature (e.g.~\cite{Sta59,DCT91,CT92,Z98,VG06,LV07,KOV14,KOV15,KOV17}) 
	where operational interpretations of mutual information and corresponding data processing inequalities have given rise to simple proofs of 
	strong technical results.

	\item The third contribution is algorithmic: based on the insights from the region interpretation of mode collapse, we propose a new GAN framework to mitigate mode collapse, which we  call {PacGAN}. 
	PacGAN can be applied to any existing GAN, and it requires only a small modification to the  discriminator architecture (see Section \ref{sec:pacgan}). 
	The key idea is to pass $m$ ``packed" or concatenated samples to the discriminator, 
	which are jointly  classified  as either real or
	generated. 
	This allows the discriminator to do 
	binary hypothesis testing based on the product distributions $(P^m,Q^m)$,  
	which naturally penalizes mode collapse (as we show in Section \ref{sec:productregion}).  
	We demonstrate  on benchmark datasets that 
	PacGAN significantly improves upon competing approaches  in mitigating mode collapse (see Section \ref{sec:exp}). 
	Further, unlike existing approaches on jointly using multiple samples, e.g.~\cite{SGZ16}, 
	PacGAN requires no hyper parameter tuning and  
	incurs only a slight overhead in the architecture.  
	

\end{enumerate}

%
%
%


%

\paragraph{Outline.}
This paper is structured as follows:
 we present the PacGAN framework in Section \ref{sec:pacgan}, and evaluate it empirically according to the metrics and experiments proposed in prior work (Section \ref{sec:exp}). 
In Section \ref{sec:theory}, we propose a \emph{new} definition of mode collapse, and provide 
analyses showing that PacGAN mitigates mode collapse. 
The proofs of the main results are provided in Section \ref{sec:proof}. 
Finally, we describe in greater detail the related work on GANs in general and mode collapse in particular in Section \ref{sec:related}.


%% file: section2-gf.tex
\section{PacGAN: A novel framework for mitigating mode collapse}
\label{sec:pacgan}

We propose a new framework for mitigating mode collapse in GANs. 
We start with an arbitrary existing GAN\footnote{For a list of some popular GANs, we refer to the GAN zoo: https://github.com/hindupuravinash/the-gan-zoo}, which is typically defined by a generator architecture, a discriminator architecture,  
and a loss function. 
Let us call this triplet the \emph{mother architecture}.

The PacGAN framework maintains the same generator architecture and loss function as the mother architecture, 
and makes a slight change only to the discriminator. 
That is, instead of using a discriminator $D(X)$  
that maps a single (either from real data or from the generator) to a (soft) label,  
we use an \emph{augmented} discriminator $D(X_1,X_2,\ldots,X_m)$ that 
maps $m$ samples, 
jointly coming from either real data or the generator, to a single (soft) label. 
These $m$ samples are drawn independently 
 from the same distribution---either  real (jointly labelled as $Y=1$) or generated (jointly labelled as $Y=0$). 
We refer to the concatenation of samples with the same label as {\em packing}, 
the resulting concatenated discriminator as a {\em packed discriminator}, 
and the number $m$ of concatenated samples as the {\em degree of packing}. 
We call this approach a framework instead of an architecture,
 because the proposed approach of packing can be applied to 
any existing GAN, using any architecture and any loss function, as long as it uses a discriminator of the form $D(X)$ 
that classifies a single input sample.  

We propose the nomenclature  ``Pac(X)($m$)'' where 
(X) is the name of the mother architecture, and 
($m$) is an integer that refers to how many samples are packed together as an input to the discriminator.
For example, if we take an original GAN and feed the discriminator three packed samples as input, 
we call this ``PacGAN3''. 
If we take the celebrated DCGAN \cite{RMC15} and feed the discriminator four packed samples as input, 
we call  this ``PacDCGAN4''. 
When we refer to the generic principle of packing, we use PacGAN without an subsequent integer.

\begin{figure}[ht]
\begin{center}
  \includegraphics[width=0.5\textwidth]{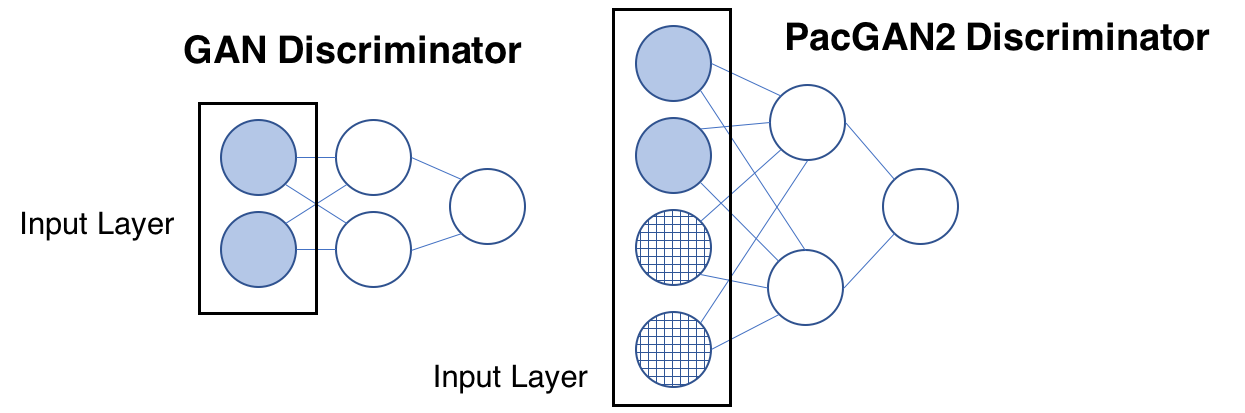}
\end{center}
  \caption{PacGAN(m) augments the input layer by a factor of m. 
  The number of edges between the first two layers are increased accordingly to preserve the connectivity of the mother architecture (typically fully-connected).
  Packed samples are fed to the input layer in a concatenated fashion; the grid-patterned nodes represent input nodes for the second input sample.}
  \label{fig:pac}
  \end{figure}

\vspace{0.1in}
\noindent
{\bf How to pack a discriminator.}  
Note that there are many ways to change the discriminator architecture to accept packed input samples.
We propose to keep all hidden layers of the discriminator exactly the same as the mother architecture, 
and only increase the number of nodes in the input layer by a factor of $m$. 
For example, in Figure \ref{fig:pac}, suppose we start with a mother architecture in which the discriminator is a fully-connected feed-forward network. 
Here, each sample $X$ lies in a space of dimension $p=2$, so the input layer has two nodes.
Now, under PacGAN2, we would multiply the size of the input layer by the packing degree (in this case, two), 
and the connections to the first hidden layer would be adjusted so that the first two layers remain fully-connected, as in the mother architecture.
The grid-patterned nodes in Figure \ref{fig:pac} represent input nodes for the second sample. 

Similarly, when packing a DCGAN, which uses convolutional neural networks for both the generator and the discriminator, 
we simply stack the images into a tensor of depth $m$.
For instance, the discriminator for PacDCGAN5 on the MNIST dataset of handwritten images \cite{mnist} would take an input of size $28\times28\times 5$, since each individual black-and-white MNIST image is $28\times 28$ pixels. 
Only the input layer and 
the number of weights in the corresponding first convolutional layer will increase in depth by a factor of five. 
By modifying only the input dimension and fixing the number of hidden and output nodes in the discriminator, 
we can focus purely on the effects of  {\em packing} in our numerical experiments in Section \ref{sec:exp}. 

\bigskip
\noindent
{\bf How to train a packed discriminator.}  
Just as in standard GANs, we train the packed discriminator with a bag of samples from the real data and the generator. 
However, each minibatch in the stochastic gradient descent now consists of 
 {\em packed} samples. 
 Each packed sample is of the form  
$(X_1,X_2,\ldots,X_m,Y)$, where the label is $Y=1$ for real data and $Y=0$ for generated data, 
and the $m$ independent samples from either class are jointly treated as a single, higher-dimensional feature $(X_1,\ldots,X_m)$. 
The discriminator learns to classify $m$ packed samples jointly. 
Intuitively,  packing helps the discriminator detect mode collapse because lack of diversity is more obvious in a set of samples than in a single sample. 
Fundamentally, packing allows the discriminator to observe samples from \emph{product distributions}, which highlight mode collapse more clearly than unmodified data and generator distributions. 
We make this statement precise in Section \ref{sec:theory}.


Notice that the computational overhead of PacGAN training is marginal, since only the input layer of the discriminator gains new parameters.
Furthermore, 
we keep all training hyperparameters  
identical to the mother architecture, 
including the stochastic gradient descent minibatch size, weight decay, learning rate, and the number of training epochs.
This is in contrast with other approaches for  mitigating mode collapse that 
require significant computational overhead and/or delicate hyperparameter selection  \cite{DBP16,DKD16,SGZ16,SVR17,MPP16}.

\paragraph{Computational complexity.} 
The exact computational complexity overhead of PacGAN (compared to GANs) is architecture-dependent, but can be computed in a straightforward manner.
For example, consider a discriminator with $w$ fully-connected layers, each containing $g$ nodes. 
Since the discriminator has a binary output, the $(w+1)$th layer has a single node, and is fully connected to the previous layer.
We seek the computational complexity of a single minibatch parameter update, where each minibatch contains $r$ samples. 
Backpropagation in such a network is dominated by the matrix-vector multiplication in each hidden layer, which has complexity $O(g^2)$ per input sample, assuming a naive implementation.
Hence the overall minibatch update complexity is $O(rwg^2)$.
Now suppose the input layer is expanded by a factor of $m$.
If we keep the same number of minibatch elements, the per-minibatch cost grows to $O((w+m)rg^2)$. 
We find that in practice, even $m=2$ or $m=3$ give good results. 

%% file: experiment.tex
\section{Experiments}
\label{sec:exp} 

On standard benchmark datasets, we compare PacGAN to several baseline GAN architectures, 
some of which are explicitly proposed to mitigate mode collapse: GAN \cite{GPM14},
minibatch discrimination (MD) \cite{SGZ16},  DCGAN \cite{RMC15}, VEEGAN \cite{SVR17}, Unrolled GANs \cite{MPP16}, and ALI \cite{DBP16}. We also implicitly compare against BIGAN \cite{DKD16}, which is conceptually identical to ALI.
To isolate the effects of packing, we make minimal choices in the architecture and hyperparameters of our packing implementation.
For each experiment, we evaluate packing by taking a standard, baseline GAN implementation that was \emph{not} designed to prevent mode collapse, and adding packing in the discriminator.
In particular, our goal for this section is to reproduce experiments from existing literature, 
apply the packing framework to the simplest GAN among those in the baseline, 
and showcase how packing affects the performance. 
All of our experiments are available at \url{https://github.com/fjxmlzn/PacGAN}, and were run with support from \cite{towns2014xsede,nystrom2015bridges}.

\paragraph{Metrics.} 
For consistency with prior work, we measure several previously-used metrics. 
On datasets with clear, known modes (e.g., Gaussian mixtures, labelled datasets), prior papers have counted the \emph{number of modes} that are produced by a generator \cite{DKD16,MPP16,SVR17}. 
In labelled datasets, this number can be evaluated using a third-party trained classifier that classifies the generated samples \cite{SVR17}. 
In Gaussian Mixture Models (GMMs), 
for example  in \cite{SVR17}, 
a mode is considered lost if there is no sample in the generated test data 
within $x$ standard deviations from the center of that mode. 
In \cite{SVR17}, $x$ is set to be three for 2D-ring and 2D-grid. 
A second metric used in \cite{SVR17} is the \emph{number of high-quality samples}, 
which is the proportion of the samples that are within $x$ standard deviation from the center of a mode. 
Finally, the \emph{reverse Kullback-Leibler divergence} over the modes has been used to measure the quality of mode collapse as follows. 
Each of the generated test samples is assigned to its closest mode; this induces an empirical, discrete distribution 
with an alphabet size equal to the number of observed modes in the generated samples.
A similar induced discrete distribution is computed from the real data samples. 
The reverse KL divergence between the induced distribution from generated samples and the induced distribution from the real samples is used as a metric. 
Each of these three metrics has shortcomings---for example, the number of observed modes does not account for class imbalance among generated modes, and all of these metrics only work for datasets with known modes. 
Defining an appropriate metric for  evaluating GANs is an active research topic \cite{theis2015note,WBS16,SSM17}. 

\paragraph{Datasets.} 
We use a number of synthetic and real datasets for our experiments, all of which have been studied or proposed in prior work. 
The \textbf{2D-ring} \cite{SVR17} is a mixture of eight two-dimensional spherical Gaussians with means $( \cos((2\pi/8)i), \sin((2\pi/8)i) )$ and variances $10^{-4}$ in each dimension for $i\in\{1,\ldots,8\}$.
The \textbf{2D-grid} \cite{SVR17} is a mixture of 25 two-dimensional spherical Gaussians with means $(-4+2i,-4+2j)$ and variances $0.0025$ in each dimension for $i,j\in\{0,1,2,3,4\}$.


To examine real data, we use the MNIST dataset \cite{mnist}, which consists of 70,000 images of handwritten digits, each $28\times 28$ pixels. 
Unmodified, this dataset has 10 modes, one for each digit. 
As done in Mode-regularized GANs \cite{CLJ16}, Unrolled GANs \cite{MPP16}, and VEEGAN \cite{SVR17}, we augment the number of modes by \emph{stacking} the images. 
That is, we generate a new dataset of 128,000 images, in which each image consists of three randomly-selected MNIST images that are stacked into a $28\times 28\times 3$ image in RGB. 
This new dataset has (with high probability) $1000=10\times 10\times 10$ modes. 
We refer to this as the \textbf{stacked MNIST} dataset.

Finally, we include experiments on the \textbf{CelebA} dataset, which is a collection of 200,000 facial images of celebrities \cite{liu2015faceattributes}. 
We use the aligned and cropped version, in which images are 218x178 pixels.
Although CelebA images are annotated with features (e.g. `eyeglasses', `wearing hat'), we do not use these labels in our experiments.


\subsection{Synthetic data experiments}
Our first experiment evaluates the number of modes and the number of high-quality samples for the 2D-ring and the 2D-grid. 
Results are reported in Table \ref{tbl:veegan1}. 
The first two rows are the GAN baseline and ALI, respectively,
followed by PacGAN with a packing factor of 2, 3, and 4.
The hyper-parameters, network architecture, and loss function for GAN and ALI are exactly reproduced from ALI's code\footnote{https://github.com/IshmaelBelghazi/ALI}. 
All PacGANs are directly modified from an existing GAN implementation, without any further hyper-parameter tuning.
The details are described below.



\begin{figure}[ht]
	\begin{center}
	\includegraphics[width=.3\textwidth]{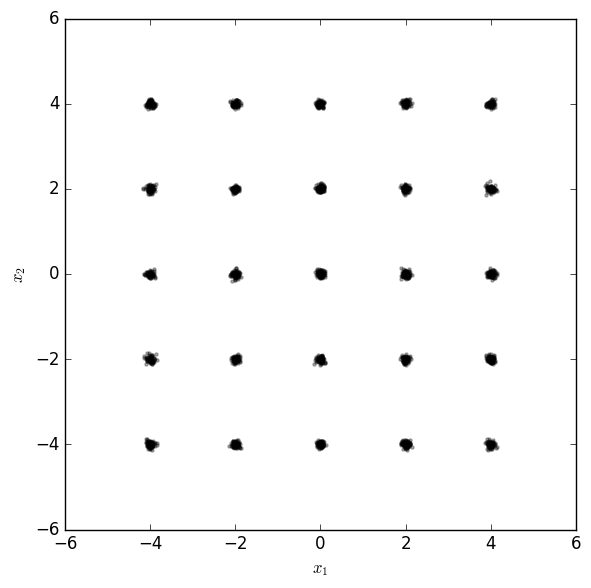}
	\put(-108,145){Target distribution}
	\includegraphics[width=.3\textwidth]{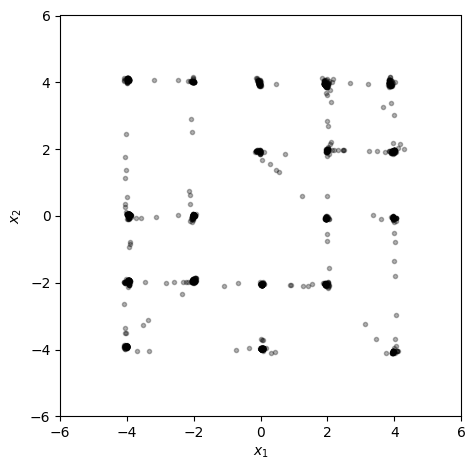}
	\put(-80,145){GAN}
	\includegraphics[width=.3\textwidth]{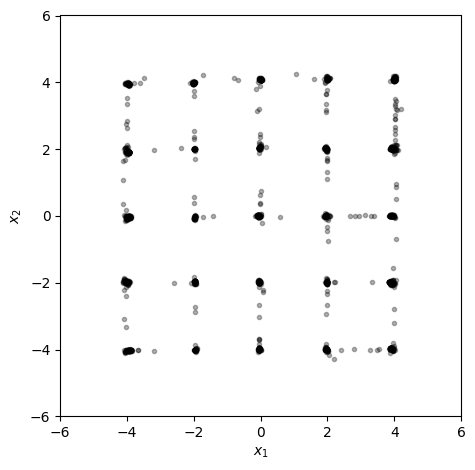}
	\put(-90,145){PacGAN2}
	\end{center}
	\caption{Scatter plot of the 2D samples from the true distribution (left) of 2D-grid and 
	the learned generators using GAN (middle) and PacGAN2 (right). PacGAN2 captures all of the 25 modes. }
	\label{fig:scatter}
\end{figure}


\bigskip\noindent
{\bf Architecture and hyper-parameters.}
All of the GANs in this experiment use the same generator architecture.
There are four hidden layers, each of which has 400 units with ReLU activation, 
trained with batch normalization \cite{IS15}.
The input noise is a two dimensional spherical Gaussian with zero mean and unit variance. 
{All of the GANs in this experiment use the same discriminator, except that the input dimension is different.}
{The discriminator has three hidden layers, 
with 200 units per hidden layer.} 
The hidden layers use LinearMaxout with 5 maxout pieces, and  
no batch normalization is used in the discriminator.  
{In addition to a generator and discriminator, ALI also has a third component, called an \emph{encoder}; 
we only used the encoder to evaluate ALI, but did not include the encoder in our PacGANs.} 
MD's discriminator is the same as GAN's discriminator, except that a minibatch discrimination layer is added before the output layer. The implementation of minibatch discrimination layer in this and all following experiments is based on the standard implementation\footnote{\url{https://github.com/openai/improved-gan}}.

We train each GAN with 100,000 total samples, and a mini-batch size of 100 samples; training is run for 400 epochs. 
The discriminator's loss function is  $\log(1+\exp(-D(\text{real data}))) + \log(1+\exp(D(\text{generated data})))$.  
The generator's loss function is $ \log(1+\exp(D(\text{real data}))) + \log(1+\exp(-D(\text{generated data})))$. 
Adam \cite{KB14} stochastic gradient descent is applied with 
the generator weights and the discriminator weights updated once per mini-batch. 
At testing, we use 2500 samples from the learned generator for evaluation. 
Each metric is evaluated and averaged over 10 trials.

\begin{table}[ht]
	\begin{center}
  	\begin{tabular}{ l  r  r r c r  r r c }
    		\hline
		& \multicolumn{3}{c}{2D-ring}  && \multicolumn{3}{c}{2D-grid}   \\ \cline{2-4} \cline{6-8} 
    		 & Modes &  high quality & reverse KL && Modes &  high quality  & reverse KL\\ 
		 & (Max 8) & samples &&& (Max 25) & samples &
		 \\ \hline
    		GAN \cite{GPM14} & 6.3$\pm$0.5 & 98.2$\pm$0.2 \% 		& 0.45$\pm$0.09&& 17.3$\pm$0.8& 94.8$\pm$0.7 \%	& 0.70$\pm$0.07\\
    		ALI  \cite{DBP16} & 6.6$\pm$0.3 &  97.6$\pm$0.4 \% &0.36$\pm$0.04			&& 24.1$\pm$0.4 & 95.7$\pm$0.6 \%	& 0.14$\pm$0.03\\
    		Minibatch Disc. \cite{SGZ16} & 4.3$\pm$0.8 & 36.6$\pm$8.8 \% & 1.93$\pm$0.11 && 23.8$\pm$0.5 & 79.9$\pm$3.2 \% & 0.17$\pm$0.03\\
		\hline
		PacGAN2 (ours) 	& 7.9$\pm$0.1& 95.6$\pm$2.0 \%	& 0.07$\pm$0.03				&& 23.8$\pm$0.7&91.3$\pm$0.8 \% &0.13$\pm$0.04	\\
		PacGAN3 (ours) 	& 7.8$\pm$0.1& 97.7$\pm$0.3 \%	&0.10$\pm$0.02				&& 24.6$\pm$0.4&94.2$\pm$0.4 \% &0.06$\pm$0.02	\\
		PacGAN4 (ours) 	& 7.8$\pm$0.1& 95.9$\pm$1.4 \%	& 0.07$\pm$0.02				&& 24.8$\pm$0.2&93.6$\pm$0.6 \%	&0.04$\pm$0.01	\\
    		\hline
  	\end{tabular}
	\end{center}
	\caption{Two measures of mode collapse proposed in \cite{SVR17} 
	for two synthetic mixtures of Gaussians: number of modes captured by the generator and percentage of high quality samples, as well as reverse KL.
	Our results are averaged over 10 trials shown with the standard error. 
	We note that 2 trials of MD in 2D-ring dataset cover no mode, which makes reverse KL intractable. This reverse KL entry is averaged over the other 8 trails.
	}
	\label{tbl:veegan1}
\end{table}

\paragraph{Results.}
Table \ref{tbl:veegan1} shows that PacGAN outperforms or matches the baseline schemes in all three metrics. 
On the 2D grid dataset, increasing the packing degree $m$ appears to increase the average number of modes recovered, as expected.
On the 2D ring dataset, PacGAN2 is able to recover almost all the modes, so further packing seems to provide little extra benefit.
The benefits of packing can be evaluated by comparing the GAN in the first row (which is the mother architecture) and PacGANs in the last rows. 
The simple change of packing the mother architecture appears to make a significant difference in performance,
and the overhead associated with implementing these changes is minimal compared to the baselines \cite{DBP16,MPP16,SVR17}. 

Note that maximizing the number of high-quality samples is not necessarily indicative of a good generative model. 
First, we expect some fraction of probability mass to lie outside the ``high-quality" boundary, and that fraction increases with the dimensionality of the dataset. 
For reference, we find empirically that the expected fraction of high-quality samples in the true data distribution for the 2D ring and grid are both 98.9\%, 
which corresponds to the theoretical ratio for a single 2D Gaussian. 
These values are higher than the fractions found by PacGAN, indicating room for improvement.
However, a generative model could output 100\% high-quality points by learning very few modes 
(as reported in \cite{SVR17}). 


We also observe that in terms of mode coverage MD performs well in 2D-grid dataset but badly in 2D-ring dataset, even with completely the same architecture. This suggests that MD is sensitive to experiment settings. In terms of high quality samples, MD performs even worse than GAN baseline in both datasets.

We wish to highlight that our goal is not to compete with the baselines of ALI or other state-of-the-art methods,
but to showcase the improvement that can be obtained with packing. 
In this spirit, we can easily apply our framework to other baselines and test ``PacALI'', ``PacUnrolledGAN'', and ``PacVEEGAN''. 
In fact, we expect that most GAN architectures can be packed to  improve sample quality.  
However, for these benchmark tests, we see that packing the simplest GAN is sufficient.

\subsubsection{The effect of parameter size: 2D Grid}
The way we implement packing introduces a potential confounding variable:  the number of parameters in the discriminator.
That is, our packed architectures have more discriminator nodes (and hence more discriminator parameters) than the mother architecture, which could artificially inflate our experimental results by giving the discriminator greater capacity.
Our next experiment aims to compare this effect to the effect of packing, again on the 2D grid dataset.
We evaluate three metrics---fraction of high-quality samples, number of modes recovered, and reverse KL divergence---for ALI, GAN, MD and PacGAN, while varying the number of \emph{total} parameters in each architecture (discriminator and encoder if one exists). 

\bigskip\noindent
{\bf Architecture and hyper-parameters.}
Compared to the previous experiment, this experiment introduces only one architectural difference, which stems from varying the number of total parameters. 
We keep the generators and encoders (if one exists) identical across experiments, and vary only the number of total parameters in the discriminator.
Recall that in our previous 2D Grid experiment, there were 200 nodes per hidden layer. 
In this experiment, we keep the input and output layers identical to our previous experiment, but alter the number of nodes per hidden layer in the discriminator. 
For each experimental setting, each hidden layer of the discriminator has the same number of hidden nodes, drawn from the set $\{50,100,150,200,250\}$. 
This hidden layer size determines the total number of parameters in the architecture, so each GAN variant is evaluated for five different parameter counts. 
There may be more sophisticated ways to evaluate the effects of discriminator and encoder size; our approach only captures the effect of hidden layer width. 

\begin{figure}
\begin{minipage}[b]{0.32\linewidth}
\centering
\includegraphics[width=\textwidth]{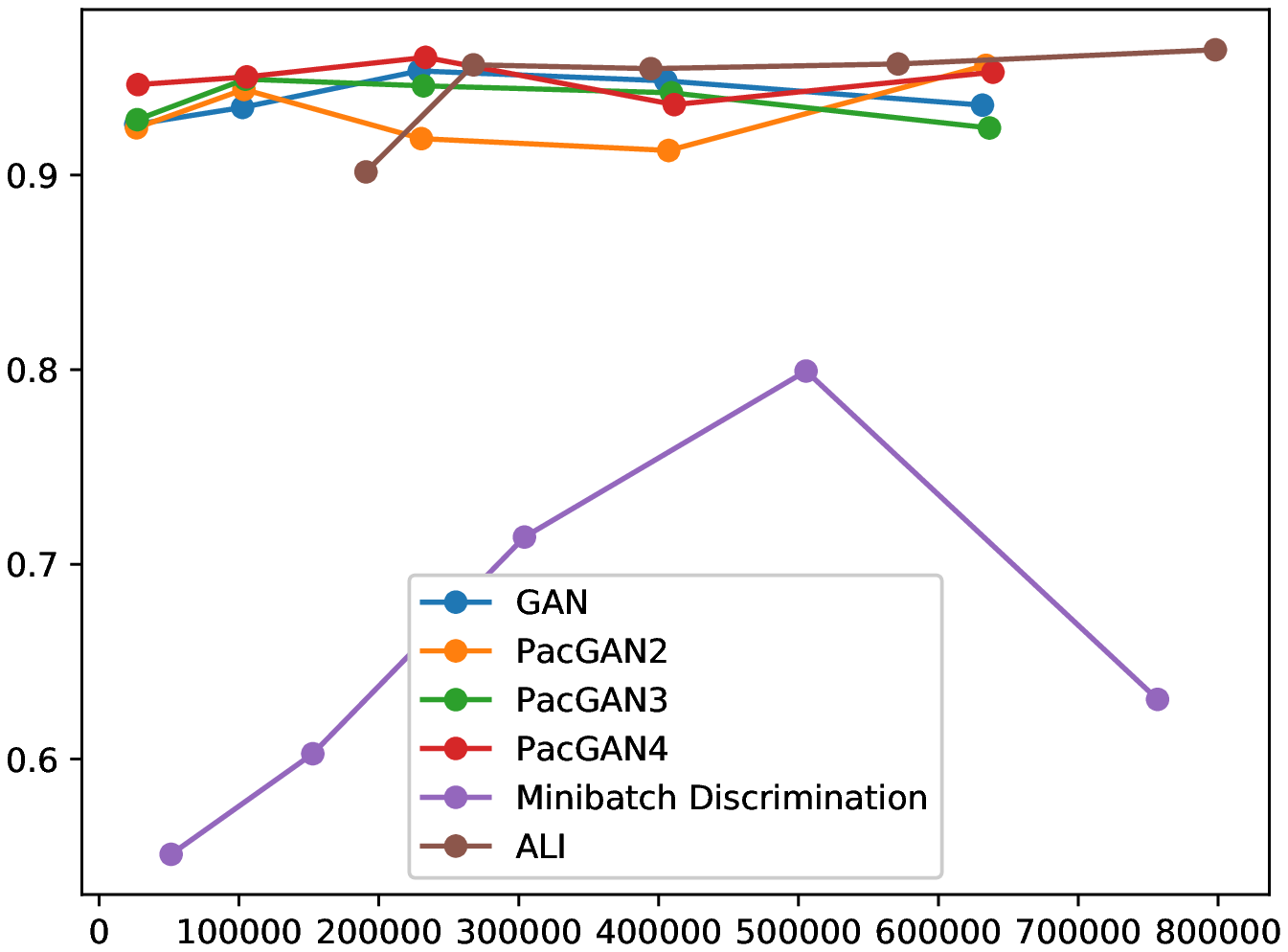}
\put(-110,-10){Parameter Count}
\caption{High-quality samples (out of 1, higher is better)}
\label{fig:params_quality}
\end{minipage}
\hspace{0.05in}
\begin{minipage}[b]{0.32\linewidth}
\centering
\includegraphics[width=\textwidth]{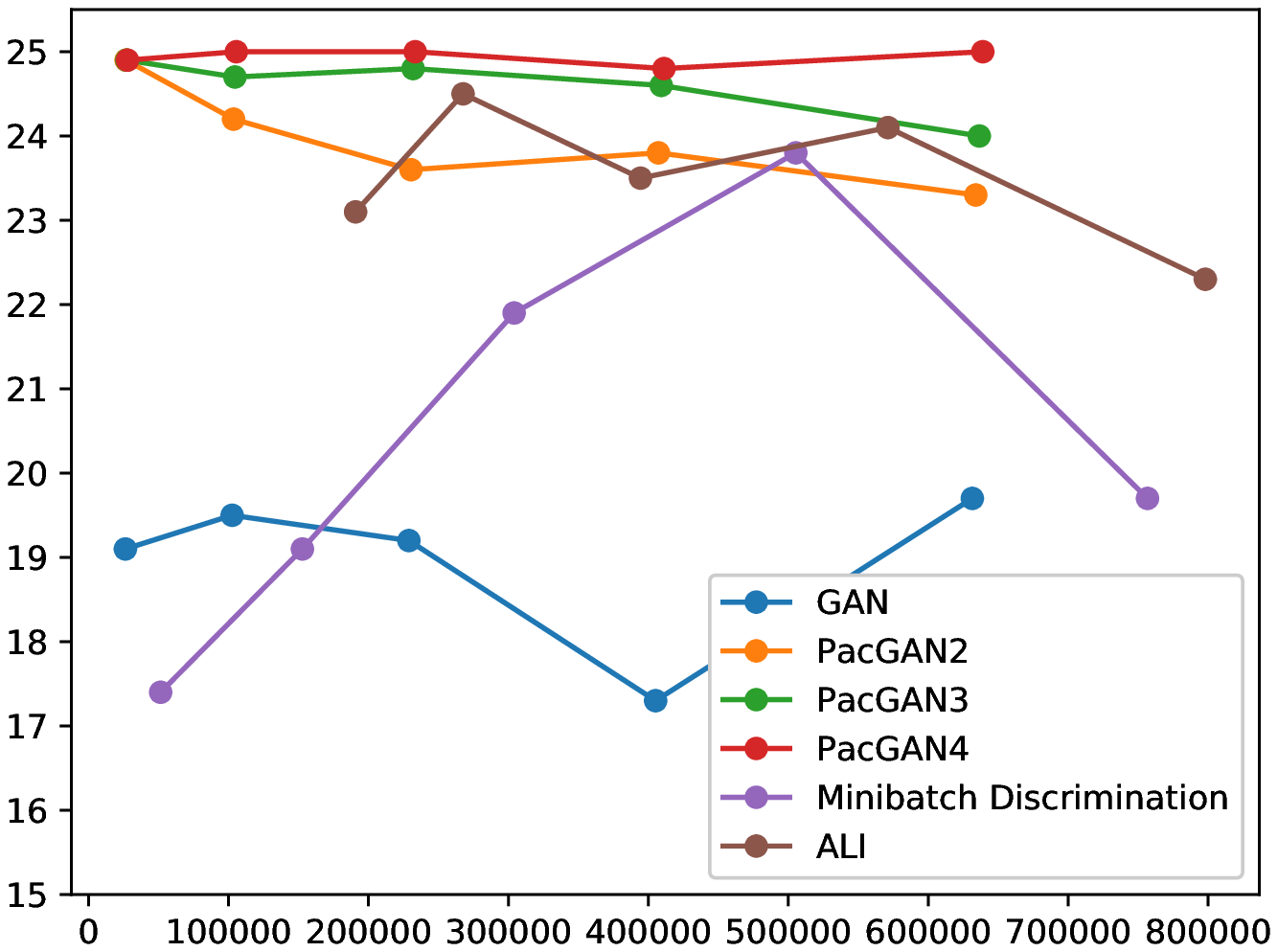}
\put(-110,-10){Parameter Count}
\caption{Modes recovered (max 25, higher is better)}
\label{fig:params_modes}
\end{minipage}
\hspace{0.05in}
\begin{minipage}[b]{0.32\linewidth}
\centering
\includegraphics[width=\textwidth]{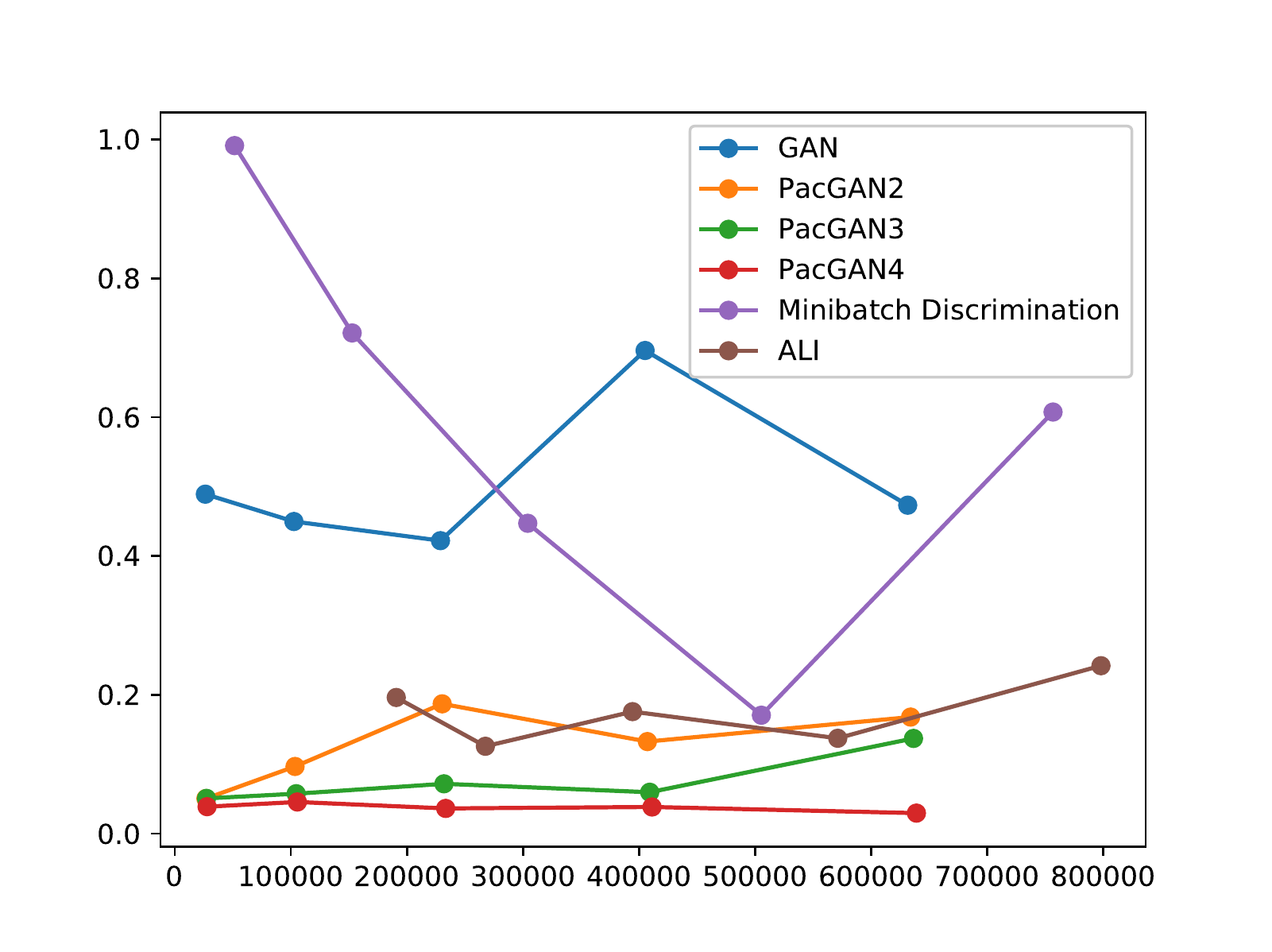}
\put(-110,-10){Parameter Count}
\caption{Reverse KL divergence (lower is better)}
\label{fig:params_kl}
\end{minipage}
\end{figure}

\paragraph{Results.}
Figures \ref{fig:params_quality}, \ref{fig:params_modes}, and \ref{fig:params_kl} show the high-quality samples, reverse KL-divergence, and number of modes recovered, respectively. 
The horizontal axis in each figure captures the total number of parameters in the discriminator and encoder (only ALI has an encoder, which is counted toward its parameter budget). 
Each data point is averaged over 10 trials, as before. 

We make a few observations: first, the number of parameters seems to have a small effect on the evaluated metrics. 
Despite varying the number of parameters by an order of magnitude, we do not see significant evidence of the metrics improving with the number of parameters, for any architecture.
This suggests that the advantages of PacGAN and ALI compared to GAN do not stem from having more parameters. 
Our second observation is that packing seems to significantly increase the number of modes recovered and the reverse KL divergence; there is a distinct improvement from GAN to PacGAN2 to PacGAN3 to PacGAN4. 
These effects are expected, as both metrics (modes recovered and reverse KL divergence) are proxies for mode collapse. 
Along these metrics, ALI seems to perform similarly to PacGAN2.
Third, packing does not appear to affect the fraction of high-quality samples.
One explanation may be that improving diversity does not necessary improve sample quality. 
We want to highlight that the standard error of these experiments is large; more trials are needed, but these preliminary results suggest that the benefits of packing do not primarily stem from having more parameters in the discriminator.  
For MD, the metrics first improve and then degrade with the number of parameters.
We suspect that this may because MD is very sensitive to experiment settings, as the same architecture of MD has very different performance on 2d-grid and 2d-ring dataset (Table~\ref{tbl:veegan1}).

\subsection{Stacked MNIST experiment}
In our next experiments, we evaluate mode collapse on the stacked MNIST dataset (described at the beginning of Section \ref{sec:exp}).    
These experiments are direct comparisons to analogous experiments in VEEGAN \cite{SVR17} and Unrolled GANs \cite{MPP16}. 
For these evaluations, we generate samples from the generator. 
Each of the three channels in each sample is classified by 
a  pre-trained third-party MNIST classifier, and the resulting three digits 
determine which of the $1,000$ modes the sample belongs to.
We measure the number of modes captured, as well as the KL divergence
between the generated distribution over modes and the expected true one (i.e., a uniform distribution over the 1,000 modes).


\begin{figure}[ht]
	\begin{center}
	\includegraphics[width=.3\textwidth]{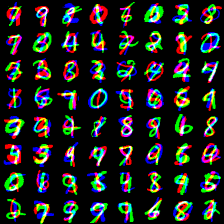}
	\put(-108,145){Target distribution}
	\hspace{0.1cm}
	\includegraphics[width=.3\textwidth]{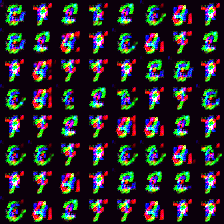}
	\put(-93,145){DCGAN}
	\hspace{0.1cm}
	\includegraphics[width=.3\textwidth]{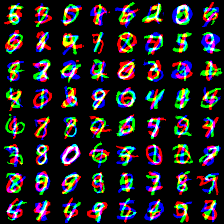}
	\put(-106,145){PacDCGAN2}
	\end{center}
	\caption{True distribution (left), DCGAN generated samples (middle), and PacDCGAN2 generated samples (right) from the stacked-MNIST dataset show PacDCGAN2 captures more diversity while producing sharper images.}
	\label{fig:stacked-mnist}
\end{figure}



\subsubsection{VEEGAN \cite{SVR17} experiment} \label{sec:veegan mnist}
In this experiment, we replicate Table 2 from \cite{SVR17}, which measured the number of observed modes in a generator trained on the stacked MNIST dataset, as well as the KL divergence of the generated mode distribution.

\paragraph{Hyperparameters.}
%
%
For this experiment, we train each GAN on 128,000 samples, with a mini-batch size of 64. 
The generator's loss function is $-\log(D(\text{generated data}))$, and the discriminator's loss function is  
-log(D(real\ data))-log(1-D(generated\ data)).
We update the generator parameters twice and the discriminator parameters once in each mini-batch,
and train the networks over 50 epochs.
For testing, we generate 26,000 samples, and evaluate the empirical KL divergence and number of modes covered. 
Finally, we average these values over 10 runs of the entire pipeline. 

\paragraph{Architecture.}
In line with prior work \cite{SVR17}, we used a DCGAN-like architecture for these experiments, which is based on the code at \url{https://github.com/carpedm20/DCGAN-tensorflow}.
In particular, the generator and discriminator architectures are as follows:

Generator:
\begin{table}[H]
	\begin{center}
		\begin{tabular}{l|l|l|l|l|l}
			\hline
			layer&number of outputs & kernel size& stride & BN & activation function\\
			\hline
			Input: $z\sim U(-1,1)^{100}$&100&&&&\\
			Fully connected &2*2*512&&&Yes&ReLU\\
			Transposed Convolution & 4*4*256&5*5&2&Yes&ReLU\\
			Transposed Convolution & 7*7*128&5*5&2&Yes&ReLU\\
			Transposed Convolution & 14*14*64&5*5&2&Yes&ReLU\\
			Transposed Convolution & 28*28*3&5*5&2&&Tanh\\
			\hline
		\end{tabular}
	\end{center}
\end{table}
Discriminator (for PacDCGAN$m$):
\begin{table}[H]
	\begin{center}
		\begin{tabular}{l|l|l|l|l|l}
			\hline
			layer&number of outputs & kernel size& stride & BN & activation function\\
			\hline
			Input: $x\sim p_{data}^m$ or $G^m$&28*28*(3*$m$)&&&&\\
			Convolution & 14*14*64&5*5&2&&LeakyReLU\\
			Convolution & 7*7*128&5*5&2&Yes&LeakyReLU\\
			Convolution & 4*4*256&5*5&2&Yes&LeakyReLU\\
			Convolution & 2*2*512&5*5&2&Yes&LeakyReLU\\
			Fully connected&1&&&&Sigmoid\\
			\hline
		\end{tabular}
	\end{center}
\end{table}
	MD uses the same architecture as GAN, except that a minibatch discrimination layer is added before the output layer of discriminator.

\paragraph{Results.}
Results are shown in Table \ref{tbl:veegan2}.
The first four rows are copied directly from \cite{SVR17}.
The last three rows are computed using a basic DCGAN, with packing in the discriminator. 
We find that packing gives good mode coverage, reaching all 1,000 modes in every trial.
Given a DCGAN that can capture at most 99 modes on average (our mother architecture), 
the principle of packing, which is a small change in the architecture, 
is able to improve performance to capture all 1,000 modes.  
Again we see that packing the simplest DCGAN is sufficient to fully capture all the modes in this benchmark tests, 
and we do not pursue packing more complex baseline architectures. 
Existing approaches to mitigate mode collapse, such as ALI, Unrolled GANs,  VEEGAN, and MD are not able to capture as many modes. 
We also observe that MD is very unstable throughout training, which makes it capture even less modes than GAN.
One factor that contributes to MD's instability may be that MD requires too many parameters.
The number of discriminator parameters in MD is 47,976,773, whereas GAN has 4,310,401 and PacGAN4 only needs 4,324,801.



\begin{table}[ht]
	\begin{center}
  	\begin{tabular}{ l  r  r    }
    		\hline
		& \multicolumn{2}{c}{Stacked MNIST}  \\ \cline{2-3} 
    		 & Modes (Max 1000)& KL \\ \hline 
    		DCGAN \cite{RMC15} 	& 99.0 & 3.40 		\\
    		ALI  \cite{DBP16} & 16.0 &  5.40 		\\
    		Unrolled GAN \cite{MPP16} & 48.7 & 4.32 		\\
    		VEEGAN \cite{SVR17} 		& 150.0 &  2.95		\\
			Minibatch Discrimination \cite{SGZ16} & 24.5$\pm$7.67& 5.49$\pm$0.418\\
		\hline
		DCGAN (our implementation) 	& 78.9$\pm$6.46& 4.50$\pm$0.127  \\
		PacDCGAN2 (ours) 	& 1000.0$\pm$0.00& 0.06$\pm$0.003  \\
		PacDCGAN3 (ours) 	& 1000.0$\pm$0.00&0.06$\pm$0.003  \\
		PacDCGAN4 (ours) 	& 1000.0$\pm$0.00&0.07$\pm$0.005 \\
    		\hline
  	\end{tabular}
	\end{center}
	\caption{Two measures of mode collapse proposed in \cite{SVR17} 
	for the stacked MNIST dataset: number of modes captured by the generator and reverse KL divergence over the generated mode distribution. 
	The DCGAN, PacDCGAN, and MD  results are averaged over 10 trials, with standard error reported. 
	}
	\label{tbl:veegan2}
\end{table}

Note that other classes of GANs may also be able to learn most or all of the modes if tuned properly.
For example, \cite{MPP16} reports that regular GANs can learn all 1,000 modes even without unrolling if the discriminator is large enough, 
and if the discriminator is half the size of the generator, unrolled GANs recover up to 82\% of the modes when the unrolling parameter is increased to 10. 
To explore this effect, we conduct further experiments on unrolled GANs in Section \ref{sec:unrolled}.

\subsubsection{Unrolled GAN \cite{MPP16} experiment}
\label{sec:unrolled}
This experiment is designed to replicate Table 1 from Unrolled GANs \cite{MPP16}.
Unrolled GANs exploit the observation that iteratively updating discriminator and generator model parameters can contribute to training instability. 
To mitigate this, they update model parameters by computing the loss function's gradient with respect to $k\geq 1$ sequential discriminator updates, where $k$ is called the unrolling parameter. 
\cite{MPP16} reports that unrolling improves mode collapse as $k$ increases, at the expense of greater training complexity.

Unlike Section \ref{sec:veegan mnist}, which reported a single metric for unrolled GANs, 
this experiment studies the effect of the unrolling parameter and the discriminator size on the number of modes learned by a generator.
The key differences between these trials and the unrolled GAN row in Table \ref{tbl:veegan2} are four: (1) the unrolling parameters are different, (2) the discriminator sizes are different, (3) the generator and discriminator architectures are chosen according to Appendix E in \cite{MPP16}, and
(4) the total training time was 5x as long as \cite{MPP16}.
PacDCGAN uses the same generators and discriminators (except for input layer) as unrolled GAN in each experiment. MD uses the same architecture, except that a minibatch discrimination layer is added before the output layer of discriminator.


\paragraph{Results.}
Our results are reported in Table \ref{tbl:unrolled}.
The first four rows are copied from \cite{MPP16}.
As before, we find that packing seems to increase the number of modes covered.
Additionally, in both experiments, PacDCGAN finds more modes on average than Unrolled GANs with $k=10$, with lower reverse KL divergences between the mode distributions.
This suggests that packing has a more pronounced effect than unrolling.

We see that compared with PacGAN, MD has worse metrics in D=1/4G setting but has similar metrics in D=1/2G setting. In addition, we should note that MD requires much more discriminator parameters: 747 for PacGAN4 and 1,226,317 for MD in D=1/4G setting; 2,213 for PacGAN4 and 2,458,533 for MD in D=1/2G setting.

\begin{table}[ht]
	\begin{center}
  	\begin{tabular}{ l  r  r  r r r  }
    		\hline
		& \multicolumn{2}{c}{D is $1/4$ size of G } && \multicolumn{2}{c}{D is $1/2$ size of G }  \\ \cline{2-3}\cline{5-6}
    		 & Modes (Max 1000)& KL  & & Modes (Max 1000) & KL  \\ \hline
    		DCGAN \cite{RMC15} &30.6$\pm$20.73&5.99$\pm$0.42&& 628.0$\pm$140.9&2.58$\pm$0.75\\
    		Unrolled GAN,  1 step \cite{MPP16} &65.4$\pm$34.75&5.91$\pm$0.14&& 523.6$\pm$55.77&2.44$\pm$0.26\\
    		Unrolled GAN, 5 steps \cite{MPP16} &236.4$\pm$63.30&4.67$\pm$0.43&& 732.0$\pm$44.98&1.66$\pm$0.09\\
    		Unrolled GAN, 10 steps \cite{MPP16} &327.2$\pm$74.67&4.66$\pm$0.46&& 817.4$\pm$37.91&1.43$\pm$0.12\\
    		Minibatch Discrimination \cite{SGZ16} & 264.1$\pm$59.02 & 3.32$\pm$0.30&& 837.1$\pm$67.46 & 0.84$\pm$0.25\\
		\hline
		DCGAN (our implementation) 	& 78.5$\pm$17.56&5.21$\pm$0.19 && 487.7$\pm$34.59 & 2.24$\pm$0.15 \\
		PacDCGAN2 (ours) 	& 484.5$\pm$32.99&2.61$\pm$0.22 && 840.7$\pm$15.92 & 1.00$\pm$0.05 \\
		PacDCGAN3 (ours) 	& 601.3$\pm$32.18& 2.00$\pm$0.17 && 866.6$\pm$12.10& 0.90$\pm$0.04 \\
		PacDCGAN4 (ours) 	& 667.4$\pm$29.00 & 1.81$\pm$0.15 && 820.2$\pm$25.50& 1.15$\pm$0.14 \\
    		\hline
  	\end{tabular}
	\end{center}
	\caption{Modes covered and KL divergence for unrolled GANs and MD as compared to PacDCGANs for various  unrolling parameters, discriminator sizes, and the degree of packing. The DCGAN and PacDCGAN results are 
	averaged over 50 trials, with standard error reported. The MD results are averaged over 10 trials, with standard error reported.
	}
	\label{tbl:unrolled}
\end{table}

%
%
%
%
%
%
%
%
%
%

\input{celebA.tex}

%% file: celebA.tex
\subsection{CelebA experiment}
\label{sec:celeba}
In this experiment, we measure the diversity of images generated from the celebA dataset as proposed by Arora et al. \cite{AZ17}.  They suggest measuring the diversity by estimating the probability of collision in a finite batch of images sampled from the generator. If there exists at least one pair of near-duplicate images in the batch it is declared to have a collision. To detect collision in a batch of samples, they select the $20$ closest pairs from it according to the Euclidean distance in pixel space, and then visually identify if any of them would be considered duplicates by humans. For visual identification, we take majority vote of three human reviewers for each batch of samples. To estimate the probability we repeat the experiment $20$ times.

\begin{figure}[ht]
	\begin{center}
	\includegraphics[width=.23\textwidth]{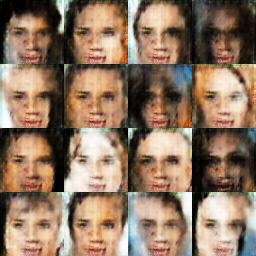}
	\put(-80,112){DCGAN}
	\hspace{0.1cm}
	\includegraphics[width=.23\textwidth]{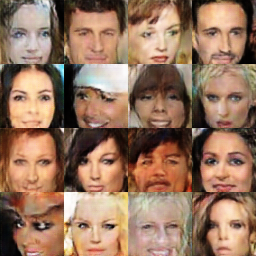}
	\put(-90,112){PacDCGAN2}
	\end{center}
	\caption{CelebA samples generated from DCGAN (left) and PacDCGAN2 (right) show PacDCGAN2 generates more diverse and sharper images.}
	\label{fig:celebA}
\end{figure}

We use DCGAN- unconditional, with JSD objective as described in \cite{RMC15} as the base architecture. We perform the experiment for different sizes of the discriminator while fixing the other hyper-parameters. The DCGAN \cite{RMC15} uses $4$ CNN layers with the number of output channels of each layer being $ dim \times\,1,2,4,8$. Thus the discriminator size is proportional to ${dim}^2$. Table \ref{tab:collision_table} shows probability of collision in a batch of size $1024$ for DCGAN and PacDCGAN2 for $dim \in \{16,32,64,80\}$.  Packing significantly improves diversity of samples. If the size of the discriminator is small, then packing also improves quality of the samples. Figure \ref{fig:celebA} shows samples generated from DCGAN and PacDCGAN2 for $dim = 16$. 
We note that DCGAN and PacDCGAN2 use approximately same number of parameters, $273$K and $274$K respectively.

\begin{table}[H]
	\begin{center}
		\begin{tabular}{l r r}
			\hline
			discriminator size & \multicolumn{2}{c}{probability of collision}  \\ \cline{2-3} 
			  & DCGAN &  PacDCGAN2 \\
			 \hline
			${d}^2$ & $1$ & $0.33$\\ 
			$4\,{d}^2$ & $0.42$ & $0$\\ 
			$16\,{d}^2$ & $0.86$ & $0$\\ 
			$25\,{d}^2$ & $0.65$ & $0.17$\\ 
			\hline
		\end{tabular}
	\end{center}
\caption{Probability of at least one pair of near-duplicate images being present in a batch of $1024$ images generated from DCGAN and PacDCGAN2 on celebA dataset show that PacDCGAN2 generates more diverse images.}
\label{tab:collision_table}
\end{table}

	\subsection{WGAN experiment} \label{app:WGAN}
	To verify that our packing idea can also work on Wasserstein loss, we compare WGAN with PacWGAN on stacked MNIST dataset. The experiment setting follows VEEGAN setting in Section \ref{sec:veegan mnist}, except: (1) remove all batch normalization layers in discriminator, and remove the sigmoid activation in the output layer of discriminator; 
	(2) use WGAN-GP loss instead of JSD loss; and (3) to showcase the difference between WGAN and PacGAN, we use smaller generators and discriminators. Specifically, the number of feature maps in each layer of discriminator and generator is a quarter of what used in Appendix \ref{sec:veegan mnist}.
	Table \ref{tbl:wgan} shows that PacWGANs discover more modes and achieves smaller KL divergence than WGAN. This suggests that the idea of packing improves mode collapse on GANs trained with 
	Wasserstein distance loss as well.
	
	\begin{table}[H]
		\begin{center}
			\begin{tabular}{ l  r  r    }
				\hline
				& \multicolumn{2}{c}{Stacked MNIST}  \\ \cline{2-3} 
				& Modes & KL \\ \hline 
				WGAN \cite{ACB17}	& 314.3$\pm$38.54& 2.44$\pm$0.170  \\
				PacWGAN2 (ours) 	& 927.6$\pm$22.19& 0.59$\pm$0.108  \\
				PacWGAN3 (ours)  	&948.7$\pm$21.43&0.50$\pm$0.089  \\
				PacWGAN4 (ours) 	& 965.7$\pm$19.07&0.42$\pm$0.094 \\
				\hline
			\end{tabular}
		\end{center}
		\caption{Two measures of mode collapse proposed in \cite{SVR17} 
			for the stacked MNIST dataset: number of modes captured by the generator and reverse KL divergence over the generated mode distribution. 
			All results are averaged over 10 trials, with standard error reported. 
		}
		\label{tbl:wgan}
	\end{table}

%% file: related.tex
\section{Related work}
\label{sec:related}

The idea of packing arises naturally by revisiting the formulation  of GANs, in a slightly different way from how GANs are  commonly explained in the  literature. 
Recall that a neural-network-based generator $G_w:\reals^d \to\reals^p$ 
maps noise $Z\in\reals^d$ with a simple distribution (such as uniform over the unit hypercube or Gaussian)
to the desired target distribution $P$ in target domain $\reals^p$. 
The target distribution $P$ is not known, but only observed via $n$ i.i.d.~samples 
$\{X_i\}_{i=1}^n$. 
Training the weights $w$ of the generator $G_w$ is of a fundamental interest:
given $\{X_i\}_{i=1}^n$, how do we train the generator weights $w$ 
such that the distribution of $G_w(Z)$ is close to the (unknown) distribution $P$?  

This question is affected by two important properties of neural-network-based generators: 
$(a)$ evaluating exact likelihoods is difficult, but sampling from the learned distribution is easy; and 
$(b)$ it is straightforward to evaluate the Jacobian matrix $J_G(Z_j)$ of this vector-valued function with respect to the weights $w$ with back-propagation. 
This has led to a natural iterative strategy of the following type:
at each iteration $t$, 
choose a function 
$D(\{\tX_j\}_{j=1}^m)$, 
called a discriminator or a critic, 
which measures how far (in distribution) 
the generated samples $\{ \tX_j \triangleq G(Z_j) \}_{j=1}^m$ are from the target samples $\{X_i\}_{i=1}^n$. 
The gradient of this function $\nabla_{\tX_j}D(\{\tX_j\}_{j=1}^m) \in \reals^p$ 
provides local information about which direction each sample $\tX_j$ should be moved to 
in order for the set of samples $\{X_j\}_{j=1}^m$ to be {\em jointly} closer to the target.   
This can be used to update the weights of the generator according to a standard gradient descent step: 
\begin{eqnarray}
	w^{(t)} \;\;=\;\; w^{(t-1)} - \alpha_t \sum_{j=1}^m \nabla_{\tX_j}D(\{\tX_j\}_{j=1}^m) \, \,J_G(Z_j) \;,
	\label{eq:rel_grad}	
\end{eqnarray}
where $J_G(Z_j)  \in \reals^{p \times d}$ is the Jacobian 
of the generator function $G_w(\cdot)$ with respect to the weights $w$ evaluated at $Z_j$, 
and $\alpha_t$ is the step size.  
The main challenge is in designing the right function $D(\{\tX_j\}_{j=1}^m)$, 
to achieve the goal of training a generator that produces realistic samples.

This framework, although presented in a different language, was introduced in the seminal paper 
of Goodfellow et al. in \cite{GPM14}, 
where a breakthrough in training generative models is achieved by 
introducing the concept of generative adversarial networks.
A particularly innovative choice of discriminator is proposed,  
based on $(a)$ 
the intuition of two competing parties playing a minimax game, 
and $(b)$ on the successes of deep neural networks in training classifiers. 
Concretely, the proposed discriminator $D(\{\tX_j\}_{j=1}^m )$ is 
implemented as a neural network 
$D_\theta: \reals^p\to\reals$ with weights $\theta$. 
With the right choice of an objective function, this can be {\em trained} to provide 
an estimator of the Jensen-Shannon divergence: $d_{\rm JS}(P,Q) \triangleq (1/2)\,d_{\rm KL}(P\| (P+Q)/2) + 
(1/2)\,d_{\rm KL}(Q\| (P+Q)/2)$. 
This has the desired property of (approximately) 
measuring how close the  samples generated 
from distribution $G_w(Z)\sim Q$ 
 are to the real distribution $P$. 
Precisely, \cite{GPM14} proposed 
using 
\begin{eqnarray}
	D(\{\tX_j\}_{j=1}^m ) \;\; = \;\; \frac1m \sum_{j=1}^m \log (1-D_{\theta^*}(\tX_j))  \;, 
	\label{eq:rel_loss}
\end{eqnarray}
where the neural network weight $\theta^*$ is 
the optimal solution trained on the 
target samples $\{X_i\}_{i=1}^n$ 
and additional fresh generated samples $\{\tX_j\}_{j=m+1}^{m+n}$, typically of the same size $n$. 
The proposed choice of the objective function is the standard cross entropy: 
\begin{eqnarray}
	 \theta^* \;\; \in \;\;  \arg \, \max_\theta \,\;\; \frac1n \sum_{i=1}^n \log (D_\theta(X_i)) + \frac1n \sum_{j=m+1}^{m+n} \log (1-D_\theta(\tX_j )) \;. 
	\label{eq:rel_gan}
\end{eqnarray}
Notice that in the formulation of \eqref{eq:rel_loss}, 
 $D(\{\tX_j\}_{j=1}^m)$ is decomposed as a summation over $m$ functions, 
 each involving only a {\em single} sample. 
Such a decomposition is not necessary, 
and leads to a natural question: 
 is there any gain in training GANs with general discriminator 
 $D(\{\tX_j\}_{j=1}^m)$ involving multiple samples {\em jointly}? 
We answer this question in the affirmative, 
with  both numerical experiments showing significant gain in mitigating mode collapse 
and theoretical analyses showing a fundamental connection between using samples jointly and mode collapse. 

We note that 
there is a family of GANs whose discriminators take pairs of images 
\cite{STS16,LCC17,BSA18}, albeit for very different reasons. 
These discriminators perform 
two-sample tests based on 
maximum mean discrepancy (MMD) by finding a kernel function, which naturally takes a pair of images as input. 
It is unknown if MMD-based GANs help in mitigating mode collapse, either theoretically or empirically. 
We believe this question is outside the scope of this paper, and poses an interesting future research direction. 
In the remainder of this section, we describe related work in this space, and how it relates to our packed interpretation of GANs.

\subsection{Challenges in training GANs}
\label{sec:rel_challenges}

The literature on GANs has documented three primary, closely-related challenges:
($i$) they are unstable to train, ($ii$) they are challenging to evaluate, and ($iii$) they exhibit mode collapse (more broadly, they do not generalize).
Much research has emerged in recent years addressing these challenges. 
Our work  explicitly addresses the challenge ($iii$). 
We give a brief overview of the related work on each of these challenges, and its relation to our work. 

\paragraph{Training instability.}
GANs' alternating generator and discriminator updates can lead to significant instability during training. 
This instability manifests itself as oscillating values of the loss function that  exceed variations  caused by minibatch processing \cite{Goo16}. 
Such variability makes it challenging to evaluate when training has converged, let alone which model one should choose among those obtained throughout the training process. 
This phenomenon is believed to arise because in practice, the learned distribution and the true distribution lie on disjoint manifolds in a high-dimensional space \cite{ACB17}. 
As such, the discriminator can often learn to perfectly distinguish generated and real samples. 
On real data, the discriminator (correctly) learns to output `1', and vice versa on generated data. 
This  is believed in GAN literature to cause 
the generator loss function to have a negligible gradient, leading to unstable parameter updates. 
In our work, we do not explicitly tackle instability; our theoretical results assume an optimal discriminator and infinite data samples. 
However, a better understanding of instability is of both practical and theoretical interest, albeit  orthogonal to the question of packing. 

Fundamentally, it is not well-understood \emph{why} training instability occurs. 
In \cite{LMPS17}, Li et al.~take a step towards theoretically understanding GAN training dynamics, suggesting that optimization techniques are partially responsible for instability. 
\cite{LMPS17} shows that for a toy distribution model (i.e., a Gaussian mixture), 
GANs using optimal discriminators are able to learn the underlying distribution, whereas first-order optimization methods exhibit oscillatory dynamics. 
Empirically, this observation seems to hold even with improved GAN optimization techniques, such as unrolled GANs \cite{MPP16}, 
despite recent work showing that gradient-descent-based optimization of GANs is locally stable \cite{NK17}.

Building on this intuition, several papers have proposed methods for mitigating instability, generally taking one of two approaches.
The first relies on changing the optimization objective function. 
Regular GANs optimize the Jensen-Shannon divergence between the true distribution and the learned one \cite{GPM14}.
Jensen-Shannon divergence can behave poorly in regions where the two distributions have nonoverlapping support \cite{ACB17},
so other works have proposed alternative distance metrics, including Wasserstein distance \cite{ACB17}, f-divergences \cite{NCT16,NCM17}, 
asymmetric KL divergences \cite{NLV17}, 
kernel maximum mean discrepancy for two-sample testing \cite{LCC17}, and neural network distance \cite{AGL17}. 
In practice, challenges can arise from trying to approximate these distance metrics; for instance, \cite{ACB17} uses the Kantorovich-Rubinstein dual to compute the Wasserstein-1 distance, which requires optimization over the set of 1-Lipschitz functions. 
This constraint is approximated in \cite{ACB17} by clipping the critic weights, which can lead to artificially stunted critic functions---a fact that was later tackled by using gradient regularization in WGAN-GP \cite{GAAD17}. 
In a similar vein, \cite{LSW17} altered the objective function by transforming the discriminator optimization into its dual form, which improves stability as we have only minimizations in both the outer and inner optimizations. 

Another method of changing the objective function is to introduce regularization.
In a sense, the fact that generators and discriminators are not trained to completion in practice is a simple form of regularization \cite{GPM14}.
Recently, other innovative choices for regularizations have been proposed, including 
weight clipping \cite{ACB17}, 
gradient regularization \cite{GAAD17,MNG17,NK17}, 
Tikhonov regularizer for training-with-noise \cite{RLN17}, 
adding noise to the gradient \cite{HRU17}, 
and 
spectral-norm regularization \cite{MKKY18}.

A conceptually different approach for improving training stability is to propose architectural changes that empirically improve training stability. 
For example, Salimans et al. proposed a number of heuristic tricks for improving the training of GANs, including minibatch discrimination, reference batch normalization, and feature mapping \cite{SGZ16}.
Our work most closely resembles minibatch discrimination from \cite{SGZ16}, which also inputs multiple images to the discriminator. 
We provide a detailed comparison between this proposed minibatch discriminator and ours later in this section. 


\paragraph{Evaluation Techniques.}
Generative models (including GANs) are notoriously difficult to evaluate.
Ideally, one would measure the distance between the true distribution and the learned one.
However, typical generative models can only produce \emph{samples} from a learned distribution, and on real datasets, the true distribution is often unknown. 
As such, prior work on GANs has used a number of heuristic evaluation techniques. 

The most common evaluation technique is visual inspection. 
Many papers produce a collection of generated images, and compare them to the underlying dataset \cite{GPM14,RMC15,Goo16}, 
or ask annotators to evaluate the realism of generated images \cite{SGZ16}.
This approach can be augmented by interpolating between two points in the latent space and illustrating that the GAN produces a semantically meaningful interpolation between the generated images \cite{DBP16}.
This approach is useful to the extent that some GANs produce visually unrealistic images, but it is expensive, unreliable, and it does not help identify generalization problems \cite{theis2015note}.
The most common attempt to estimate the visual quality of an image is the \emph{inception score}, which describes how precisely a classifier can classify generated images, thereby implicitly capturing some measure of image quality \cite{SGZ16}; this has become a de facto evaluation technique for GANs trained on realistic data \cite{SGZ16,FCA16,GAAD17,MKKY18}.

Another common approach involves estimating the likelihood of a holdout set of test data under the learned distributions. 
The learned distribution is estimated using a standard kernel density estimator (KDE)\cite{WBS16}. 
However, KDEs are known to have poor performance in high dimensions, and in practice, the error in KDE is often larger than the distance between real and learned distributions \cite{WBS16}.
Hence, it is unclear how meaningful such estimates are. 
One proposed approach uses annealed importance sampling (AIS) instead of KDE to estimate log-likelihoods \cite{WBS16}, with significantly increased accuracy levels. 

An increasing number of papers are using \emph{classification-based evaluation metrics}.
Naively, GANs trained on labelled datasets can pass their outputs through a pre-trained classifier. The classifier outputs indicate which modes are represented in the generated samples \cite{DBP16,SGZ16,SVR17}. 
This is useful for measuring the first type of mode collapse (missing modes), but it cannot reveal the second type (partial collapse within a mode). 
To provide a more nuanced view of the problem, \cite{SSM17} recently proposed a more general classification-based evaluation metric, in which they train a classifier on generated data and real data, and observe differences in classifier performance on a holdout set of test data. 
While this approach does not directly evaluate partial mode collapse, it is more likely to implicitly measure it by producing weaker classifiers when trained on generated data.
On datasets that are not labelled, some papers have relied on \emph{human} classification, asking human annotators to `discriminate' whether an image is real or generated \cite{DCF15}. 

 In a recent work in \cite{richardson2018gans},  
 it was empirically shown that Gaussian mixture models (GMM)
 can also generates realistic samples if trained efficiently, although the images are not as sharp as GAN generated samples. 
 However, trained GMMs do not suffer from mode collapse, 
 capture the underlying distribution more faithfully, and 
 provide interpretable representation of the statistical structures. 
One of the main contribution is a new evaluation technique. 
The domain of the samples is partitioned into 
bins in a data dependent manner on the training data. 
The histograms of the training data and the generated data on the bins 
are compared to give a measure on how close those two distributions are.

\paragraph{Mode Collapse/Generalization.}
Mode collapse collectively refer to the phenomenon of lack of divergence in the generated samples. 
This includes trained generators assigning low probability mass to significant subsets of the data distribution's support, and hence losing some modes.
This also includes the phenomenon of trained generators mapping two latent vectors that are far apart to the same or similar data samples. 
Mode collapse is a byproduct of poor generalization---i.e., the generator does not learn the true data distribution; this phenomenon is a topic of recent interest \cite{AGL17,AZ17}.
Prior work has observed two types of mode collapse: entire modes from the input data are missing from the generated data (e.g., in a dataset of animal pictures, lizards never appear), or the generator only creates images within a subset of a particular mode (e.g., lizards appear, but only lizards that are a particular shade of green) \cite{Goo16,TGB17,AZ17,DKD16,MPP16,reed2016generative}. 
These phenomena are not well-understood, but a number of explanatory hypotheses have been proposed:
\begin{enumerate}
\item The objective function is ill-suited to the problem \cite{ACB17},
potentially causing distributions that exhibit mode collapse to be local minima in the optimization objective function. 
\item Weak discriminators cannot detect mode collapse, either due to low capacity or a poorly-chosen architecture \cite{MPP16,SGZ16,AGL17,LMPS17}.
\item The maximin solution to the GAN game is not the same as the minimax solution \cite{Goo16}.
\end{enumerate}
The impact and interactions of these hypotheses are not well-understood, but we show in this paper that a packed discriminator can significantly reduce mode collapse, both theoretically and in practice.
In particular, the method of packing is simple, and leads to clean theoretical analyses.
We compare the proposed approach of packing to three main approaches in the literature for mitigating mode collapse:

(1) \emph{Joint Architectures.} 
The most common approach to address mode collapse involves an encoder-decoder architecture, in which the GAN learns an encoding $G^{-1}(X)$ from the data space to a lower-dimensional latent space, on top of the usual decoding $G(Z)$ from the latent space to the data space.
Examples include bidirectional GANs \cite{DBP16}, adversarially learned inference (ALI) \cite{DKD16}, and VEEGAN \cite{SVR17}. 
These joint architectures feed both the latent and the high-dimensional representation of each data point into the discriminator: 
$\{(Z_i,G(Z_i))\}$ for the generated data and $\{(G^{-1}(X_i),X_i)\}$ for the real data.
In contrast, classical GANs use only the decoder, and feed only high-dimensional representations into the discriminator. 
Empirically, training these components jointly seems to improve the GAN performance overall, while also producing useful feature vectors that can be fed into downstream tasks like classification.
Nonetheless, we find experimentally that using the same generator architectures and discriminator architectures, 
packing captures more  modes  than these joint architectures, with significantly less overhead in the architecture and computation.
Indeed, recent work shows theoretically that encoder-decoder architectures may be fundamentally unable to prevent mode collapse  \cite{arora2017theoretical}.



(2) \emph{Augmented Discriminators.} 
Several papers have observed that discriminators lose discriminative power by observing only one (unlabeled) data sample at a time \cite{Goo16,SGZ16}.
A natural solution for labelled datasets is to provide the discriminator with image labels.
This has been found to work well in practice \cite{CLJ16}, though it does not generalize to unlabelled data.
A more general technique is {\em minibatch discrimination} \cite{SGZ16}.
Like our proposed packing architecture, minibatch discrimination feeds an array of data samples to the discriminator. 
However, unlike packing, minibatch discrimination proposed in \cite{SGZ16} is complicated both computationally and conceptually, and highly sensitive to the delicate hyper-parameter choices. 
At a high level, the main idea in minibatch discrimination is to give the discriminator some side information coming 
from a minibatch, and use it together with each of the individual examples in the minibatch to classify each sample. 
The proposed complex architecture to achieve this goal is as follows. 

Let $f(X_i)$ denote a vector of (latent) features for input $X_i$ produced by some intermediate layer in the discriminator. A tensor $T$ is learned such that the tensor product $T[\id,\id,f(X_i)]$  gives a 
latent matrix representation $M_i$ of the input $X_i$. 
The notation $T[\id,\id,f(X_i)]$ indicates 
a tensor to matrix linear mapping, where you take the third dimension and apply 
a vector $f(X_i)$. 
The $L_1$ distance across the rows of the $M_i$'s are computed
for each pair of latent matrices in the minibatch to give a measure 
$c_b(X_i,X_j) = \exp(-\|M_{i,b} - M_{j,b}\|_{L_1}))$. 
This minibatch layer outputs $o(X_i)_b = \sum_{j=1}^n c_b(X_i,X_j)$. 
This is concatenated with the original latent feature $f(X_i)$ to be passed through the upper layers of the discriminator architecture.
While the two approaches start from a similar intuition that batching or packing multiple samples gives stronger discriminator, the proposed architectures are completely different. 
PacGAN is  easier to implement, quantitatively shows strong performance in experiments, and is principled: our theoretical analysis rigorously shows that packing is 
a principled way to use multiple samples at the discriminator.

More recently, a breakthrough in training GANs was achieved in 
\cite{KAL17}. 
By progressively training GANs of increasing resolutions, 
the authors were able to train, for the first time, 
on high quality CelebA datasets with size $1024\times1024$. 
This  produces 
 by far the most realistic looking faces.  
 One of the main innovations in the paper is to 
 compute a new feature ``minibatch std'' that intuitively captures how diverse the minibatch is, 
 and to append it to the rest of your features for the discriminator to see. 
 This is a much simpler way of capturing minibatch statistics, that resolves the 
 issue of sensitivity to hyperparameter tuning of the original minibatch idea of \cite{SGZ16}.



(3) \emph{Optimization-based solutions.}
Another potential source of mode collapse is imperfect optimization algorithms. 
Exact optimization of the GAN minimax objective function is computationally intractable, so
GANs typically use iterative parameter updates between the generator and discriminator: 
for instance, we update the generator parameters through $k_1$ gradient descent steps, followed by $k_2$ discriminator parameter updates. 
Recent work has studied the effects of this compromise, showing that iterative updates can lead to non-convergence in certain settings \cite{LMPS17}---a worse problem than mode collapse.
Unrolled GANs \cite{MPP16} propose a middle ground, in which the optimization takes $k$ (usually five) gradient steps into account when computing gradients. 
These unrolled gradients affect the generator parameter updates by better predicting how the discriminator will respond.
This approach is conjectured to spread out the generated samples, making it harder for the discriminator to distinguish real and generated data.
The primary drawback of this approach is computational cost; packing  achieves better empirical performance with smaller computational overhead  
and training complexity. 

\subsubsection{Theoretical Understanding of Generalization}
In parallel with efforts to reduce mode collapse, there has been work on fundamentally understanding the generalization properties of GANs. 
Our work is implicitly related to generalization in that packing allows GAN training to converge to distributions that are closer to the true underlying distribution, in the sense of exhibiting less $(\epsilon,\delta)$-mode collapse. 
However, we do not explicitly analyze the baseline generalization properties of existing GANs. 
Arora et al. made a breakthrough on this front ~in \cite{AGL17}. 
Recall that typical assumption in theoretical analyses are: $(a)$ infinite samples, which allow us to work with population expectations, and 
$(b)$ infinite expressive power at the discriminator. 
\cite{AGL17} addresses both of these assumptions in the following way:
first, to show that existing losses (such as Wasserstein loss \cite{ACB17} and cross entropy loss \cite{GPM14}) do not generalize, 
\cite{AGL17} relaxes both $(a)$ and $(b)$. 
Under this quite general setting,  a GAN is trained on these typical choices of losses with a target distribution of a spherical Gaussian. 
Then, using a discriminator with enough expressive power, the training loss will converge to its maximum, which is proven to be strictly bounded away from zero for this Gaussian example.  
The implication of this analysis is that a perfect generator with infinite expressive power still will not be able to generate the target Gaussian distribution, as it is penalized severely in the 
empirical loss defined by the training samples. 
This observation leads to the second contribution of the paper, where a proper distance is defined, called {\em neural network divergence}, 
which takes into account the finite expressive power of neural networks. 
It is shown that the neural  network  divergence  has 
much  better  generalization  properties  than  Jensen-Shannon  divergence  or  Wasserstein  distance. 
This implies that this new neural network distance can better capture how the GAN performs for a specific choice of loss function. 

Liu et al.~study the effects of discriminator family with finite expressive power and the distributional convergence properties of 
various choices of the loss functions in \cite{LBC17}.
It is shown that the restricted expressive power of the discriminator (including the popular neural-network-based discriminators) 
have the effect of encouraging moment-matching conditions to be satisfied. 
Further, it is shown that for a broad class of loss functions, convergence in the loss function implies 
distributional weak convergence, which generalizes known convergence results of \cite{SGF10,ACB17}. 
This work does not consider the finite-data regime of \cite{AGL17}.
A more fine-grained theoretical characterization of the distribution 
induced by the optimal generator is provided in \cite{LC18}. 
This is achieved by analyzing a restricted version of f-GAN and 
showing that the learned distribution is a solution to a mixture of maximum likelihood and method of moments.

Finally, Feizi et al.~address the effect of generator and discriminator architectures 
 for a simpler case of learning a single Gaussian distribution in \cite{FSXT17}. 
By connecting the loss function to supervised learning, 
the generalization performance of a simple LQG-GAN is analyzed where 
the generator is linear, the loss is quadratic, and the data is coming from a Gaussian distribution. 
An interesting connection between principal component analysis and the optimal generator of this particular GAN is made. 
The sample complexity of this problem is shown to be linear in the dimension, if the discriminator is constrained to be quadratic, where as for general discriminators the sample complexity can be much larger.

%% file: region.bbl
\begin{thebibliography}{10}

\bibitem{ACB17}
Martin Arjovsky, Soumith Chintala, and L{\'e}on Bottou.
\newblock Wasserstein {GAN}.
\newblock {\em arXiv preprint arXiv:1701.07875}, 2017.

\bibitem{AGL17}
Sanjeev Arora, Rong Ge, Yingyu Liang, Tengyu Ma, and Yi~Zhang.
\newblock Generalization and equilibrium in generative adversarial nets
  ({GANs}).
\newblock {\em arXiv preprint arXiv:1703.00573}, 2017.

\bibitem{arora2017theoretical}
Sanjeev Arora, Andrej Risteski, and Yi~Zhang.
\newblock Theoretical limitations of encoder-decoder gan architectures.
\newblock {\em arXiv preprint arXiv:1711.02651}, 2017.

\bibitem{AZ17}
Sanjeev Arora and Yi~Zhang.
\newblock Do gans actually learn the distribution? an empirical study.
\newblock {\em arXiv preprint arXiv:1706.08224}, 2017.

\bibitem{BSA18}
Miko{\l}aj Bi{\'n}kowski, Dougal~J Sutherland, Michael Arbel, and Arthur
  Gretton.
\newblock Demystifying mmd gans.
\newblock {\em arXiv preprint arXiv:1801.01401}, 2018.

\bibitem{Bla53}
David Blackwell.
\newblock Equivalent comparisons of experiments.
\newblock {\em The Annals of Mathematical Statistics}, 24(2):265--272, 1953.

\bibitem{BJPD17}
Ashish Bora, Ajil Jalal, Eric Price, and Alexandros~G Dimakis.
\newblock Compressed sensing using generative models.
\newblock {\em arXiv preprint arXiv:1703.03208}, 2017.

\bibitem{BPD18}
Ashish Bora, Eric Price, and Alexandros~G. Dimakis.
\newblock Ambientgan: Generative models from lossy measurements.
\newblock In {\em International Conference on Learning Representations (ICLR)},
  2018.

\bibitem{CLJ16}
Tong Che, Yanran Li, Athul~Paul Jacob, Yoshua Bengio, and Wenjie Li.
\newblock Mode regularized generative adversarial networks.
\newblock {\em arXiv preprint arXiv:1612.02136}, 2016.

\bibitem{CT92}
Thomas~M Cover and A~Thomas.
\newblock Determinant inequalities via information theory.
\newblock {\em SIAM journal on Matrix Analysis and Applications},
  9(3):384--392, 1988.

\bibitem{defferrard2016convolutional}
Micha{\"e}l Defferrard, Xavier Bresson, and Pierre Vandergheynst.
\newblock Convolutional neural networks on graphs with fast localized spectral
  filtering.
\newblock In {\em Advances in Neural Information Processing Systems}, pages
  3844--3852, 2016.

\bibitem{DCT91}
Amir Dembo, Thomas~M Cover, and Joy~A Thomas.
\newblock Information theoretic inequalities.
\newblock {\em Information Theory, IEEE Transactions on}, 37(6):1501--1518,
  1991.

\bibitem{DCF15}
Emily~L Denton, Soumith Chintala, Rob Fergus, et~al.
\newblock Deep generative image models using a laplacian pyramid of adversarial
  networks.
\newblock In {\em Advances in neural information processing systems}, pages
  1486--1494, 2015.

\bibitem{DKD16}
Jeff Donahue, Philipp Kr{\"a}henb{\"u}hl, and Trevor Darrell.
\newblock Adversarial feature learning.
\newblock {\em arXiv preprint arXiv:1605.09782}, 2016.

\bibitem{DBP16}
Vincent Dumoulin, Ishmael Belghazi, Ben Poole, Alex Lamb, Martin Arjovsky,
  Olivier Mastropietro, and Aaron Courville.
\newblock Adversarially learned inference.
\newblock {\em arXiv preprint arXiv:1606.00704}, 2016.

\bibitem{FSXT17}
Soheil Feizi, Changho Suh, Fei Xia, and David Tse.
\newblock Understanding gans: the lqg setting.
\newblock {\em arXiv preprint arXiv:1710.10793}, 2017.

\bibitem{FCA16}
Chelsea Finn, Paul Christiano, Pieter Abbeel, and Sergey Levine.
\newblock A connection between generative adversarial networks, inverse
  reinforcement learning, and energy-based models.
\newblock {\em arXiv preprint arXiv:1611.03852}, 2016.

\bibitem{Goo16}
Ian Goodfellow.
\newblock Nips 2016 tutorial: Generative adversarial networks.
\newblock {\em arXiv preprint arXiv:1701.00160}, 2016.

\bibitem{GPM14}
Ian Goodfellow, Jean Pouget-Abadie, Mehdi Mirza, Bing Xu, David Warde-Farley,
  Sherjil Ozair, Aaron Courville, and Yoshua Bengio.
\newblock Generative adversarial nets.
\newblock In {\em Advances in neural information processing systems}, pages
  2672--2680, 2014.

\bibitem{GAAD17}
Ishaan Gulrajani, Faruk Ahmed, Martin Arjovsky, Vincent Dumoulin, and Aaron
  Courville.
\newblock Improved training of {W}asserstein {GAN}s.
\newblock {\em arXiv preprint arXiv:1704.00028}, 2017.

\bibitem{HRU17}
Martin Heusel, Hubert Ramsauer, Thomas Unterthiner, Bernhard Nessler, and Sepp
  Hochreiter.
\newblock Gans trained by a two time-scale update rule converge to a local nash
  equilibrium.
\newblock In {\em Advances in Neural Information Processing Systems 30}, pages
  6629--6640. 2017.

\bibitem{Hin10}
Geoffrey Hinton.
\newblock A practical guide to training restricted boltzmann machines.
\newblock {\em Momentum}, 9(1):926, 2010.

\bibitem{ilyas2017robust}
Andrew Ilyas, Ajil Jalal, Eirini Asteri, Constantinos Daskalakis, and
  Alexandros~G Dimakis.
\newblock The robust manifold defense: Adversarial training using generative
  models.
\newblock {\em arXiv preprint arXiv:1712.09196}, 2017.

\bibitem{IS15}
Sergey Ioffe and Christian Szegedy.
\newblock Batch normalization: Accelerating deep network training by reducing
  internal covariate shift.
\newblock In {\em International Conference on Machine Learning}, pages
  448--456, 2015.

\bibitem{IZZ16}
Phillip Isola, Jun-Yan Zhu, Tinghui Zhou, and Alexei~A Efros.
\newblock Image-to-image translation with conditional adversarial networks.
\newblock {\em arXiv preprint arXiv:1611.07004}, 2016.

\bibitem{KOV14}
Peter Kairouz, Sewoong Oh, and Pramod Viswanath.
\newblock Extremal mechanisms for local differential privacy.
\newblock In {\em Advances in Neural Information Processing Systems (NIPS)},
  pages 2879--2887, 2014.

\bibitem{KOV15}
Peter Kairouz, Sewoong Oh, and Pramod Viswanath.
\newblock Secure multi-party differential privacy.
\newblock In {\em Advances in Neural Information Processing Systems (NIPS)},
  2015.

\bibitem{KOV17}
Peter Kairouz, Sewoong Oh, and Pramod Viswanath.
\newblock The composition theorem for differential privacy.
\newblock {\em IEEE Transactions on Information Theory}, 63(6):4037--4049, June
  2017.

\bibitem{KAL17}
Tero Karras, Timo Aila, Samuli Laine, and Jaakko Lehtinen.
\newblock Progressive growing of {GAN}s for improved quality, stability, and
  variation.
\newblock {\em arXiv preprint arXiv:1710.10196}, 2017.

\bibitem{KB14}
Diederik Kingma and Jimmy Ba.
\newblock Adam: A method for stochastic optimization.
\newblock {\em arXiv preprint arXiv:1412.6980}, 2014.

\bibitem{kingma2018glow}
Diederik~P Kingma and Prafulla Dhariwal.
\newblock Glow: Generative flow with invertible 1x1 convolutions.
\newblock {\em arXiv preprint arXiv:1807.03039}, 2018.

\bibitem{KW13}
Diederik~P Kingma and Max Welling.
\newblock Auto-encoding variational bayes.
\newblock {\em arXiv preprint arXiv:1312.6114}, 2013.

\bibitem{kipf2016semi}
Thomas~N Kipf and Max Welling.
\newblock Semi-supervised classification with graph convolutional networks.
\newblock {\em arXiv preprint arXiv:1609.02907}, 2016.

\bibitem{KH09}
Alex Krizhevsky and Geoffrey Hinton.
\newblock Learning multiple layers of features from tiny images.
\newblock 2009.

\bibitem{mnist}
Yann LeCun.
\newblock The mnist database of handwritten digits.
\newblock {\em http://yann. lecun. com/exdb/mnist/}, 1998.

\bibitem{LTH16}
Christian Ledig, Lucas Theis, Ferenc Husz{\'a}r, Jose Caballero, Andrew
  Cunningham, Alejandro Acosta, Andrew Aitken, Alykhan Tejani, Johannes Totz,
  and Zehan Wang.
\newblock Photo-realistic single image super-resolution using a generative
  adversarial network.
\newblock {\em arXiv preprint arXiv:1609.04802}, 2016.

\bibitem{LCC17}
Chun-Liang Li, Wei-Cheng Chang, Yu~Cheng, Yiming Yang, and Barnabas Poczos.
\newblock Mmd gan: Towards deeper understanding of moment matching network.
\newblock In {\em Advances in Neural Information Processing Systems 30}, pages
  2200--2210. 2017.

\bibitem{LMPS17}
Jerry Li, Aleksander Madry, John Peebles, and Ludwig Schmidt.
\newblock Towards understanding the dynamics of generative adversarial
  networks.
\newblock {\em arXiv preprint arXiv:1706.09884}, 2017.

\bibitem{LSW17}
Yujia Li, Alexander Schwing, Kuan-Chieh Wang, and Richard Zemel.
\newblock Dualing gans.
\newblock In {\em Advances in Neural Information Processing Systems 30}, pages
  5611--5621. 2017.

\bibitem{LBC17}
Shuang Liu, Olivier Bousquet, and Kamalika Chaudhuri.
\newblock Approximation and convergence properties of generative adversarial
  learning.
\newblock {\em arXiv preprint arXiv:1705.08991}, 2017.

\bibitem{LC18}
Shuang Liu and Kamalika Chaudhuri.
\newblock The inductive bias of restricted f-gans.
\newblock {\em arXiv preprint arXiv:1809.04542}, 2018.

\bibitem{LV07}
Tie Liu and Pramod Viswanath.
\newblock An extremal inequality motivated by multiterminal
  information-theoretic problems.
\newblock {\em Information Theory, IEEE Transactions on}, 53(5):1839--1851,
  2007.

\bibitem{LLW15}
Ziwei Liu, Ping Luo, Xiaogang Wang, and Xiaoou Tang.
\newblock Deep learning face attributes in the wild.
\newblock In {\em Proceedings of the IEEE International Conference on Computer
  Vision}, pages 3730--3738, 2015.

\bibitem{liu2015faceattributes}
Ziwei Liu, Ping Luo, Xiaogang Wang, and Xiaoou Tang.
\newblock Deep learning face attributes in the wild.
\newblock In {\em Proceedings of International Conference on Computer Vision
  (ICCV)}, 2015.

\bibitem{MNG17}
Lars Mescheder, Sebastian Nowozin, and Andreas Geiger.
\newblock The numerics of gans.
\newblock In {\em Advances in Neural Information Processing Systems 30}, pages
  1823--1833. 2017.

\bibitem{MPP16}
Luke Metz, Ben Poole, David Pfau, and Jascha Sohl-Dickstein.
\newblock Unrolled generative adversarial networks.
\newblock {\em arXiv preprint arXiv:1611.02163}, 2016.

\bibitem{MSC13}
Tomas Mikolov, Ilya Sutskever, Kai Chen, Greg~S Corrado, and Jeff Dean.
\newblock Distributed representations of words and phrases and their
  compositionality.
\newblock In {\em Advances in neural information processing systems}, pages
  3111--3119, 2013.

\bibitem{MT17}
Kyle Mills and Isaac Tamblyn.
\newblock Phase space sampling and operator confidence with generative
  adversarial networks.
\newblock {\em arXiv preprint arXiv:1710.08053}, 2017.

\bibitem{MKKY18}
Takeru Miyato, Toshiki Kataoka, Masanori Koyama, and Yuichi Yoshida.
\newblock Spectral normalization for generative adversarial networks.
\newblock In {\em International Conference on Learning Representations (ICLR)},
  2018.

\bibitem{NK17}
Vaishnavh Nagarajan and J.~Zico Kolter.
\newblock Gradient descent gan optimization is locally stable.
\newblock In I.~Guyon, U.~V. Luxburg, S.~Bengio, H.~Wallach, R.~Fergus,
  S.~Vishwanathan, and R.~Garnett, editors, {\em Advances in Neural Information
  Processing Systems 30}, pages 5591--5600. 2017.

\bibitem{NLV17}
Tu~Nguyen, Trung Le, Hung Vu, and Dinh Phung.
\newblock Dual discriminator generative adversarial nets.
\newblock In {\em Advances in Neural Information Processing Systems}, pages
  2667--2677, 2017.

\bibitem{NCM17}
Richard Nock, Zac Cranko, Aditya~K Menon, Lizhen Qu, and Robert~C Williamson.
\newblock f-gans in an information geometric nutshell.
\newblock In {\em Advances in Neural Information Processing Systems 30}, pages
  456--464. 2017.

\bibitem{NCT16}
Sebastian Nowozin, Botond Cseke, and Ryota Tomioka.
\newblock f-gan: Training generative neural samplers using variational
  divergence minimization.
\newblock In {\em Advances in Neural Information Processing Systems}, pages
  271--279, 2016.

\bibitem{nystrom2015bridges}
Nicholas~A Nystrom, Michael~J Levine, Ralph~Z Roskies, and J~Scott.
\newblock Bridges: a uniquely flexible hpc resource for new communities and
  data analytics.
\newblock In {\em Proceedings of the 2015 XSEDE Conference: Scientific
  Advancements Enabled by Enhanced Cyberinfrastructure}, page~30. ACM, 2015.

\bibitem{RMC15}
Alec Radford, Luke Metz, and Soumith Chintala.
\newblock Unsupervised representation learning with deep convolutional
  generative adversarial networks.
\newblock {\em arXiv preprint arXiv:1511.06434}, 2015.

\bibitem{reed2016generative}
Scott Reed, Zeynep Akata, Xinchen Yan, Lajanugen Logeswaran, Bernt Schiele, and
  Honglak Lee.
\newblock Generative adversarial text to image synthesis.
\newblock {\em arXiv preprint arXiv:1605.05396}, 2016.

\bibitem{richardson2018gans}
Eitan Richardson and Yair Weiss.
\newblock On gans and gmms.
\newblock {\em arXiv preprint arXiv:1805.12462}, 2018.

\bibitem{RLN17}
Kevin Roth, Aurelien Lucchi, Sebastian Nowozin, and Thomas Hofmann.
\newblock Stabilizing training of generative adversarial networks through
  regularization.
\newblock In {\em Advances in Neural Information Processing Systems}, pages
  2015--2025, 2017.

\bibitem{SW17}
Yunus Saatci and Andrew Wilson.
\newblock Bayesian gans.
\newblock In {\em Advances in Neural Information Processing Systems}, pages
  3624--3633, 2017.

\bibitem{SGZ16}
Tim Salimans, Ian Goodfellow, Wojciech Zaremba, Vicki Cheung, Alec Radford, and
  Xi~Chen.
\newblock Improved techniques for training gans.
\newblock In {\em Advances in Neural Information Processing Systems}, pages
  2234--2242, 2016.

\bibitem{SSM17}
Shibani Santurkar, Ludwig Schmidt, and Aleksander Madry.
\newblock A classification-based perspective on {GAN} distributions.
\newblock {\em arXiv preprint arXiv:1711.00970}, 2017.

\bibitem{SGF10}
Bharath~K Sriperumbudur, Arthur Gretton, Kenji Fukumizu, Bernhard
  Sch{\"o}lkopf, and Gert~RG Lanckriet.
\newblock Hilbert space embeddings and metrics on probability measures.
\newblock {\em Journal of Machine Learning Research}, 11(Apr):1517--1561, 2010.

\bibitem{SVR17}
Akash Srivastava, Lazar Valkov, Chris Russell, Michael Gutmann, and Charles
  Sutton.
\newblock Veegan: Reducing mode collapse in gans using implicit variational
  learning.
\newblock {\em arXiv preprint arXiv:1705.07761}, 2017.

\bibitem{Sta59}
AJ~Stam.
\newblock Some inequalities satisfied by the quantities of information of
  fisher and shannon.
\newblock {\em Information and Control}, 2(2):101--112, 1959.

\bibitem{STS16}
Dougal~J Sutherland, Hsiao-Yu Tung, Heiko Strathmann, Soumyajit De, Aaditya
  Ramdas, Alex Smola, and Arthur Gretton.
\newblock Generative models and model criticism via optimized maximum mean
  discrepancy.
\newblock {\em arXiv preprint arXiv:1611.04488}, 2016.

\bibitem{theis2015note}
Lucas Theis, A{\"a}ron van~den Oord, and Matthias Bethge.
\newblock A note on the evaluation of generative models.
\newblock {\em arXiv preprint arXiv:1511.01844}, 2015.

\bibitem{thekumparampil2018attention}
Kiran~K Thekumparampil, Chong Wang, Sewoong Oh, and Li-Jia Li.
\newblock Attention-based graph neural network for semi-supervised learning.
\newblock {\em arXiv preprint arXiv:1803.03735}, 2018.

\bibitem{TGB17}
Ilya Tolstikhin, Sylvain Gelly, Olivier Bousquet, Carl-Johann Simon-Gabriel,
  and Bernhard Sch{\"o}lkopf.
\newblock Adagan: Boosting generative models.
\newblock {\em arXiv preprint arXiv:1701.02386}, 2017.

\bibitem{towns2014xsede}
John Towns, Timothy Cockerill, Maytal Dahan, Ian Foster, Kelly Gaither, Andrew
  Grimshaw, Victor Hazlewood, Scott Lathrop, Dave Lifka, Gregory~D Peterson,
  et~al.
\newblock Xsede: accelerating scientific discovery.
\newblock {\em Computing in Science \& Engineering}, 16(5):62--74, 2014.

\bibitem{VG06}
Sergio Verd{\'u} and Dongning Guo.
\newblock A simple proof of the entropy-power inequality.
\newblock {\em IEEE Transactions on Information Theory}, 52(5):2165--2166,
  2006.

\bibitem{VPT16}
Carl Vondrick, Hamed Pirsiavash, and Antonio Torralba.
\newblock Generating videos with scene dynamics.
\newblock In {\em Advances in Neural Information Processing Systems 29}, pages
  613--621. 2016.

\bibitem{WBS16}
Yuhuai Wu, Yuri Burda, Ruslan Salakhutdinov, and Roger Grosse.
\newblock On the quantitative analysis of decoder-based generative models.
\newblock {\em arXiv preprint arXiv:1611.04273}, 2016.

\bibitem{yu2017seqgan}
Lantao Yu, Weinan Zhang, Jun Wang, and Yong Yu.
\newblock Seqgan: Sequence generative adversarial nets with policy gradient.
\newblock In {\em AAAI}, pages 2852--2858, 2017.

\bibitem{zaheer2017deep}
Manzil Zaheer, Satwik Kottur, Siamak Ravanbakhsh, Barnabas Poczos, Ruslan~R
  Salakhutdinov, and Alexander~J Smola.
\newblock Deep sets.
\newblock In {\em Advances in Neural Information Processing Systems}, pages
  3391--3401, 2017.

\bibitem{Z98}
Ram Zamir.
\newblock A proof of the fisher information inequality via a data processing
  argument.
\newblock {\em Information Theory, IEEE Transactions on}, 44(3):1246--1250,
  1998.

\end{thebibliography}
